\documentclass[10pt,journal,compsoc]{IEEEtran}

\usepackage{ragged2e}
\usepackage{times}
\usepackage{soul}
\usepackage{url}

\usepackage{hyperref}

\usepackage{microtype}
\usepackage{graphicx}
\usepackage{subfigure}
\usepackage{booktabs} 

\usepackage{hyperref}

\usepackage{amsmath}
\usepackage{amssymb}
\usepackage{mathtools}
\usepackage{amsthm}

\usepackage{enumitem}

\usepackage[colorinlistoftodos,prependcaption,textsize=tiny]{todonotes}

\usepackage{amsmath}

  \def\mC{{\mathcal C}}
  \def\mD{{\mathcal D}}

  \def\mL{{\mathcal L}}

  \def\mR{{\mathcal R}}

  \def\mX{{\mathcal X}}

  \DeclareMathAlphabet\mathbfcal{OMS}{cmsy}{b}{n}

  \def\0{{\bf 0}}
  \def\1{{\bf 1}}

  \def\bmm{{\bf m}}

  \def\bx{{\bf x}}
  \def\by{{\bf y}}

  \def\bx{{\bf x}}
  
  \def\by{{\bf y}}

  \DeclareMathOperator*{\argmax}{arg\,max}

\def\eg{\emph{e.g.}} 
\def\ie{\emph{i.e.}} 
 
\def\etc{\emph{etc.}} \def\vs{\emph{vs.}}
\def\wrt{{w.r.t.~}}

\usepackage[capitalize,noabbrev]{cleveref}

\def\camera{\textcolor{black}}
\def\black{\textcolor{black}}

\usepackage{balance}
\usepackage{xspace}
\usepackage{pifont}
\usepackage{dashrule}

\newcommand{\myeata}{EATA\xspace}
\newcommand{\myeta}{ETA\xspace}
\newcommand{\myeataC}{EATA-C\xspace}

\definecolor{chengreen}{RGB}{17, 120, 100}

\usepackage{threeparttable}
\usepackage{multicol,multirow}
\usepackage{booktabs} 
\usepackage{bbding}

\theoremstyle{plain}

\theoremstyle{definition}

\theoremstyle{remark}

\usepackage{colortbl}  
\usepackage[textsize=tiny]{todonotes}

\usepackage{wrapfig}

\ifCLASSOPTIONcompsoc
  \usepackage[nocompress]{cite}
\else
  \usepackage{cite}
\fi
\ifCLASSINFOpdf
\else
\fi

\usepackage{algorithm}
\usepackage{algorithmic}

\def\mytitle{Uncertainty-Calibrated Test-Time Model Adaptation without Forgetting}

\begin{document}
%
\title{\mytitle}
%
%
%
%

\author{Mingkui~Tan*, Guohao~Chen*, Jiaxiang Wu, Yifan Zhang, Yaofo Chen, Peilin Zhao,
        and~Shuaicheng Niu$^\dagger$
\IEEEcompsocitemizethanks{\IEEEcompsocthanksitem Mingkui Tan, Guohao Chen, Yaofo Chen, and Shuaicheng Niu are with the School of Software Engineering, South China University of Technology. Mingkui Tan is also with the Key Laboratory of Big Data and Intelligent Robot, Ministry of Education, Guangzhou, China. Guohao Chen is also with Pazhou Laboratory, Guangzhou, China. Shuaicheng Niu is also with Nanyang Technological University, Singapore. Email: mingkuitan@scut.edu.cn, secasper@mail.scut.edu.cn, chenyaofo@scut.edu.cn, niushuaicheng@gmail.com.
\IEEEcompsocthanksitem Jiaxiang Wu is with ByteDance, China. The majority of this work was conducted while at Tencent AI Lab. Email: jiaxiang.wu.90@gmail.com.
\IEEEcompsocthanksitem Yifan Zhang is with the School of Computing, National University of Singapore. Email: yifan.zhang@u.nus.edu.
\IEEEcompsocthanksitem Peilin Zhao is with Tencent AI Lab, China. Email: masonzhao@tencent.com.
}
\thanks{* Authors contributed equally. $\dagger$ Corresponding author.}
}

\markboth{Journal of \LaTeX\ Class Files, 2024}%
{Shell \MakeLowercase{\textit{et al.}}: \mytitle}

\IEEEtitleabstractindextext{%
\begin{abstract}
\justifying
Test-time adaptation (TTA) seeks to tackle potential distribution shifts between training and testing data by adapting a given model w.r.t. any testing sample. This task is particularly important when the test environment changes frequently. Although some recent attempts have been made to handle this task, we still face two key challenges: 1) prior methods have to perform backpropagation for each test sample, resulting in unbearable optimization costs to many applications; 2) while existing TTA solutions can significantly improve the test performance on out-of-distribution data, they often suffer from severe performance degradation on in-distribution data after TTA (known as catastrophic forgetting). To this end, we have proposed an Efficient Anti-Forgetting Test-Time Adaptation (EATA) method which develops an active sample selection criterion to identify reliable and non-redundant samples for test-time entropy minimization. To alleviate forgetting, EATA introduces a Fisher regularizer estimated from test samples to constrain important model parameters from drastic changes.
However, in \myeata, the adopted entropy loss consistently assigns higher confidence to predictions even when the samples are underlying uncertain, leading to overconfident predictions that underestimate the data uncertainty. To tackle this, we further propose \myeata with Calibration (\myeataC) to separately exploit the reducible model uncertainty and the inherent data uncertainty for calibrated TTA. Specifically, we compare the divergence between predictions from the full network and its sub-networks to measure the reducible model uncertainty, on which we propose a test-time uncertainty reduction strategy with divergence minimization loss to encourage consistent predictions instead of overconfident ones. To further re-calibrate predicting confidence on different samples, we utilize the disagreement among predicted labels as an indicator of the data uncertainty. Based on this, we devise a min-max entropy regularization to selectively increase and decrease predicting confidence for confidence re-calibration.
Note that \myeataC and \myeata are different on the adaptation objective, while \myeataC still benefits from the active sample selection criterion and anti-forgetting Fisher regularization proposed in \myeata. 
Extensive experiments on image classification and semantic segmentation verify the effectiveness of our proposed methods. 

\end{abstract}

}

\maketitle

\IEEEdisplaynontitleabstractindextext

\IEEEpeerreviewmaketitle

\ifCLASSOPTIONcompsoc
\IEEEraisesectionheading{\section{Introduction}\label{sec:introduction}}
\else
\section{Introduction}
\label{sec:introduction}
\fi

\IEEEPARstart{D}{eep}
neural networks (DNNs) have achieved excellent performance in many challenging tasks, including image classification~\cite{he2016deep}, video recognition~\cite{wang2018nonlocal,zeng2020dense,zeng2021graph,chen2021rspnet}, and many other areas~\cite{choi2018stargan,fan2020inf,xu2021towards}. One prerequisite behind the success of DNNs is that the test samples are drawn from the same distribution as the training data, which, however, is often violated in many real-world applications. In practice, test samples may encounter natural variations or corruptions (also called \emph{distribution shift}), such as changes in lighting resulting from weather changes and unexpected noises resulting from sensor degradation~\cite{hendrycks2019benchmarking,koh2021wilds}. Unfortunately, models are often very sensitive to such distribution shifts and suffer severe performance degradation. 

Recently, several attempts~\cite{sun2020test,wang2021tent,liu2021ttt++,zhang2021memo,zhang2022Self,wang2022continual} have been proposed to handle the distribution shifts by online adapting a model at test time (called \textit{test-time adaptation}). Test-time training (TTT)~\cite{sun2020test} first proposes this pipeline. Given a test sample, TTT first fine-tunes the model via rotation classification~\cite{gidaris2018unsupervised} and then makes a prediction using the updated model. Without the need of training an additional self-supervised head, Tent~\cite{wang2021tent} and MEMO~\cite{zhang2021memo} further leverage the prediction entropy for test-time adaptation, in which the adaptation only involves test samples and a trained model. Although recent test-time adaptation methods are effective at handling test shifts, in real-world applications, they still suffer from the following limitations.

\begin{table*}[t!]
\caption{Characteristics of problem settings that adapt a trained model to a potentially shifted test domain. `Offline' adaptation assumes access to the entire source or target dataset, while `Online' adaptation can predict a single or batch of incoming test samples immediately.}
\vspace{-0.15in}
\label{tab:diff-settings}
\newcommand{\tabincell}[2]{\begin{tabular}{@{}#1@{}}#2\end{tabular}}
\begin{center}
\begin{threeparttable}
    \resizebox{1.0\linewidth}{!}{
 	\begin{tabular}{l|cccccc|cc}
 	\toprule
 	 Setting & Source Data & Target Data & Training Loss & Testing Loss & Offline & Online & Source Acc. & Prediction Uncertainty \\
 	\midrule
        Fine-tuning   & \XSolidBrush & $\bx^t, y^t$ & $\mL(\bx^t,y^t)$ & -- & \Checkmark & \XSolidBrush & Not Considered & Not Considered \\
        Continual learning   & \XSolidBrush & $\bx^t, y^t$ & $\mL(\bx^t,y^t)$ & -- & \Checkmark & \XSolidBrush & Maintained & Not Considered \\
        Unsupervised domain adaptation   & $\bx^s,y^s$ & $\bx^t$ & $\mL(\bx^s,y^s)+\mL(\bx^s,\bx^t)$ & -- & \Checkmark & \XSolidBrush & Maintained & Not Considered \\
        Test-time training   & $\bx^s,y^s$ & $\bx^t$ & $\mL(\bx^s,y^s)+\mL(\bx^s)$ & $\mL(\bx^t)$ & \XSolidBrush & \Checkmark &  Not Considered  & Not Considered \\
        Fully test-time adaptation (FTTA) & \XSolidBrush & $\bx^t$ & \XSolidBrush & $\mL(\bx^t)$ & \XSolidBrush & \Checkmark & Not Considered  & Not Considered \\
        \midrule
        \myeata (ours)   & \XSolidBrush & $\bx^t$ & \XSolidBrush & $\mL(\bx^t)$ & \XSolidBrush & \Checkmark & Maintained  & Not Considered \\
        \myeata-C (ours)   & \XSolidBrush & $\bx^t$ & \XSolidBrush & $\mL(\bx^t)$ & \XSolidBrush & \Checkmark & Maintained  & Calibrated \\
    \bottomrule
	\end{tabular}
	}
\end{threeparttable}
\end{center}
\vspace{-0.1in}
\end{table*}

\textbf{Latency Constraints.} Since TTA adapts a given model during inference, the adaptation efficiency is paramount in scenarios where latency is a critical factor. Previous methods, such as Test-Time Training (TTT)~\cite{sun2020test} and MEMO~\cite{zhang2021memo}, often require performing multiple backward propagations for each test sample. However, the computation-intensive nature of backward propagation renders these methods impractical in situations where low latency is non-negotiable or computational resources are limited.

\textbf{Forgetting on In-Distribution Samples.} Prior methods often focus on boosting the performance of a trained model on out-of-distribution (OOD) test samples, ignoring that the model after test-time adaptation suffers a severe performance degradation (named \textit{forgetting}) on in-distribution (ID) test samples (see Figure~\ref{fig:imageC-forgetting-level-5}). An eligible test-time adaptation approach should perform well on both OOD and ID test samples simultaneously, since test samples often come from both ID and OOD domains in real-world applications.

\textbf{Over-Confident Predictions.} Existing methods like Tent~\cite{wang2021tent} and SAR~\cite{niu2023sar} primarily rely on test-time entropy minimization for model adaptation, which greedily enhances the model's confidence and minimizes the predictive uncertainty for test samples, without distinguishing between model-induced and data-induced uncertainties. Consequently, even when the input data is naturally complex or highly corrupted (\ie, with \textit{irreducible} data uncertainty), the model is forced to make one-hot confident predictions where it should remain uncertain, leading to over-confident and potentially incorrect outputs.
This phenomenon is particularly concerning in high-risk applications, such as autonomous driving~\cite{bojarski2016driving} and medical diagnosis~\cite{anwar2018medical}, posing potential safety risks.

To address the efficiency and forgetting issue, we have proposed an Efficient Anti-forgetting Test-time Adaptation (\myeata) method consisting of a sample-efficient optimization strategy and a weight regularizer. \myeata excludes unreliable samples characterized by high entropy values and redundant samples that are highly similar throughout the adaptation. In this case, we can reduce the total number of backward updates of test data streaming (improving efficiency) and enhance the model performance on OOD samples. Furthermore, \myeata devises an anti-forgetting regularizer to prevent the important weights of the model from changing a lot during the adaptation, where the weights' importance is measured based on Fisher information~\cite{kirkpatrick2017overcoming} via a small set of test samples. With this regularization, the model can continually adapt to OOD test samples without performance degradation on ID test samples.

To mitigate overconfidence, we differentiate between various origins of uncertainty in TTA: 1)~Reducible model uncertainty, which arises from not knowing the optimal model parameters to describe the data due to \textit{insufficient training}~\cite{gal2016dropout}; 2)~Irreducible data uncertainty that arises from inherent noise or variability in the data, \textit{cannot be reduced by additional training}~\cite{kendall2017uncertainties}. Based on their characteristics, we aim to reduce the model uncertainty at test time for domain adaptation, while accurately reflecting data uncertainty in model predictions to ensure confidence calibration.

To this end, we further devise \myeata with Calibration, namely \myeataC. Specifically, \myeataC estimates model uncertainty by measuring the divergence between predictions from the full network and its randomly sampled sub-networks.
By minimizing this divergence, our \myeataC reduces model uncertainty and promotes consistent, rather than overconfident, predictions for model adaptation during testing.
Additionally, we introduce a data uncertainty indicator based on prediction disagreement, which effectively detects ambiguous samples near decision boundaries where conflicting predictions are more likely to occur.
We then incorporate a min-max entropy regularizer to selectively adjust the prediction confidence based on this data uncertainty estimation.
Note that \myeataC differs from \myeata on the adaptation objective, while it still benefits from the active sample selection criterion and anti-forgetting Fisher regularization proposed in \myeata.
We summarize our main contributions as follows.

\begin{itemize}[leftmargin=*]
    \item \black{We propose an Efficient Anti-forgetting Test-time Adaptation (\myeata) method. Specifically, we reveal that test samples contribute differently to adaptation, and develop an active sample identification scheme to filter out non-reliable and redundant test samples from adaptation, thereby improving TTA efficiency. Moreover, we extend the label-dependent Fisher regularizer to test samples with pseudo label generation, which prevents drastic changes in important model weights and helps alleviate the issue of model forgetting on in-distribution test samples.}
    
    \item We further introduce \myeata with Calibration (\myeataC), which differentiates between reducible and irreducible uncertainty during testing to design a calibration-driven learning objective. Specifically, \myeataC estimates model uncertainty using the divergence between full network and sub-network predictions, incorporating a consistency loss to reduce this uncertainty for adaptation. Regarding data uncertainty, \myeataC leverages prediction disagreements and applies min-max entropy regularization to selectively adjust confidence for calibration enhancement.
    
    \item We demonstrate that our proposed \myeata method improves both the performance and efficiency of test-time adaptation and also alleviates the long-neglected catastrophic forgetting issue. Our \myeataC further achieves better performance and calibration, with computational and memory efficiency comparable to \myeata.
\end{itemize}

A short version of this work was published in ICML 2022~\cite{niu2022eata}. This paper extends our preliminary version from the following aspects: 1) We explore the calibrated test-time adaptation, which aims to provide calibrated predicting confidence that reflects the true likelihood of correctness during unsupervised adaptation; 2) To solve the over-confident issue, we develop a test-time consistency loss that leverages the reducible model uncertainty for calibrated uncertainty reduction, and devise a min-max entropy regularizer to re-calibrate predicting confidence based on the inherent data uncertainty; 3) We provide analyses about the impact of different uncertainty reduction strategies, empirically verifying that our consistency loss overcomes the issue of over-fitting and over-confident in entropy minimization loss when adapting to the test data; 4) We provide extensive new empirical evaluations on image classification and semantic segmentation tasks with various model architectures, demonstrating that \myeataC achieves substantially better performance and calibration over \myeata, \eg, {improving accuracy by 6.5\%, while reducing calibration error by relatively 64.9\% on ImageNet-C dataset with ViT-Base~\cite{dosovitskiy2020vit}}. 

\begin{figure*}[t]

\centering
\includegraphics[width=0.96\linewidth]{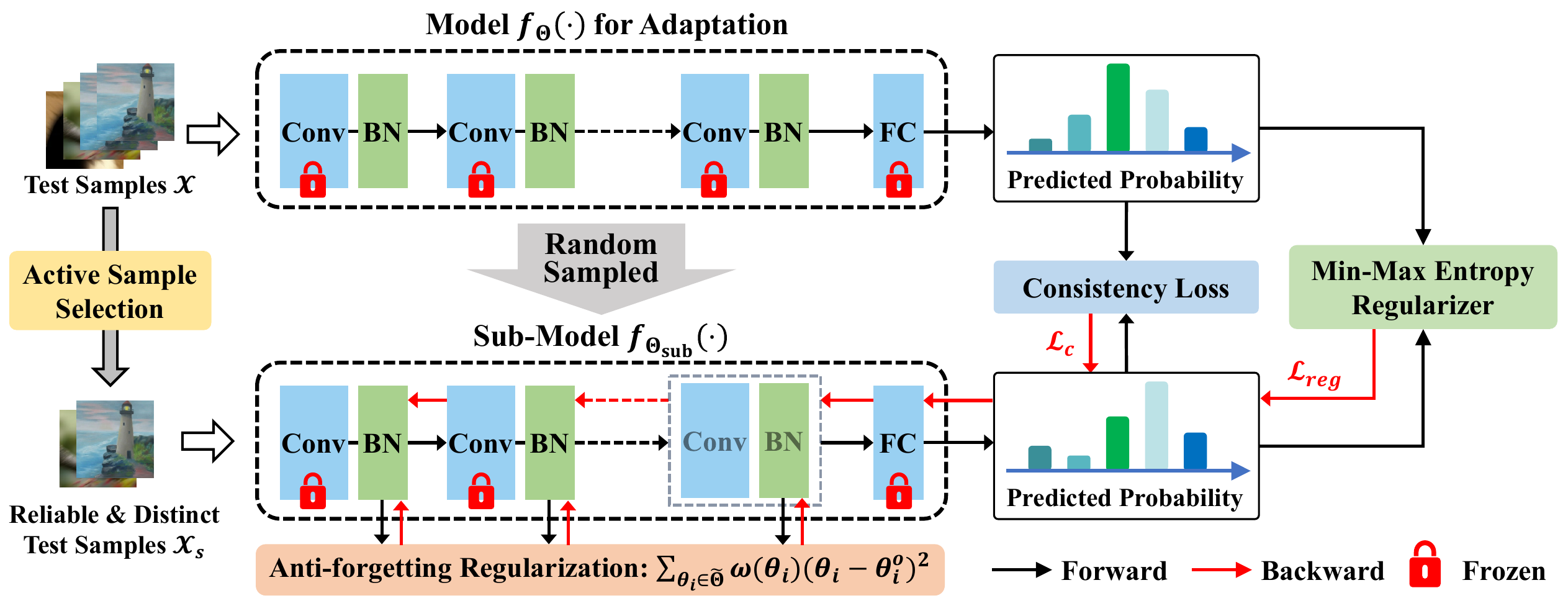}
\vspace{-0.1in}
\caption{An illustration of our proposed Efficient Anti-forgetting Test-time Adaptation with Calibration (\myeataC) method. During the test-time adaptation process, we update only the affine parameters of normalization layers in $f_{\Theta}$ and keep all other parameters frozen. Given a batch of incoming test samples $\mX\small{=}\{\bx_b\}_{b=1}^{B}$, we select the reliable and non-redundant ones $\mX_s$ with an active sample selection criterion to conduct model update, thereby enhancing adaptation efficiency. These samples are then used for calculating the proposed unidirectional consistency loss to minimize the model uncertainty. Additionally, we devise a min-max entropy regularizer for confidence re-calibration based on the data uncertainty of each sample. Lastly, we introduce an anti-forgetting regularizer which prevents the important model parameters in $\Theta$ from changing too much.}
\label{fig:eata-c overall}
\end{figure*}

\section{Related Work}

We divide the discussion on related works based on the different adaptation settings summarized in Table~\ref{tab:diff-settings} and further review existing methods for model's uncertainty calibration. 

\noindent\textbf{Test-Time Adaptation} (TTA) aims to improve model accuracy on OOD test data through model adaptation with  test samples.
Existing test-time training methods, \eg, TTT~\cite{sun2020test},  TTT++~\cite{liu2021ttt++}, TTT-MAE~\cite{gandelsman2022tmae}, and MT3~\cite{bartler2022mt3}, jointly train a source model via both supervised and self-supervised objectives, and then adapt the model via self-supervised objective at test time. This pipeline, however, necessitates both self-supervised head and test data in adaptation, while training such self-supervised head can be computation-consuming~\cite{gandelsman2022tmae}. 
To address this, some methods have been proposed to adapt a model with only test data, including batchnorm statistics adaptation~\cite{nado2020evaluating,schneider2020improving,reddy2024towards},  prediction consistency maximization over different augmentations~\cite{fleuret2021test}, and classifier adjustment~\cite{iwasawa2021test}.
Specifically, Tent~\cite{wang2021tent} updates the model to minimize the entropy of predictions at test time. MEMO~\cite{zhang2021memo} further augments test samples for marginal entropy minimization to enhance robustness.
Our work also alleviates the dependency on self-supervision heads and seeks to address the key limitations of prior works (\ie, efficiency hurdle, catastrophic forgetting, and overconfidence) to make TTA more practical in real-world applications.

\noindent\textbf{Continual Learning} (CL) aims to help the model remember the essential concepts that have been learned previously, alleviating the catastrophic forgetting issue when learning a new task~\cite{kirkpatrick2017overcoming,li2017learning,rolnick2019experience,farajtabar2020orthogonal,niu2021disturbance,Mittal_2021_CVPR}. 
In our work, we share the same motivation as CL and point out that test-time adaptation also suffers catastrophic forgetting (\ie, performance degradation on ID test samples), which makes TTA approaches unstable to deploy. To conquer this, we propose a simple yet effective solution to maintain the model performance on ID test samples (by only using test data) and meanwhile improve the performance on OOD test samples.

\noindent\textbf{Unsupervised Domain Adaptation (UDA).} Conventional UDA tackles distribution shifts by jointly optimizing a source model on both labeled source data and unlabeled target data, such as devising a domain discriminator to learn domain-invariant features~\cite{pei2018multi,saito2018maximum,zhang2020collaborative,zhang2020covid}. 
To avoid access to source data, recently CPGA~\cite{Qiu2021CPGA} generates feature prototypes for each category with pseudo-labeling. SHOT~\cite{liang2020we} learns a target-specific feature extractor by information maximization for representations alignment.
Nevertheless, such methods optimize offline via multiple epochs and losses. In contrast, our method adapts in an online manner and selectively performs once backward propagation for one given target sample, which is more efficient during inference.

\noindent\textbf{Uncertainty Calibration.} 
A calibrated model refers to whose predicting confidence reflects the true likelihood of correctness. 
Post-training processing methods \cite{guo2017calibration, naeini2015obtaining, zhang2020mix} re-calibrate a trained model by leveraging a labeled dataset within the target domain to estimate calibration error. In contrast, regularization-based methods~\cite{kumar2018trainable, seo2019learning, park2020calibrated, wang2020transferable} introduce auxiliary objectives to improve calibration at the training phase. Recently, SB-ECE~\cite{karandikar2021soft} proposes a differentiable estimation of calibration error as regularization to be jointly minimized. ESD~\cite{yoon2023esd} further reformulates the calibration objective in a class-wise manner to enhance calibration performance.
Nevertheless, these methods necessitate labeled data from the source or target domain, which limits their applicability. 
Unlike these methods, we seek to improve calibration with only access to unlabeled test data in an online manner in TTA context.

\section{Problem Formulation}

Without loss of generality, let $P\left( \bx \right)$ be the distribution of training data $\{\bx_i\}_{i=1}^{N}$  (namely $\bx_i \sim P\left( \bx \right)$) and $f_{\Theta^o}(\bx)$ be a \textbf{base model} trained on labeled training data $\{(\bx_i,y_i)\}_{i=1}^{N}$, where $\Theta^o$ denotes the model parameters. Due to the training process, the model $f_{\Theta^o}(\bx)$ tends to fit (or overfit) the training data.
During the inference state, the model shall perform well for  the in-distribution test data, namely $\bx \sim P\left( \bx \right)$. However, in practice, due to possible distribution shifts between training and test data, we may encounter many out-of-distribution test samples, namely $\bx \sim Q\left( \bx \right)$, where $Q\left( \bx \right) \neq P\left( \bx \right)$. In this case, the prediction would be very unreliable and the performance is also very poor.  

Test-time adaptation (TTA)~\cite{wang2021tent,zhang2021memo} aims at boosting the  out-of-distribution prediction performance by  doing model adaptation on test data only. Specifically, given a set of test samples $\{\bx_j\}_{j=1}^{M}$, where $\bx_j \sim Q \left( \bx \right)$ and $Q\left( \bx \right) \neq P\left( \bx \right)$, one needs to adapt $f_{\Theta}(\bx)$ to improve the prediction performance on test data in any cases. To achieve this, existing methods often seek to update the model by minimizing some unsupervised objective defined on test samples:
\begin{equation}\label{eq:tta_formula}
    \min_{\Tilde{\Theta}} \mL(\bx;\Theta),~ \bx \sim Q \left( \bx \right),
\end{equation}
where $\Tilde{\Theta} \subseteq \Theta$ denotes the free model parameters that should be updated. In general, the test-time learning objective $\mL(\cdot)$ can be formulated  as an entropy minimization problem~\cite{wang2021tent} or prediction consistency maximization over data augmentations~\cite{zhang2021memo}, \etc.

For existing TTA methods like  TTT~\cite{sun2020test} and MEMO~\cite{zhang2021memo}), during the test-time adaptation, we shall need to compute one round or even multiple rounds of backward computation for each sample, which is very time-consuming and also not favorable for latency-sensitive applications. Moreover, most methods assume that all the test samples are drawn from out-of-distribution (OOD). However, in practice, the test samples may come from both in-distribution (ID) and OOD. Simply optimizing the model on OOD test samples may lead to severe performance degradation on ID test samples. We empirically validate the existence of this issue in Figure \ref{fig:imageC-forgetting-level-5}, where the updated model has a consistently lower accuracy on ID test samples than the original model.

Moreover, existing entropy-based test-time adaptation methods like Tent~\cite{wang2021tent} and SAR~\cite{niu2023sar} consistently encourage the model to produce one-hot highly confident predictions. However, in practice, input test samples can be naturally complex and severely corrupted~\cite{hendrycks2019benchmarking}, resulting in irreducible data uncertainty. Ideally, these samples should be predicted with relatively low confidence to reflect their ambiguity. Nevertheless, the data uncertainty is often overlooked by methods based on entropy minimization,causing the adapted model to produce highly confident predictions (\textit{called overconfidence}), even when predictions should remain uncertain. Such misleading predictions raise potential safety concerns for real-world application scenarios.
We empirically demonstrate the issue of overconfidence in Figure~\ref{fig:consistency_entropy_comparison} and Table~\ref{tab:imagenet-c-level-5}.

\section{Uncertainty-Calibrated Efficient Anti-forgetting Test-time Adaptation}
In this section, we first propose an \textbf{Efficient Anti-forgetting Test-time Adaptation (\myeata)} method, which aims  to improve the efficiency of test-time adaptation (TTA) and tackle the catastrophic forgetting issue brought by existing TTA strategies  simultaneously.  
\myeata consists of two strategies.  \textbf{1)}  \emph{Sample-efficient entropy minimization} (c.f. Section~\ref{sec:efficient_adaptation}) aims to conduct efficient adaptation relying on an active sample selection strategy. Here, the sample selection process is to choose only active samples for backward propagation and therefore  improve the overall TTA efficiency  (\ie, less gradient backward propagation). To this end, we devise an active sample selection score, denoted by $S(\bx)$, to detect those reliable and non-redundant test samples from the test set for TTA. \textbf{2)} \emph{Anti-forgetting weight regularization} (c.f. Section~\ref{sec:adaptation_wo_forgetting}) seeks to alleviate knowledge forgetting by enforcing that the parameters, important for the ID domain, do not change too much in TTA. In this way, the catastrophic forgetting issue can be significantly alleviated. We illustrate EATA in Figure~\ref{suppl:fig:eata overall} in Supplementary.

To further address the overconfidence issue, we propose an \textbf{Efficient Anti-forgetting Test-time Adaptation with Calibration (\myeataC)} method. As shown in Figure~\ref{fig:eata-c overall}, we introduce a new consistency-based test-time learning objective for model uncertainty reduction (c.f. Section~\ref{sec:ttdm}), follow up a min-max entropy regularizer to re-calibrate the prediction uncertainty according to the inherent data uncertainty (c.f. Section~\ref{sec:min-max entropy}).

\subsection{Sample Efficient Entropy Minimization}\label{sec:efficient_adaptation}

For efficient test-time adaptation, we propose an active sample identification strategy to select  samples for backward propagation. Specifically, we design an active sample selection score for each sample, denoted by $S(\bx)$, based on two criteria: 1) samples should be \textbf{reliable} for test-time adaptation, and  2) the samples involved in optimization should be  \textbf{non-redundant}. By setting $S(\bx)\small{=}0$ for non-active samples, namely the unreliable and redundant samples, we can reduce unnecessary backward computation during test-time adaptation, thereby improving the prediction efficiency.

Relying on the sample score $S(\bx)$, following~\cite{wang2021tent,zhang2021memo}, we use entropy loss for model adaptations. Then, the sample-efficient entropy minimization is to minimize the following objective: 
\begin{align}
    \min_{\tilde{\Theta}}S(\bx) E(\bx;\Theta)\small{=-}S(\bx)\sum_{y\in\mC}f_{\Theta}(y|\bx)\text{log}f_{\Theta}(y|\bx), \label{eq:selective_entropy_optimization} 
\end{align}
where $\mC$ is the model output space. Here, the entropy loss $E(\cdot)$ is calculated over a batch of samples each time (similar to Tent~\cite{wang2021tent}) to avoid a trivial solution, \ie, assigning all probability to the most probable class. For efficient adaptation, we update $\tilde{\Theta}\small{\subseteq}\Theta$ with the affine parameters of all normalization layers.

\begin{figure}[t]
\centering\hspace{-0.1in}
\includegraphics[width=0.9\linewidth]{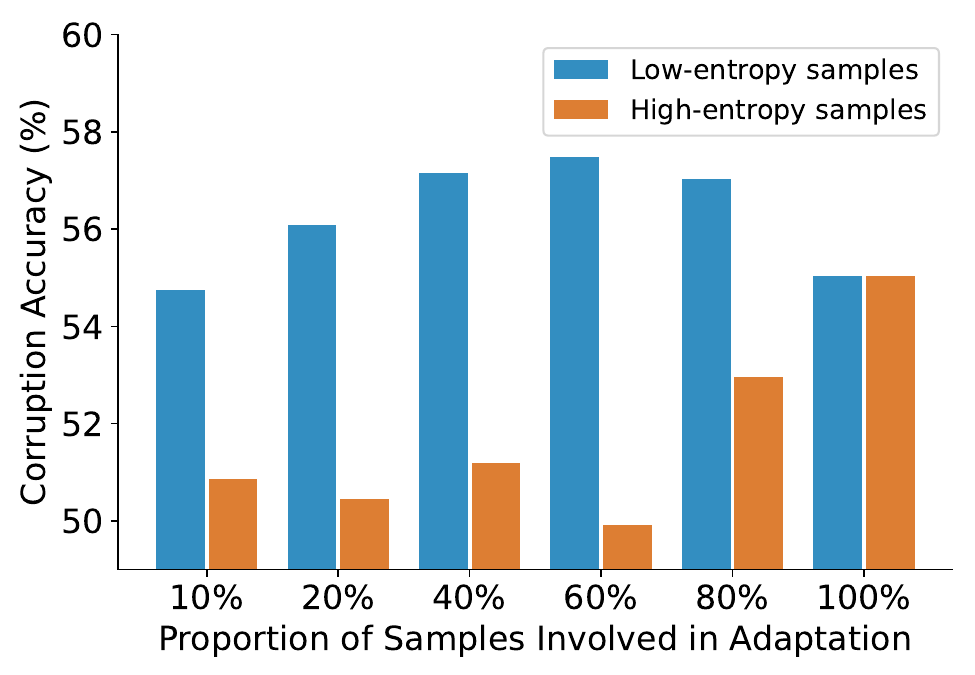}\vspace{-0.1in}
\caption{Effect of different test samples in test-time entropy minimization~\cite{wang2021tent}. We adapt a model on partial samples (top $p$\% samples with the highest or lowest entropy values), and then evaluate the adapted model on all test samples. Results are obtained on ImageNet-C (Gaussian noise, level 3) and ResNet-50 (base accuracy is 27.6\%). Introducing more samples with high entropy values into adaptation will hurt the adaptation performance.}
\label{fig:selective_entropy_motivation}
\end{figure}

\textbf{Reliable Sample Identification.}
Our intuition is that different test samples produce various effects in adaptation. To verify this, we conduct a preliminary study, where we select different proportions of samples (the samples are pre-sorted according to their entropy values $E(\bx;\Theta)$) for adaptation, and the resulting model is evaluated on all test samples. From Figure~\ref{fig:selective_entropy_motivation}, we find that: 1) adaptation on low-entropy samples makes more contribution than high-entropy ones, and 2) adaptation on test samples with very high entropy may hurt performance. The possible reason is that predictions of high-entropy samples are uncertain, so their gradients produced by entropy loss may be biased and unreliable. Following this, we name these low-entropy samples as reliable samples.
Based on the above observation, we propose an entropy-based weighting scheme to identify reliable samples and emphasize their contributions during adaptation. Formally, the entropy-based weight is given by:
\begin{align}
    S^{ent}(\bx) = \frac{1}{\exp \left[ E(\bx;\Theta) - E_0 \right]} \cdot \mathbb{I}_{\{E(\bx;\Theta)<E_0\}}(\bx), \label{eq:lambda_1st}
\end{align}
where $\mathbb{I}_{\{\cdot\}}(\cdot)$ is an indicator function,    $E(\bx;\Theta)$ is the predicted entropy regarding sample $\bx$, and $E_0$ is a pre-defined threshold. The above weighting function excludes high-entropy samples from adaptation and assigns higher weights to test samples with lower prediction uncertainties, allowing them to contribute more to model updates. Note that evaluating $S^{ent}(\bx)$ does not involve any gradient back-propagation.

\textbf{Non-redundant Sample Identification.}
Although Eqn.~(\ref{eq:lambda_1st}) helps to exclude partial unreliable samples, the remaining test samples may still have redundancy. For example, given two test samples that are mutually similar and both have a lower prediction entropy than $E_0$, we still need to perform gradient back-propagation for each of them according to Eqn.~(\ref{eq:lambda_1st}). However, this is unnecessary as these two samples produce similar gradients for model adaptation.

To further improve efficiency, we propose to exploit the samples that can produce different gradients for model adaptation. Recall that the entropy loss only relies on final model outputs (\ie, classification logits),  we further filter samples by ensuring the remaining samples have diverse model outputs. To this end, one straightforward method is to save the model outputs of all previously seen samples, and then compute the similarity between the outputs of incoming test samples and all saved model outputs for filtering. However, this method is computationally expensive at test time and memory-consuming with the increase of test samples. 

To address this, we exploit an exponential moving average technique to track the average model outputs of all seen test samples used for model adaptation. To be specific, given a set of model outputs of test samples, the moving average vector is updated recursively:
\begin{align}\label{eq:moving_average}
\bmm^t=\left\{\begin{array}{ll}
\bar{\by}^1, & \text { if } t=1 \\
\alpha\bar{\by}^t + (1-\alpha)\bmm^{t-1}, & \text { if } t>1
\end{array},\right.
\end{align}
where $\bar{\by}^t=\frac{1}{n}\sum_{k=1}^n\hat{\by}_k^t$ is the average model prediction of a mini-batch of $n$ test samples at the iteration $t$, and $\alpha\in[0,1]$. Following that, given a new test sample $\bx$ received at iteration $t > 1$, we compute the cosine similarity between its prediction $f_{\Theta}(\bx)$ and the moving average $\bmm^{t-1}$ (\ie, $cos(f_{\Theta}(\bx), \bmm^{t-1})$), which is then used to determine the diversity-based weight:
\begin{equation}
S^{div} \left( \bx \right) = \mathbb{I}_{\{cos(f_{\Theta}(\bx), \bmm^{t-1}) <\epsilon\}}(\bx),
\label{eq:lambda_div}
\end{equation}
where $\epsilon$ is a pre-defined threshold for cosine similarities. The overall sample-adaptive weight is then given by:
\begin{equation}\label{eq:lambda_2nd}
    S \left( \bx \right) = S^{ent} \left( \bx \right) \cdot S^{div} \left( \bx \right),
\end{equation}
which combines both entropy-based (as in Eqn.~\ref{eq:lambda_1st}) and diversity-based terms (as in Eqn.~\ref{eq:lambda_div}). Since we only perform gradient back-propagation for test samples with $S(\bx)>0$, the algorithm efficiency is further improved.

\textbf{Remark.} Given $M$ test samples $\mD_{test}=\{\bx_j\}_{j=1}^{M}$, the total number of reduced backward computations is given by $\mathbb{E}_{\bx\sim\mD_{test}} [\mathbb{I}_{\{S(\bx)=0\}}(\bx)]$, which is jointly determined by test data $\mD_{test}$, entropy threshold $E_{0}$, and cosine similarity threshold $\epsilon$.

\subsection{Anti-Forgetting with Fisher Regularization}\label{sec:adaptation_wo_forgetting}

In this section, we propose a new weighted Fisher regularizer (called anti-forgetting regularizer) to alleviate the catastrophic forgetting issue caused by test-time adaptation, \ie, the performance of a test-time adapted model may significantly degrade on in-distribution (ID) test samples. We achieve this through weight regularization, which only affects the loss function and does not incur any additional computational overhead for model adaptation. To be specific, we apply an importance-aware regularizer $\mR$ to prevent model parameters, important for the in-distribution domain, from changing too much during the test-time adaptation process~\cite{kirkpatrick2017overcoming}:
\begin{equation}\label{eq:weight_regularization}
    \mR(\tilde{\Theta},\tilde{\Theta}^o)=\sum_{\theta_i\in\tilde{\Theta}}\omega(\theta_i)(\theta_i - \theta_i^{o})^2,
\end{equation}
where $\tilde{\Theta}$ are parameters used for model update and $\tilde{\Theta}^o$ are the corresponding parameters of the original model. $\omega(\theta_i)$ denotes the importance of $\theta_i$ and we measure it via the diagonal Fisher information matrix as in elastic weight consolidation~\cite{kirkpatrick2017overcoming}. Here, the calculation of Fisher information $\omega(\theta_i)$ is non-trivial since we are inaccessible to any labeled training data. For the convenience of presentation, we leave the details of calculating $\omega(\theta_i)$ in the next subsection.

After introducing the anti-forgetting regularizer, the final optimization formula for \myeata is formulated as:
\begin{equation}\label{eq:overall_loss}
    \min_{\tilde{\Theta}} S(\bx) E(\bx;\Theta) + \beta \mR(\tilde{\Theta},\tilde{\Theta}^o),
\end{equation}
where $\beta$ is a trade-off parameter, $S(\bx)$ and $E(\bx;\Theta)$ are defined in Eqn.~(\ref{eq:selective_entropy_optimization}).

\textbf{Measurement of Weight Importance $\omega(\theta_i)$.} 
The calculation of Fisher information typically involves a set of labeled ID  training samples. However, in our problem setting, we are inaccessible to training data and the test samples are only unlabeled, which makes it non-trivial to measure the weight importance. To conquer this, we first collect a small set of unlabeled ID test samples $\{\bx_q\}_{q=1}^Q$, and then use the original trained model $f_{\Theta}(\cdot)$ to predict all these samples for obtaining the corresponding hard pseudo-label $\hat{y}_q$. Following that, we construct a pseudo-labeled ID test set $\mD_{F}=\{\bx_q, \hat{y}_q\}_{q=1}^Q$, based on which we calculate the fisher importance of model weights by:
\begin{equation}\label{eq:fisher_information}
    \omega(\theta_i) = \frac{1}{Q}\sum_{\bx_q\in\mD_{F}}\big(\frac{\partial}{\partial\theta_i^{o}}\mL_{CE}(f_{\Theta^o}(\bx_q),\hat{y}_q)\big)^2,
\end{equation}
where $\mL_{CE}$ is the cross-entropy loss. Here, we only need to calculate $\omega(\theta_i)$ once before performing test-time adaptation. 
\camera{Once calculated, we keep $\omega(\theta_i)$ fixed and apply it to any types of distribution shifts.}
Moreover, the unlabeled ID test samples are collected based on out-of-distribution detection techniques~\cite{liu2020energy,berger2021confidence}, which are easy to implement. Note that there is no need to collect too many ID test samples for calculating $\omega(\theta_i)$, \eg, 500 samples are enough for the ImageNet-C dataset. More empirical studies regarding this can be found in Figure~\ref{fig:number_samples_fisher}.

\subsection{Consistency-Based Uncertainty Minimization}
\label{sec:ttdm}

As mentioned in Section~\ref{sec:efficient_adaptation}, \myeata conducts model adaptation by prediction entropy minimization. This strategy aims to reduce uncertainty in predictions and learn decision boundaries in low-density regions of the test samples~\cite{chapelle2005semi,liang2023ttasurvey}. 
However, a persistent limitation of entropy minimization is its tendency to yield overly certain predictions, where the model is forced to make one-hot confident predictions even for ambiguous input data with irreducible data uncertainty. Thus, \myeata may still result in overconfident predictions that do not accurately reflect the inherent data uncertainty. To address this, we further propose \myeata with Calibration (\myeataC), shall be depicted in Sections~\ref{sec:ttdm} and~\ref{sec:min-max entropy}.

In \myeataC, we first propose a consistency-based loss to quantify and optimize the model uncertainty. Our method is inspired by MC Dropout~\cite{gal2016mcdropout}, which has shown promising performance in estimating the model uncertainty through the divergence of multiple dropout-enabled predictions. In our context, considering adaptation efficiency, we define the model uncertainty as the KL divergence~\cite{hinton2015kldivergence} between the full network prediction and randomly sampled sub-network prediction. Here, we use only two predictions and the former is indispensable since we select it as the final prediction. 
During TTA, we minimize this divergence to promote consistent predictions for model updates, rather than greedily increasing confidence which can result in overconfident outputs.

\textbf{Consistency Loss.}
Formally, let $\hat{\by}=f_{\Theta}(\bx)$ be the prediction of the full network \wrt sample $\bx$, and $\hat{\by}_{sub}=f_{\Theta_{sub}}(\bx)$ be that of the sub-network. The consistency loss is defined as follows:
\begin{align}
    \mL_{c}(\bx) &= D_{KL}(\hat{\by}_{sub}, \hat{\by}_{fuse}),
    \label{eq:consistency loss} \\
    \hat{\by}_{fuse} = (&\hat{\by} + (1-p)\cdot \hat{\by}_{sub})/(2-p), \label{eq:output_smoothing}
\end{align}
where $D_{KL}(\cdot||\cdot)$ denotes Kullback-Leibler divergence~\cite{hinton2015kldivergence} and $p$ is a constant for smoothing. Here, we calculate the divergence between $\hat{\by}_{sub}$ and $\hat{\by}_{fuse}$ (rather than $\hat{\by}$), since encouraging the sub-network to achieve the same performance as the full one is relatively hard. Thus, inspired by label smoothing~\cite{szegedy2016labelsm}, we softly fuse $\hat{\by}$ and $\hat{\by}_{sub}$ in Eqn.~(\ref{eq:output_smoothing}) for divergence optimization. Note that, during optimization, we conduct a unidirectional alignment from the sub-network to the full network, as the full network typically exhibits stronger generalization capabilities. To this end, we detach the gradient from $\hat{\by}$ and concentrate the optimization solely on $\hat{\by}_{sub}$. This strategy is designed to facilitate knowledge transfer from the full network to its sub-network, thereby enhancing the sub-network's performance while reducing the full networks' model uncertainty to adapt the network to the test domain.

\textbf{Remark on Efficiency.} Although consistency loss requires two forward passes from both the full network and the sub-network for each sample, the full network’s forward pass is gradient-free without back-propagation and the sub-network’s forward/backward pass is less computationally intensive. Moreover, we only perform sub-network's forward/backward passes on the selected reliable and non-redundant samples as outlined in Algorithm~\ref{alg:EATA-C overall}. As a result, the use of consistency loss remains efficient as shown in Table~\ref{tab:efficiency_comparison}.

\begin{algorithm}[t]
	\caption{The pipeline of proposed \myeata and \myeataC.}
	\label{alg:EATA-C overall}
	\begin{algorithmic}[1]
    \REQUIRE{Test samples $\mD_{test}\small{=}\{\bx_j\}_{j=1}^{M}$, the trained model $f_{\Theta}(\cdot)$,  ID samples $\mD_{F}\small{=}\{\bx_q\}_{q=1}^Q$, batch size $B$.}
    \FOR{a batch $\mX\small{=}\{\bx_b\}_{b=1}^{B}$ in $\mD_{test}$}
    \STATE Calculate predictions $\hat{y}$ for all $\bx \in \mX$ via $f_{\Theta}(\cdot)$.
    \STATE \texttt{For \myeata:} 
    \STATE ~~~Calculate sample selection score $S(\bx)$ via Eqn.~(\ref{eq:lambda_2nd}).
    \STATE ~~~Update model ($\tilde{\Theta} \subseteq\Theta$) with Eqn.~(\ref{eq:overall_loss}).
    \STATE \texttt{For \myeataC:}
    \STATE ~~~Select reliable and distinct samples $\mX_{s}$ via Eqn.~(\ref{eq:eata-c selection}).
    \STATE ~~~Sample $f_{\Theta_{sub}}(\cdot)$ from $f_{\Theta}(\cdot)$ via stochastic depth~\cite{huang2016depth}.
    \STATE ~~~Calculate predictions $\hat{y}_{sub}$ for $\bx \in \mX_{s}$ via $f_{\Theta_{sub}}(\cdot)$.
    \STATE ~~~Compute consistency loss $\mL_{c}(\bx)$ based on Eqn.~(\ref{eq:consistency loss})
    \STATE ~~~Calibrate confidence with entropy loss via Eqn.~(\ref{eq:min_max_entropy_regularization})
    \STATE ~~~Update model ($\tilde{\Theta} \subseteq\Theta$) with Eqn.~(\ref{eq:eata-c objective}).
    \ENDFOR
    \ENSURE The predictions $\{\hat{y}\}_{j=1}^M$ for all $\bx \in \mD_{test}$.
	\end{algorithmic}
\end{algorithm}

\subsection{Calibrated Min-Max Entropy Regularization}
\label{sec:min-max entropy}

In this section, we re-calibrate the model's prediction uncertainty in a manner that is sensitive to individual samples. This process involves categorizing samples into two distinct groups—`certain' and `uncertain'—based on the aforementioned prediction consistency. This design is inspired by margin-based learning approaches~\cite{balcan2007margin,khan2019striking} which indicated that samples near decision boundaries are inherently more uncertain and have been well justified with theoretical guarantees. Specifically, we achieve categorization by comparing the predicted labels between the full network and a sub-network. Samples that exhibit mismatched predictions are deemed `uncertain', suggesting their proximity to decision boundaries and high intrinsic data uncertainty.
Note that unlike the consistency loss that measures model uncertainty, in which samples across the data space may yield low consistency loss, data uncertainty is reflected more prominently through prediction disagreements near the decision boundary (see Figure~\ref{fig:toy_figure} for illustration). For identified uncertain samples, we aim to lower their predictive confidence by maximizing the prediction entropy, effectively acknowledging the model's lack of confidence in these cases. Conversely, for samples where predictions are consistent, labeled as `certain', we conduct the opposite strategy, \ie, boosting prediction confidence through entropy minimization. Formally, this min-max entropy regularization optimization problem is defined by:
\begin{align}
    \label{eq:min_max_entropy_regularization}
    \min_{\tilde{\Theta}_{sub}} C(\bx) E(\bx;\Theta_{sub}),~~~~~~~~~~~~~~~~~ \\
    \label{eq:complementary_certainty}
     C(\bx) = 
    \begin{cases}
        1, & \text{if $\argmax(\hat{\by})=\argmax(\hat{\by}_{sub})$,} \\
        -1, & \text{if $\argmax(\hat{\by})\neq\argmax( \hat{\by}_{sub})$,}\\
    \end{cases}
\end{align}
where $\hat{\by}$ and $\hat{\by}_{sub}$ denote the prediction of the full and sub-network respectively, $\Theta_{sub}$ denotes the parameters of the sub-network and $\tilde{\Theta}_{sub}\subset\Theta_{sub}$ denotes parameters involved in model adaptation. Note that we only update the affine parameters of the sub-network considering efficiency as mentioned in Section~\ref{sec:ttdm}.

\textbf{Overall Objective.}
\black{The methods proposed in Sections~\ref{sec:ttdm} and \ref{sec:min-max entropy} are devised to address the overconfident issue in TTA, but still suffer from catastrophic forgetting when important model weights for the in-distribution domain are significantly modified during adaptation. Therefore, we jointly optimize the model with the anti-forgetting regularizer and further reduce the required backward computations with the active sample selection criterion in EATA. Then, the overall objective of \myeataC can be formulated as: }
\begin{equation}\label{eq:eata-c objective}
    \min_{\tilde{\Theta}} S_{c}(\bx)\Big(\mL_{c}(\bx) + \alpha C(\bx) E(\bx;\Theta_{sub})\Big) + \beta \mR(\tilde{\Theta},\tilde{\Theta}^o).
\end{equation}
where $\alpha$ and $\beta$ are balance factors, $\mR(\tilde{\Theta},\tilde{\Theta}^o)$ is the fisher regularizer defined in Eqn.~(\ref{eq:weight_regularization}), and $S_{c}(\bx)$ is the joint indicator function in Eqn.~(\ref{eq:lambda_1st}) and Eqn.~(\ref{eq:lambda_div}) to select reliable and non-redundant samples. To be specific, $S_{c}(\bx)$ is defined by:
\begin{equation}\label{eq:eata-c selection}
     S_{c}(\bx) =  S^{div}(\bx) \cdot \mathbb{I}_{\{E(\bx;\Theta)<E_0\}}(\bx).
\end{equation}
We summarize the overall pipeline of our proposed \myeataC and \myeata in Algorithm~\ref{alg:EATA-C overall}.

\section{Experiments}

\begin{table*}[t]
    \caption{Comparison with state-of-the-art methods on ImageNet-C with the highest severity level 5 regarding \textbf{Corruption Accuracy(\%, $\uparrow$)} and \textbf{Expected Calibration Error(\%,$\downarrow$)}. ``BN" and ``LN" denote batch and layer normalization, respectively. The \textbf{bold} number indicates the best result and the \underline{underlined} number indicates the second best result. All results are evaluated under the \textbf{lifelong adaptation scenario} except for Tent~\cite{wang2021tent} and MEMO~\cite{zhang2021memo}, which suffer severely from error accumulation. We use * and $^\dagger$ to denote episodic and single-domain adaptation, respectively.}
    \vspace{-0.1in}
    \label{tab:imagenet-c-level-5}
\newcommand{\tabincell}[2]{\begin{tabular}{@{}#1@{}}#2\end{tabular}}
 \begin{center}
 \begin{threeparttable}
    \resizebox{1.0\linewidth}{!}{
        \begin{tabular}{c|l|c|ccc|cccc|cccc|cccc|c|cc}
        \multicolumn{3}{c}{} & \multicolumn{3}{c}{Noise} & \multicolumn{4}{c}{Blur} & \multicolumn{4}{c}{Weather} & \multicolumn{4}{c}{Digital} & \multicolumn{1}{c}{} & \multicolumn{2}{c}{Average} \\
        \toprule
         Model & Method & Metric & Gauss. & Shot & Impul. & Defoc. & Glass & Motion & Zoom & Snow & Frost & Fog & Brit. & Contr. & Elastic & Pixel & JPEG & Avg. & \#Forwards & \#Backards \\
        \midrule

        \multirow{22}{*}[-4.5ex]{R-50 (BN)} & \multirow{2}{*}{Source} & Acc. & 1.8 & 3.0 & 1.7 & 18.2 & 10.1 & 13.4 & 20.8 & 14.0 & 22.1 & 21.9 & 58.7 & 5.3 & 17.6 & 22.1 & 37.5 & 17.88 & \multirow{2}{*}{50,000} & \multirow{2}{*}{0} \\ & & ECE & 16.6 & 16.1 & 15.9 & 1.8 & 10.7 & 10.7 & 14.7 & 25.3 & 12.9 & 16.7 & 2.3 & 6.7 & 22.9 & 10.5 & 6.0 & 12.64 & & \\
                \cmidrule{2-21}
        & \multirow{2}{*}{BN Adapt}  & Acc. & 15.8 & 16.7 & 15.3 & 18.7 & 19.3 & 29.8 & 41.7 & 35.8 & 35.0 & 50.5 & 65.9 & 18.1 & 49.3 & 51.7 & 42.0 & 33.70 & \multirow{2}{*}{50,000} & \multirow{2}{*}{0} \\ & & ECE & 1.1 & 0.8 & 1.0 & 3.0 & 1.3 & 0.8 & 3.4 & 1.1 & 1.0 & 5.4 & 1.4 & 7.6 & 4.3 & 3.8 & 4.8 & \textbf{2.72} & &\\
        \cmidrule{2-21}
        & \multirow{2}{*}{Tent$^\dagger$}   & Acc. & 28.0 & 30.1 & 28.1 & 29.9 & 29.5 & 42.2 & 49.7 & 46.2 & 41.5 & 57.7 & 67.1 & 30.0 & 55.7 & 58.3 & 52.5 & 43.11 & \multirow{2}{*}{50,000} & \multirow{2}{*}{50,000} \\ & & ECE & 11.7 & 11.2 & 11.1 & 12.6 & 12.3 & 7.7 & 5.4 & 6.5 & 8.8 & 3.4 & 2.9 & 21.9 & 3.5 & 3.7 & 4.0 & 8.46 & &\\
        \cmidrule{2-21}
        & \multirow{2}{*}{MEMO*}   & Acc. & 6.8 & 8.5 & 7.5 & 20.5 & 13.4 & 19.8 & 25.8 & 22.1 & 27.7 & 27.6 & 60.9 & 11.3 & 24.4 & 32.2 & 37.9 & 23.09 & \multirow{2}{*}{50,000$\times$65} & \multirow{2}{*}{50,000$\times$64} \\ & & ECE & 24.1 & 24.2 & 22.9 & 5.3 & 19.3 & 14.8 & 23.4 & 30.4 & 18.7 & 24.6 & 7.2 & 14.9 & 29.4 & 19.3 & 13.6 & 19.47 & &\\
        \cmidrule{2-21}
        & \multirow{2}{*}{CoTTA}   & Acc. & 19.9 & 31.8 & 35.3 & 30.4 & 34.4 & 40.2 & 43.2 & 39.3 & 38.6 & 47.7 & 51.8 & 36.0 & 43.5 & 46.7 & 43.1 & 38.80 & \multirow{2}{*}{152,315} & \multirow{2}{*}{50,000} \\ & & ECE & 3.6 & 17.4 & 21.5 & 27.8 & 30.4 & 31.2 & 31.6 & 33.2 & 34.2 & 32.5 & 33.1 & 36.0 & 34.2 & 36.1 & 35.9 & 29.25 & &\\
        \cmidrule{2-21}
        & \multirow{2}{*}{SAR}   & Acc. & 29.6 & 38.4 & 37.8 & 31.5 & 32.8 & 41.4 & 48.6 & 42.9 & 40.2 & 53.3 & 63.7 & 37.7 & 53.0 & 56.3 & 52.3 & 43.96 & \multirow{2}{*}{85,964} & \multirow{2}{*}{68,145} \\ & & ECE & 3.7 & 7.4 & 9.1 & 14.4 & 15.1 & 12.6 & 9.6 & 11.7 & 13.3 & 8.4 & 7.0 & 16.8 & 8.5 & 8.4 & 8.5 & 10.29 & &\\
        \cmidrule{2-21}
        & \multirow{2}{*}{ROID}   & Acc. & 36.7 & 38.6 & 35.9 & 29.1 & 28.8 & 40.6 & 46.5 & 49.5 & 41.8 & 55.6 & 65.3 & 43.5 & 53.6 & 56.1 & 52.8 & 44.95 & \multirow{2}{*}{80,380} & \multirow{2}{*}{80,380} \\ & & ECE & 36.6 & 38.5 & 35.8 & 29.0 & 28.7 & 40.5 & 46.4 & 49.4 & 41.7 & 55.5 & 65.2 & 43.4 & 53.5 & 56.0 & 52.7 & 44.85 & &\\
        \cmidrule{2-21}
        & \multirow{2}{*}{RDump}   & Acc. & 35.6 & 34.4 & 36.1 & 32.7 & 34.6 & 45.5 & 51.6 & 50.3 & 44.1 & 59.9 & 66.8 & 46.3 & 56.8 & 59.1 & 54.7 & 47.23 & \multirow{2}{*}{50,000} & \multirow{2}{*}{26,375} \\ & & ECE & 10.5 & 11.4 & 12.6 & 13.9 & 14.4 & 9.4 & 8.1 & 8.1 & 9.4 & 6.3 & 5.5 & 11.8 & 6.1 & 7.0 & 6.3 & 9.40 & &\\
        \cmidrule{2-21}
        & \multirow{2}{*}{TEA}   & Acc. & 18.5 & 0.2 & 0.1 & 0.1 & 0.2 & 0.2 & 0.1 & 0.1 & 0.1 & 0.1 & 0.2 & 0.1 & 0.2 & 0.1 & 0.1 & 1.36 & \multirow{2}{*}{50,000$\times$23} & \multirow{2}{*}{50,000$\times$22} \\ & & ECE & 7.9 & 5.4 & 2.4 & 2.0 & 1.8 & 1.9 & 3.8 & 19.5 & 67.6 & 76.4 & 86.2 & 91.0 & 90.4 & 91.3 & 91.3 & 42.60 & &\\
        \cmidrule{2-21}
        & & Acc. & 35.6 & 38.7 & 37.5 & 35.9 & 36.1 & 47.6 & 53.1 & 50.6 & 45.6 & 59.5 & 67.1 & 45.2 & 57.6 & 59.8 & 55.4 & \underline{48.36} & & \\ &  \multirow{-2}{*}{EATA (Ours)} & ECE & 10.5 & 13.4 & 14.7 & 18.5 & 18.5 & 14.7 & 12.8 & 14.1 & 16.1 & 11.6 & 10.4 & 18.6 & 13.3 & 13.1 & 13.9 & 14.28 & \multirow{-2}{*}{50,000} & \multirow{-2}{*}{29,721}\\
        \cmidrule{2-21}
        &  & Acc. & 37.2 & 40.9 & 39.9 & 36.6 & 37.1 & 48.5 & 52.7 & 51.9 & 47.2 & 60.4 & 67.0 & 49.0 & 57.6 & 60.1 & 56.1 & \textbf{49.48} & &\\ &  \multirow{-2}{*}{EATA-C (Ours)} & ECE & 7.1 & 6.7 & 7.2 & 10.1 & 9.5 & 5.9 & 4.9 & 4.8 & 5.8 & 3.2 & 3.2 & 6.0 & 4.0 & 3.9 & 3.9 & \underline{5.74} & \multirow{-2}{*}{83,312} & \multirow{-2}{*}{33,312}\\
        \midrule
        \multirow{20}{*}[-4ex]{ViT (LN)} & \multirow{2}{*}{Source} & Acc. & 12.9 & 17.6 & 11.7 & 34.4 & 27.7 & 43.7 & 36.2 & 43.4 & 45.4 & 52.8 & 73.3 & 45.5 & 37.9 & 54.7 & 60.2 & 39.84  & \multirow{2}{*}{50,000} & \multirow{2}{*}{0}\\ & & ECE & 14.2 & 11.3 & 15.2 & 2.2 & 8.2 & 6.0 & 10.7 & 7.5 & 8.3 & 5.0 & 2.1 & 2.3 & 12.2 & 4.1 & 2.9 & \underline{7.48} & &\\
        \cmidrule{2-21}
        & \multirow{2}{*}{Tent$^\dagger$} & Acc. & 33.4 & 42.1 & 41.4 & 48.8 & 45.2 & 54.9 & 48.2 & 55.6 & 55.1 & 64.4 & 75.2 & 62.5 & 51.6 & 65.5 & 65.0 & 53.92 & \multirow{2}{*}{50,000} & \multirow{2}{*}{50,000} \\ & & ECE & 24.1 & 10.1 & 11.7 & 8.6 & 10.4 & 7.5 & 12.4 & 8.0 & 8.0 & 4.8 & 2.4 & 5.2 & 10.0 & 4.3 & 4.1 & 8.78 & &\\
        \cmidrule{2-21}
        & \multirow{2}{*}{MEMO*}   & Acc. & 32.2 & 35.1 & 32.6 & 37.5 & 28.5 & 43.3 & 40.1 & 45.2 & 47.0 & 53.9 & 73.3 & 53.3 & 39.6 & 59.5 & 62.8 & 45.60 & \multirow{2}{*}{50,000$\times$65} & \multirow{2}{*}{50,000$\times$64} \\ & & ECE & 33.0 & 32.5 & 33.2 & 36.7 & 45.7 & 41.0 & 47.2 & 39.8 & 38.8 & 33.7 & 20.0 & 26.7 & 48.5 & 30.4 & 27.2 & 35.63 & &\\
        \cmidrule{2-21}
        & \multirow{2}{*}{CoTTA}   & Acc. &45.6 & 58.8 & 58.4 & 40.1 & 50.9 & 50.4 & 44.4 & 46.1 & 51.4 & 52.2 & 57.7 & 36.5 & 53.9 & 55.8 & 55.4 & 50.51 & \multirow{2}{*}{50,000$\times$3} & \multirow{2}{*}{50,000}\\ & & ECE & 9.0 & 15.8 & 21.0 & 29.3 & 28.3 & 29.1 & 34.5 & 29.8 & 29.6 & 30.0 & 29.2 & 36.3 & 29.0 & 31.2 & 31.8 & 27.59 & &\\
        \cmidrule{2-21}
        & \multirow{2}{*}{SAR}   & Acc. & 43.1 & 50.8 & 52.9 & 50.4 & 51.0 & 57.5 & 53.2 & 58.6 & 61.2 & 66.0 & 76.1 & 61.2 & 54.7 & 67.8 & 67.7 & 58.15 & \multirow{2}{*}{91,605} & \multirow{2}{*}{82,277} \\ & & ECE & 8.0 & 8.4 & 9.5 & 9.1 & 11.4 & 9.6 & 13.0 & 9.9 & 9.6 & 7.8 & 4.5 & 10.0 & 13.8 & 7.5 & 7.5 & 9.30 & &\\
        \cmidrule{2-21}
        & \multirow{2}{*}{ROID}   & Acc. & 48.8 & 49.5 & 49.0 & 54.1 & 54.2 & 58.8 & 55.7 & 62.7 & 61.3 & 69.7 & 77.0 & 65.5 & 64.1 & 69.6 & 68.3 & 60.56 & \multirow{2}{*}{80,739} & \multirow{2}{*}{80,739}\\ & & ECE & 48.7 & 49.4 & 48.9 & 54.0 & 54.1 & 58.7 & 55.6 & 62.6 & 61.2 & 69.6 & 76.9 & 65.4 & 64.0 & 69.5 & 68.2 & 60.46 & &\\
        \cmidrule{2-21}
        & \multirow{2}{*}{RDump}   & Acc. & 50.5 & 48.4 & 51.8 & 54.0 & 55.3 & 59.4 & 55.9 & 63.3 & 60.7 & 70.8 & 76.8 & 66.8 & 60.9 & 69.5 & 68.4 & 60.84 & \multirow{2}{*}{50,000} & \multirow{2}{*}{32,050} \\ & & ECE & 10.6 & 9.7 & 10.5 & 8.9 & 9.4 & 7.9 & 10.0 & 6.9 & 7.4 & 5.4 & 3.2 & 6.7 & 7.6 & 5.1 & 5.0 & 7.62 & &\\
        \cmidrule{2-21}
        & \multirow{2}{*}{TEA}   & Acc. & 46.9 & 48.0 & 48.1 & 46.7 & 47.0 & 53.9 & 53.5 & 57.4 & 56.0 & 62.2 & 71.7 & 55.4 & 57.1 & 63.1 & 61.7 & 55.24 & \multirow{2}{*}{50,000$\times$23} & \multirow{2}{*}{50,000$\times$22} \\ & & ECE & 10.6 & 13.7 & 13.7 & 14.5 & 14.4 & 11.3 & 11.6 & 9.4 & 10.2 & 7.8 & 3.9 & 10.5 & 9.8 & 7.3 & 7.4 & 10.40 & &\\
        \cmidrule{2-21}
        & & Acc. & 50.5 & 55.6 & 56.0 & 54.9 & 56.3 & 61.1 & 59.8 & 64.3 & 64.0 & 70.1 & 77.4 & 65.5 & 63.1 & 70.4 & 69.5 & \underline{62.57}  &  & \\ & \multirow{-2}{*}{EATA (Ours)} & ECE & 10.6 & 13.9 & 15.7 & 16.4 & 17.3 & 15.9 & 17.5 & 14.8 & 15.7 & 12.8 & 9.0 & 15.4 & 17.5 & 13.3 & 13.9 & 14.63 & \multirow{-2}{*}{50,000} & \multirow{-2}{*}{36,688}\\
        \cmidrule{2-21}
        & & Acc. & 56.8 & 60.2 & 59.8 & 58.0 & 60.9 & 65.2 & 65.5 & 69.7 & 67.8 & 74.0 & 78.9 & 66.7 & 70.3 & 74.1 & 71.8 & \textbf{66.65} &  & \\ & \multirow{-2}{*}{EATA-C (Ours)} & ECE & 5.2 & 5.1 & 5.9 & 7.1 & 6.2 & 5.6 & 5.9 & 5.0 & 5.0 & 4.1 & 3.3 & 5.4 & 4.7 & 4.2 & 4.7 & \textbf{5.14} & \multirow{-2}{*}{83,184} & \multirow{-2}{*}{33,184} \\
    \bottomrule
        \end{tabular}}
         \end{threeparttable}
         \end{center}
     \vspace{-0.15in}
\end{table*}

\begin{table}[t]
    \caption{Comparison on ImageNet-R. Results are evaluated in the single-domain adaptation scenario. We use * to denote episodic adaptation.}
    \vspace{-0.1in}
        \label{tab:imagenet-r}
\newcommand{\tabincell}[2]{\begin{tabular}{@{}#1@{}}#2\end{tabular}}
 \begin{center}
 \begin{threeparttable}
    \resizebox{1.0\linewidth}{!}{
        \begin{tabular}{l|cccc}\toprule
        \multirow{1}{*}{\tabincell{c}{Model}}&\multirow{1}{*}{\tabincell{c}{Acc. (\%)}}&\multirow{1}{*}{\tabincell{c}{ECE (\%)}}&\multirow{1}{*}{\tabincell{c}{\#Forwards}}&\multirow{1}{*}{\tabincell{c}{\#Backwards}} \\
        
        \midrule
         ResNet-50(BN) & 38.0 & 17.7 & 30,000 & 0     \\
         ~~$\bullet~$BN~\cite{schneider2020improving}   & 40.4$_{(+2.4)}$ & 13.4$_{(-4.3)}$ & 30,000 & 0      \\
         ~~$\bullet~$Tent~\cite{wang2021tent} & 42.3$_{(+4.3)}$ & 17.8$_{(+0.1)}$ & 30,000 & 30,000      \\
         ~~$\bullet~$MEMO*~\cite{zhang2021memo} &       41.9$_{(+3.9)}$ & 26.9$_{(+9.2)}$   & 30,000$\times$65  & 30,000$\times$64   \\
         ~~$\bullet~$CoTTA~\cite{wang2022cotta} & 42.4$_{(+4.4)}$ & 15.8$_{(-1.9)}$ & 90,000 & 30,000  \\
         ~~$\bullet~$SAR~\cite{niu2023sar} & 42.7$_{(+4.7)}$ & 14.6$_{(-3.1)}$ & 47,755 & 32,877 \\
         ~~$\bullet~$ROID~\cite{marsden2024universal} & \textbf{48.6$_{(+10.5)}$} & 48.1$_{(+30.4)}$ & 48,303 & 48,303 \\
         ~~$\bullet~$TEA~\cite{yuan2024tea} & 42.8$_{(+4.8)}$ & 14.2$_{(-3.5)}$ & 30,000$\times$23 & 30,000$\times$22 \\
         ~~$\bullet~$EATA (Ours) & 44.9$_{(+6.9)}$ & 16.7$_{(-1.0)}$ & 30,000 & 5,417 \\
         ~~$\bullet~$EATA-C (Ours) & 47.1$_{(+9.1)}$ & \textbf{13.3$_{(-4.4)}$} & 35,122 & 5,122 \\
         \midrule
         ViT(LN) & 52.5 & 5.0 & 30,000 & 0     \\
         ~~$\bullet~$Tent~\cite{wang2021tent} & 54.2$_{(+1.7)}$ & 7.4$_{(+2.4)}$ & 30,000 & 30,000   \\
         ~~$\bullet~$MEMO*~\cite{zhang2021memo} & 57.5$_{(+5.0)}$ & 32.1$_{(+27.1)}$ & 30,000$\times$65  & 30,000$\times$64  \\
         ~~$\bullet~$CoTTA~\cite{wang2022cotta} & 56.4$_{(+3.9)}$ & 7.4$_{(+2.4)}$ & 90,000 & 30,000  \\
         ~~$\bullet~$SAR~\cite{niu2023sar} & 55.0$_{(+2.5)}$ & 5.2$_{(+0.2)}$ & 47,119 & 33,844 \\
         ~~$\bullet~$ROID~\cite{marsden2024universal} & 62.2$_{(+9.7)}$ & 61.7$_{(+56.7)}$ & 49,795 & 49,795 \\
         ~~$\bullet~$TEA~\cite{yuan2024tea} & 60.1$_{(+7.6)}$ & 7.4$_{(+2.4)}$ & 30,000$\times$23 & 30,000$\times$22\\
         ~~$\bullet~$EATA (Ours)& 58.2$_{(+5.7)}$ & 5.8$_{(+0.8)}$ & 30,000 & 6,053 \\
         ~~$\bullet~$EATA-C (Ours)& \textbf{64.2$_{(+11.7)}$} & \textbf{3.9$_{(-1.1)}$} & 36,395 & 6,395 \\
        \bottomrule
        \end{tabular}}
         \end{threeparttable}
         \end{center}
         \vspace{-0.2in}
\end{table}

\noindent\textbf{Datasets and Models.} 
We conduct experiments on three benchmark datasets for OOD generalization: ImageNet-C~\cite{hendrycks2019benchmarking} (contains corrupted images in 15 types of 4 main categories and each type has 5 severity levels) and ImageNet-R~\cite{hendrycks2021many} for image classification; and ACDC~\cite{SDV21acdc} for semantic segmentation.
We use ResNet-50 (R-50)~\cite{he2016deep} and ViT-Base (ViT)~\cite{dosovitskiy2020vit} for ImageNet experiments, and Segformer-B5~\cite{xie2021segformer} for ACDC~\cite{SDV21acdc} experiments. The models are trained on ImageNet or CityScapes~\cite{cordts2016cityscapes} training set with stochastic depth regularization~\cite{huang2016depth} and tested on clean or OOD test sets.

\noindent\textbf{Compared Methods.}
We compare with the following state-of-the-art methods. BN adaptation~\cite{schneider2020improving} updates batch normalization statistics on test samples. Tent~\cite{wang2021tent} minimizes the entropy of test samples during testing. MEMO~\cite{zhang2021memo} maximizes the prediction consistency of different augmented copies regarding a given test sample. SAR~\cite{niu2023sar} selects reliable samples for test time sharpness-aware entropy minimization. CoTTA~\cite{wang2022cotta} and DAT~\cite{ni2024distribution} minimize the cross entropy between the student network and its mean teacher during testing. RDump~\cite{press2024rdumb} periodically resets model parameters based on our \myeata. TEA~\cite{yuan2024tea} employs energy-based contrastive learning with negative sample generation. ROID~\cite{marsden2024universal} minimizes the diversity-weighted soft likelihood ratio loss. We denote \myeata without weight regularization in Eqn.~(\ref{eq:weight_regularization}) as \textbf{e}fficient \textbf{t}est-time \textbf{a}daptation \textbf{(\myeta)}.
More ablative methods can be found in Table~\ref{tab:eata-c_ablation}.

\noindent\textbf{Adaptation Scenarios.} We conduct experiments under three adaptation scenarios: 1) \textit{Episodic}, the model parameters will be reset immediately after each optimization step of a test sample or batch; 2) \textit{Single-domain}, the model is online adapted through the entire evaluation of one given test dataset (\eg, gaussian noise level 5 of ImageNet-C). Once the adaptation on this dataset is finished, the model parameters will be reset; 3) \textit{Lifelong}, the model is online adapted and the parameters will never be reset (as shown in Figure~\ref{fig:imageC-forgetting-level-5} (\textbf{Right})), which is more challenging but practical.

\noindent\textbf{Evaluation Metrics.}
1) Clean accuracy/error (\%) on original in-distribution (ID) test samples, \eg, the original test images of ImageNet. Note that we measure the clean accuracy of all methods via \textbf{(re)adaptation}; 2) Out-of-distribution (OOD) accuracy/error (\%) on OOD test samples, \eg, the corruption accuracy on ImageNet-C; 3) Expected Calibration Error (ECE)~\cite{naeini2015ece} on OOD test samples, which measures the average discrepancies between model's confidence and accuracy within multiple confidence intervals; 4) The number of forward and backward passes during the entire TTA process.
Note that the fewer \#forwards and \#backwards indicate less computation, leading to higher efficiency.

\begin{table*}[t]
        \caption{
        Semantic segmentation results (mIoU in \%) on the Cityscapes-to-ACDC lifelong test-time adaptation scenario. The model is continually adapted to the four adverse conditions for ten rounds without model reset. All results are evaluated based on the Segformer-B5 architecture. Following~\cite{wang2022cotta}, we only show the results of the first, fourth, seventh, and last rounds due to page limits. Full results can be found in the supplementary material.}
 \vspace{-0.1in}
        \label{tab:acdc}
\newcommand{\tabincell}[2]{\begin{tabular}{@{}#1@{}}#2\end{tabular}}
 \begin{center}
 \begin{threeparttable}
    \resizebox{1.0\linewidth}{!}{
        \begin{tabular}{l|cccc|cccc|cccc|cccc|l}
        \toprule
        Round & \multicolumn{4}{|l}{1} & \multicolumn{4}{|l}{4} & \multicolumn{4}{|l}{7} & \multicolumn{4}{|l|}{10} & All \\
        \midrule
        Condition & Fog & Night & Rain & Snow & Fog & Night & Rain & Snow & Fog & Night & Rain & Snow & Fog & Night & Rain & Snow & Mean \\
        \midrule
        Source & 69.1 & 40.3 & 59.7 & 57.8 & 69.1 & 40.3 & 59.7 & 57.8 & 69.1 & 40.3 & 59.7 & 57.8 & 69.1 & 40.3 & 59.7 & 57.8 & 56.7 \\
        BN Stats Adapt & 62.3 & 38.0 & 54.6 & 53.0 & 62.3 & 38.0 & 54.6 & 53.0 & 62.3 & 38.0 & 54.6 & 53.0 & 62.3 & 38.0 & 54.6 & 53.0 & 52.0$_{(-4.7)}$ \\
        Tent (lifelong) & 69.0 & 40.2 & 60.0 & 57.3 & 66.6 & 36.6 & 58.9 & 54.2 & 64.6 & 33.4 & 55.9 & 51.6 & 62.5 & 30.4 & 52.6 & 48.7 & 52.7$_{(-4.0)}$ \\
        CoTTA & 70.9 & 41.1 & 62.4 & 59.7 & 70.8 & 40.6 & 62.6 & 59.7 & 70.8 & 40.5 & 62.6 & 59.7 & 70.8 & 40.5 & 62.6 & 59.7 & 58.4$_{(+1.7)}$ \\
        DAT & 71.7 & 44.6 & 63.8 & 62.2 & 68.0 & 42.0 & 60.9 & 59.4 & 66.1 & 40.6 & 59.8 & 57.8 & 63.8 & 39.6 & 58.2 & 55.4 & 57.0$_{(+0.3)}$ \\
        \midrule
        \myeata & 69.1 & 40.5 & 59.8 & 58.1 & 69.3 & 41.8 & 60.1 & 58.6 & 68.8 & 42.5 & 59.4 & 57.9 & 67.9 & 42.8 & 57.7 & 56.3 & 57.0$_{(+0.3)}$ \\ 
        \myeataC & 71.0 & 44.3 & 63.1 & 61.1 & 72.0 & 47.3 & 64.9 & 63.8 & 71.8 & 48.2 & 64.2 & 64.2 & 72.0 & 48.7 & 64.3 & 64.1 & \textbf{61.6$_{(+4.9)}$} \\
        \bottomrule
        \end{tabular}}
         \end{threeparttable}
         \end{center}
  \vspace{-0.1in}
\end{table*}

\noindent\textbf{Implementation Details.}
For test time adaptation, we use SGD as the update rule, with a momentum of 0.9 and a batch size of 64. 
In \myeata and \myeta, the learning rate is set to 0.00025/0.001 for ResNet-50/ViT-Base on ImageNet, and 7.5$\times10^{-5}$ on ACDC, respectively (following Tent, SAR and CoTTA). In \myeataC, the learning rate is set to 0.005/0.1 for ResNet-50/ViT-Base on ImageNet, and 0.0005 on ACDC, respectively. The sub-network is obtained via stochastic depth regularization~\cite{huang2016depth} with a drop ratio of 0.2/0.6 for ImageNet/ACDC. For both \myeata and \myeataC, we use 2,000/20 samples for ImageNet/ACDC to calculate $\omega(\theta_i)$ in Eqn.~(\ref{eq:fisher_information}).The
effect and sensitivity of each hyperparameter is investigated in Section~\ref{sec:ablation}. More details are put in Supplementary.

\subsection{Comparisons \wrt OOD Performance, Efficiency and Calibration Error}

\textbf{Results on ImageNet-C.}
From Table~\ref{tab:imagenet-c-level-5}, our \myeata and \myeataC consistently surpass existing approaches regarding adaptation accuracy, \eg, the average accuracy of 48.4\% (EATA) \textit{vs.} 45.0\% (ROID) on ResNet-50. Importantly, \myeata yields a remarkable performance gain over its counterpart Tent, \eg, $33.4\%\rightarrow50.5\%$ on Gaussian Noise with ViT-Base, suggesting the significance of removing samples with unreliable gradients and tackling samples differently in the TTA process. Our enhanced method, EATA-C, further boosts adaptation accuracy by a large margin, which consistently outperforms TEA and ROID in all 15 corruption types over both ResNet-50 and ViT-Base, suggesting our effectiveness.
Note that besides achieving strong OOD performance, \myeata also alleviates the forgetting on ID samples (see Figure~\ref{fig:imageC-forgetting-level-5}), showing the effectiveness of our anti-forgetting regularization without limiting the learning ability during adaptation (see also Table~\ref{tab:eata-c_ablation} for ablation).

In terms of computational efficiency, \myeata requires only 29,721 backward passes on ResNet-50, which is much fewer than methods that require extensive data augmentations (\ie, TEA at 50,000$\times$22) or multiple optimization iterations (\eg, SAR at 68,145 on ResNet-50). Compared with Tent (\eg, 50,000 backward passes), \myeata saves computation by excluding samples with high prediction entropy and redundant samples out of test-time optimization, resulting in higher efficiency. While our \myeataC\linebreak uses additional forward passes, its forward passes with the full network are gradient-free, and the lightweight sub-network forward/backward passes are conducted only on the selected samples, maintaining overall computational efficiency comparable to \myeata~(see Table~\ref{tab:efficiency_comparison} for detailed results and discussions on wall-clock time and memory usage). Although optimization-free methods (such as BN adaptation) do not need backward updates, their applicability scope and OOD performances are limited.

Regarding calibration, existing methods consistently exhibit substantial calibration error (\eg, ROID and CoTTA are 60.46\% and 27.59\% on ECE with ViT-Base), suggesting miscalibration as a prevalent issue in the unsupervised test-time adaptation. By filtering unreliable samples to reduce noisy learning signals, \myeata improves calibration over Tent (\eg, $11.7\%\rightarrow10.5\%$ on Gaussian Noise with ResNet-50), though miscalibration is yet significant. By further decreasing reducible model uncertainty and reflecting data uncertainty in model predictions, our enhanced method, \myeataC, reduces the ECE of \myeata by relatively 59.8\% on ResNet-50 and 64.9\% on ViT-Base, demonstrating the strong calibration effect of \myeataC across diverse datasets and architectures. In summary, while \myeataC improves performance and efficiency over the state of the art, our \myeataC further achieves high accuracy, well-calibrated prediction, and efficient computation simultaneously, 
establishing a new benchmark for test-time adaptation.

\noindent\textbf{Results on ImageNet-R.}
From Table~\ref{tab:imagenet-r}, \myeata consistently achieves a favorable balance between performance and efficiency, significantly improving accuracy on both ResNet-50 and ViT-Base while requiring much fewer backpropagation steps. For instance, \myeata improves accuracy from 42.8\% (TEA) to 44.9\% on ResNet-50, while reducing the backpropagation steps from 30,000$\times$22 to 5,417. EATA-C further improves accuracy substantially (\eg, by 6.0\% over \myeata on ViT-Base), while maintaining computational efficiency comparable to \myeata. Importantly, \myeataC is the only method that reduces calibration error on both ResNet-50 and ViT-Base and uniquely lowers ECE on ViT-Base, suggesting the effectiveness of our calibration-driven objective in TTA.

\noindent\textbf{Results on CityScapes-to-ACDC.} Following~\cite{wang2022cotta}, we evaluate our method on the semantic segmentation task in a lifelong adaptation scenario. From Table~\ref{tab:acdc}, while DAT initially achieves higher mIOU, it tends to overfit, leading to significant performance degradation in subsequent adaptations. In contrast, our \myeata maintains a more stable performance compared to DAT and Tent, by filtering unreliable predictions and preventing drastic changes to important model parameters. Moreover, by replacing entropy minimization with our consistency maximization objective for more robust learning signals, \myeataC achieves state-of-the-art performance, surpassing \myeata by 4.6\% and CoTTA by 3.2\% in mIoU over ten adaptation rounds. More critically, our EATA-C showcases consistent improvement over lifelong adaptation, increasing the average mIOU on four datasets from 59.8\% (first round) to 62.3\% (tenth round), further highlighting our long-term effectiveness.

\subsection{Demonstration of Preventing Forgetting}

\begin{figure*}[t]
\centering
\subfigure{\label{fig:imageC-forgetting-level-5-each-reset}\includegraphics[width=73mm]{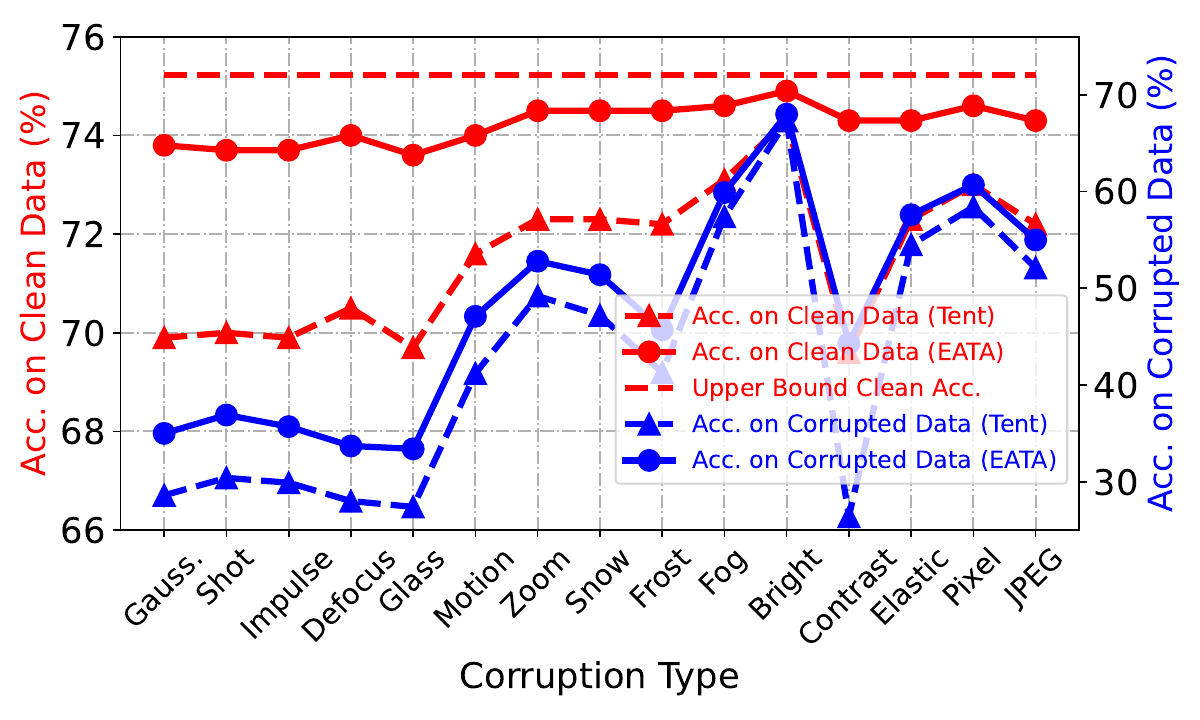}}
\subfigure{\label{fig:imageC-forgetting-level-5-lifelong}\includegraphics[width=72mm]{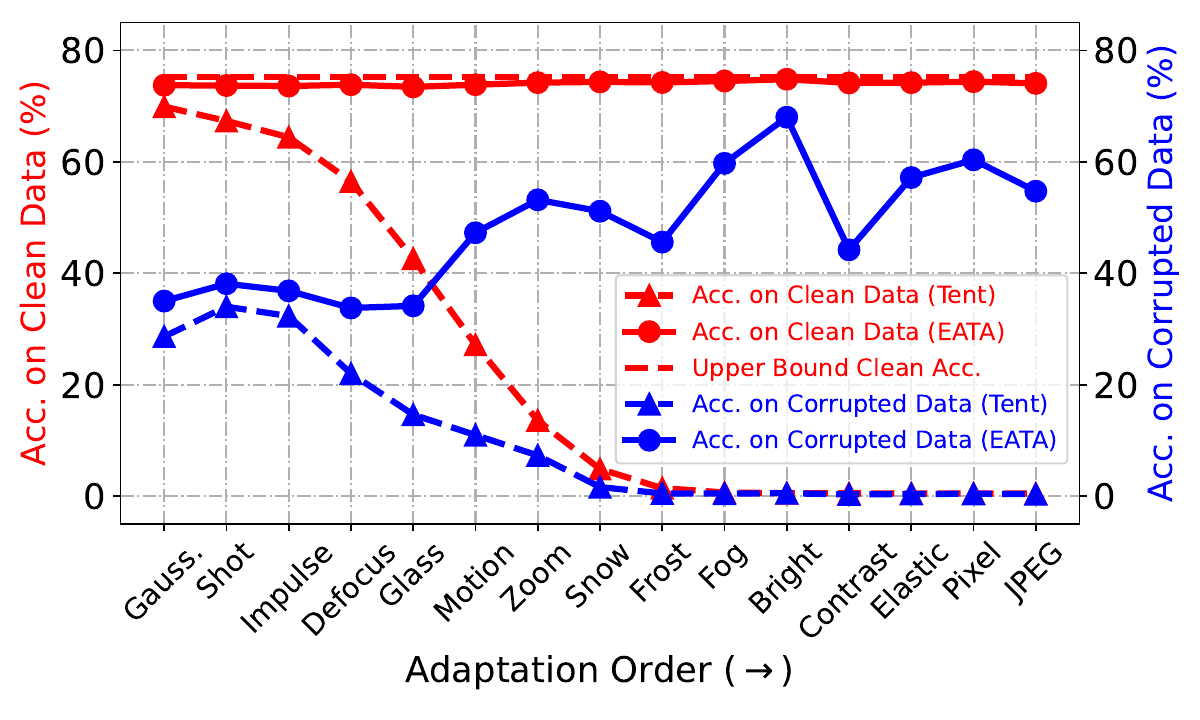}}
\vspace{-0.15in}
\caption{Comparison of preventing forgetting on ImageNet-C (severity level 5) with ResNet-50. We record the OOD corruption accuracy on each corrupted test set and the ID clean accuracy (after OOD adaptation). In \textbf{Left}, the model parameters of both Tent and our \myeata are reset before adapting to a new corruption type. In \textbf{Right}, the model performs lifelong adaptation and the parameters will never be reset, namely Tent (lifelong) and our \myeata (lifelong). The upper bound clean accuracy is estimated with the source model without adaptation on corrupted OOD data, which does not suffer from forgetting. \myeata achieves higher OOD accuracy and meanwhile maintains the ID clean accuracy.}
\vspace{-0.05in}
\label{fig:imageC-forgetting-level-5}
\end{figure*}

In this section, we investigate the ability of our \myeata in preventing ID forgetting during test-time adaptation. The experiments are conducted on ImageNet-C with ResNet-50. We measure the anti-forgetting ability by comparing the model's clean accuracy (\ie, on original validation data of ImageNet) before and after adaptation. To this end, we first perform test-time adaptation on a given OOD test set, and then evaluate the clean accuracy of the updated model.
Here, we consider two adaptation scenarios: the single-domain adaptation, and the lifelong adaptation.
We report the results of severity level 5 in Figure~\ref{fig:imageC-forgetting-level-5} and put the results of severity levels 1-4 into Supplementary.

From Figure~\ref{fig:imageC-forgetting-level-5}, our \myeata consistently outperforms Tent regarding the OOD corruption accuracy and meanwhile maintains the clean accuracy on ID data (in both two adaptation scenarios), demonstrating our effectiveness. It is worth noting that the performance degradation in lifelong adaptation scenario is much more severe (see Figure~\ref{fig:imageC-forgetting-level-5} \textbf{Right}). 
More critically, in lifelong adaptation, both the clean and corruption accuracy of Tent decreases rapidly (until degrades to 0\%) after adaptation of the first three corruption types, showing that Tent in lifelong adaptation is not stable enough. In contrast, during the whole lifelong adaptation process, our \myeata achieves good corruption accuracy and the clean accuracy is also very close to the clean accuracy of the model without any OOD adaptation (\ie, original clean accuracy, tested using Tent). These results demonstrate the superiority of our anti-forgetting Fisher regularizer in terms of overcoming the forgetting on ID data.

\subsection{Ablation Studies}
\label{sec:ablation}

\noindent\textbf{Effect of Components in $S(\bx)$ (Eqn.~\ref{eq:lambda_2nd}).}
Our \myeata accelerates test-time adaptation by excluding two types of samples out of optimization: 1) samples with high prediction entropy values (Eqn.~\ref{eq:lambda_1st}) and 2) samples that are similar (Eqn.~\ref{eq:lambda_2nd}). We ablate both of them in Table~\ref{tab:effects_diff_components}. Compared with the baseline $S(\bx)\small{=}1$ (the same as Tent), introducing $S^{ent}(\bx)$ in Eqn.~(\ref{eq:lambda_1st}) achieves better accuracy and fewer backwards (\eg, 49.6\% (37,636) \vs~33.4\% (50,000) on level 5). This verifies our motivation in Figure~\ref{fig:selective_entropy_motivation} that some high-entropy samples may hurt the performance since their gradients are unreliable.
When further removing some redundant samples that are similar (Eqn.~\ref{eq:lambda_2nd}), our \myeata further reduces the number of back-propagation (\eg, 37,636$\rightarrow$28,168 on level 5) and achieves comparable OOD error (\eg, 50.4\% \vs~49.6\%), demonstrating the effectiveness of our sample-efficient optimization strategy.

\noindent\textbf{Effect of Components in \myeataC.}
Our EATA-C aims to achieve a favorable balance between accuracy, calibration, and efficiency. We conduct an ablation study to verify the effectiveness of each module as in Table~\ref{tab:eata-c_ablation}. The results indicate the following findings: 1)~\textit{Consistency Loss}: Incorporating the consistency loss alone substantially enhances the source model's robustness and reduces ECE; 2)~\textit{Entropy Regularization}: The min-max entropy regularizer further calibrates prediction confidence and leads to a slight improvement in accuracy, \eg, accuracy increases from 48.8\% (Exp 10) to 49.0\% (EATA-C), and ECE decreases from 5.1\% to 4.6\%; 3)~\textit{Fisher Regularization}: This anti-forgetting regularizer contributes to TTA stability, as in the lifelong TTA of Table~\ref{tab:imagenet-c-level-5} and Figure~\ref{fig:imageC-forgetting-level-5}. In single-domain TTA, it also positively affects both ECE and accuracy, \eg, ECE decreases from 5.4\% (ETA-C) to 4.6\% (\myeataC) and accuracy improves from 48.9\% to 49.0\%; 4)~\textit{Active Sample Selection}: By filtering out unreliable and redundant test samples, active sample selection significantly boosts computational efficiency while maintaining or improving accuracy, \eg, accuracy increases from 48.5\% (Exp 6) to 49.0\% (EATA-C) while reducing the required backward passes by 35\%. More discussions on wall-clock time and memory usage are provided in Table~\ref{tab:efficiency_comparison}. These results collectively underscore the effectiveness of each component.

\begin{figure*}[t]
    \centering
    \begin{minipage}[t]{0.32\linewidth}
        \centering
        \includegraphics[width=1.0\linewidth]{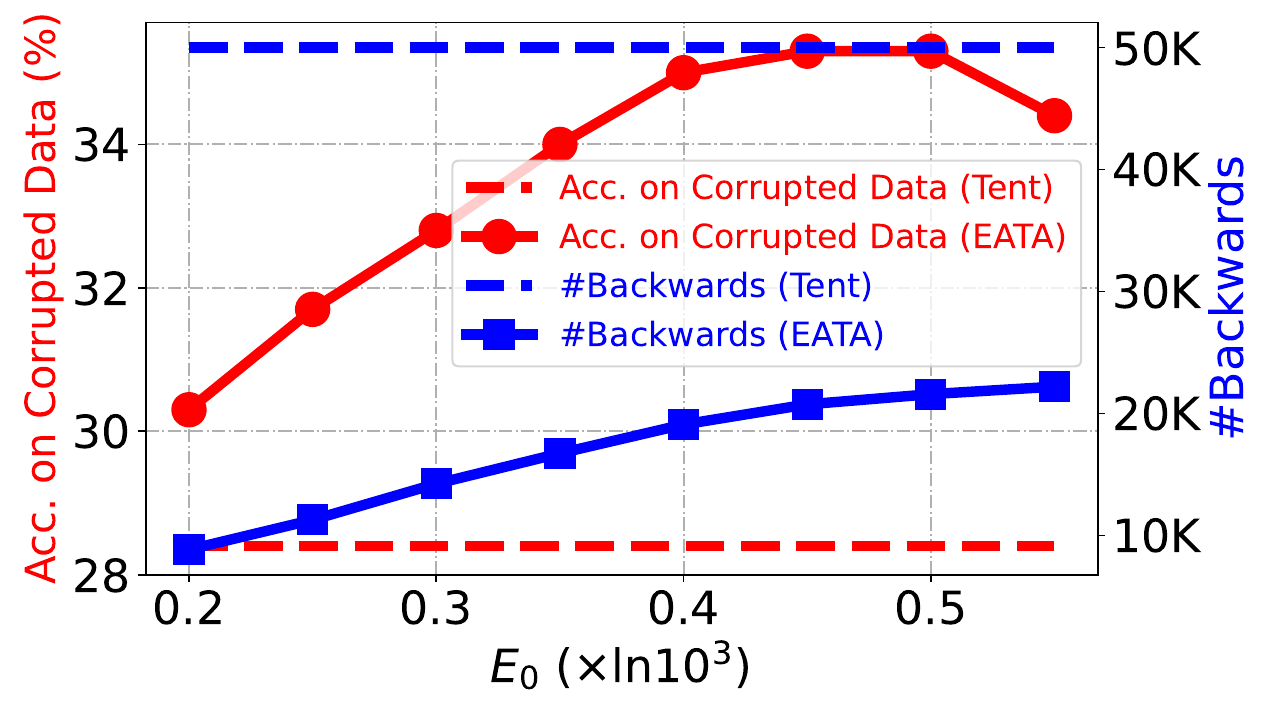}
        \vspace{-0.3in}
        \caption{Effect of different entropy margins $E_0$ in Eqn.~(\ref{eq:lambda_1st}). Results obtained on ImageNet-C(Gaussian, level 5) with ResNet-50.}
        \label{fig:entropy0_effects}
    \end{minipage}
    ~
    \begin{minipage}[t]{0.31\linewidth}
        \centering
        \includegraphics[width=1.0\linewidth]{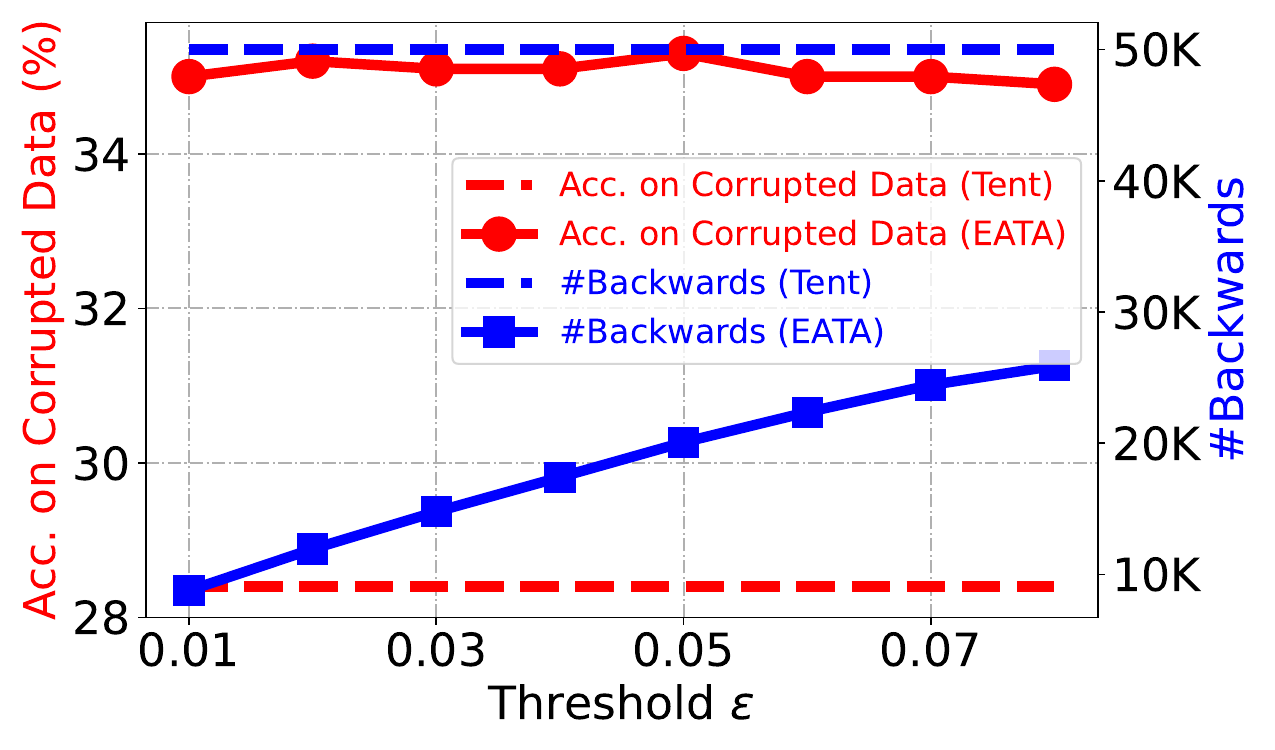}
        \vspace{-0.3in}
        \caption{Effect of different similarity threshold $\epsilon$ in Eqn.(~\ref{eq:lambda_div}). Results obtained on ImageNet-C(Gaussian, level 5) with ResNet-50.}
        \label{fig:d_influence}
    \end{minipage}
    ~
    \begin{minipage}[t]{0.32\linewidth}
        \centering
        \includegraphics[width=1.0\linewidth]{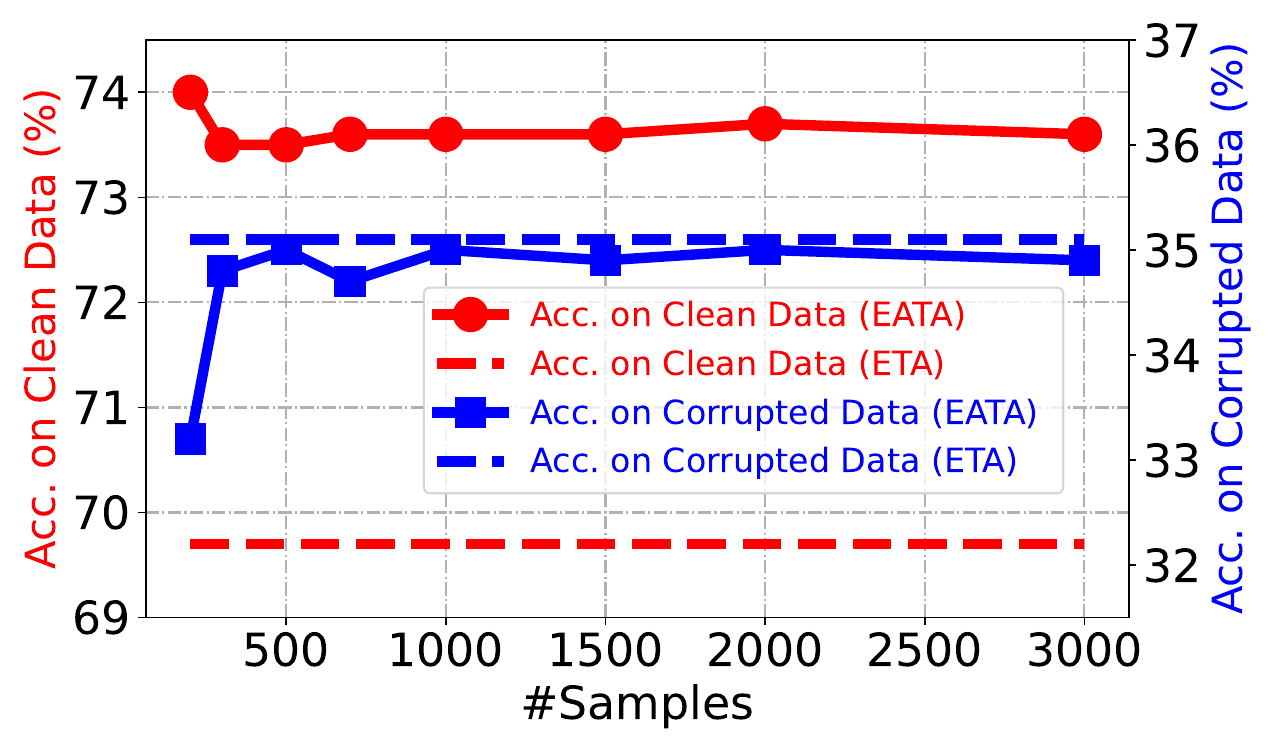}
        \vspace{-0.3in}
        \caption{Effect of \#samples for calculating Fisher in Eqn.~(\ref{eq:fisher_information}). Results obtained on ImageNet-C(Gaussian, level 5) with ResNet-50.}
        \label{fig:number_samples_fisher}
    \end{minipage}
\end{figure*}

\begin{figure*}[t]
    \centering
    \begin{minipage}[t]{0.31\linewidth}
        \centering
        \includegraphics[width=1.0\linewidth]{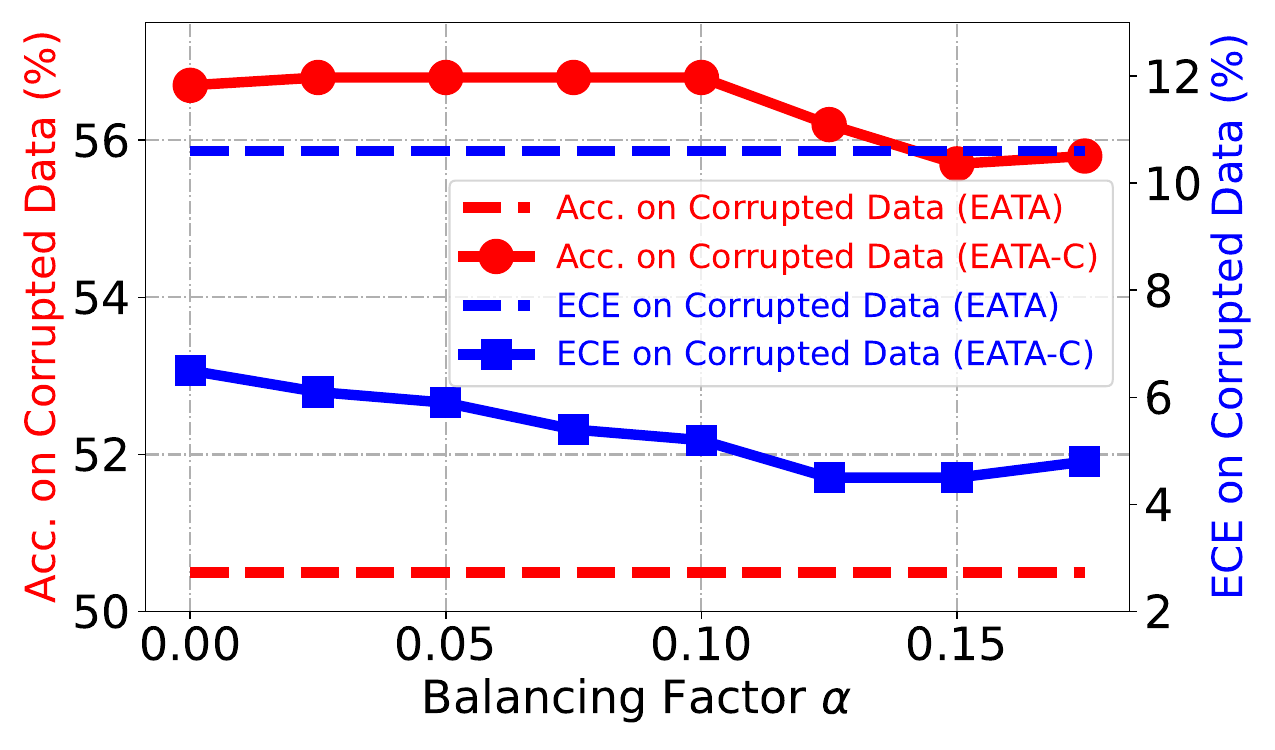}
        \vspace{-0.3in}
        \caption{Effect of different balancing factor $\alpha$ in Eqn.~(\ref{eq:overall_loss}). Results obtained on ImageNet-C(Gaussian, level 5) with ViT-Base.}
        \label{fig:alpha_influence}
    \end{minipage}
    ~
    \begin{minipage}[t]{0.31\linewidth}
        \centering
        \includegraphics[width=1.0\linewidth]{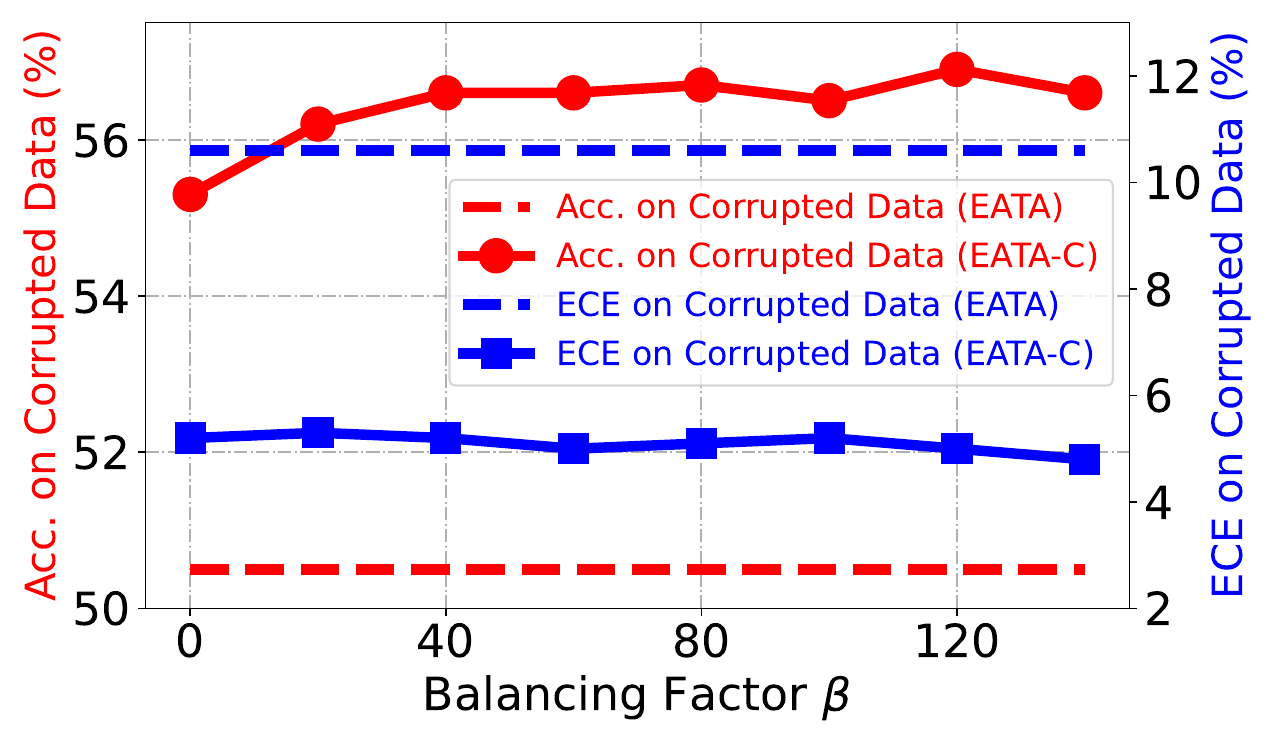}
        \vspace{-0.3in}
        \caption{Effect of different balancing factor $\beta$ in Eqn.~(\ref{eq:overall_loss}). Results obtained on ImageNet-C(Gaussian, level 5) with ViT-Base.}
        \label{fig:beta_influence}
    \end{minipage}
    ~
    \begin{minipage}[t]{0.31\linewidth}
        \centering
        \includegraphics[width=1.0\linewidth]{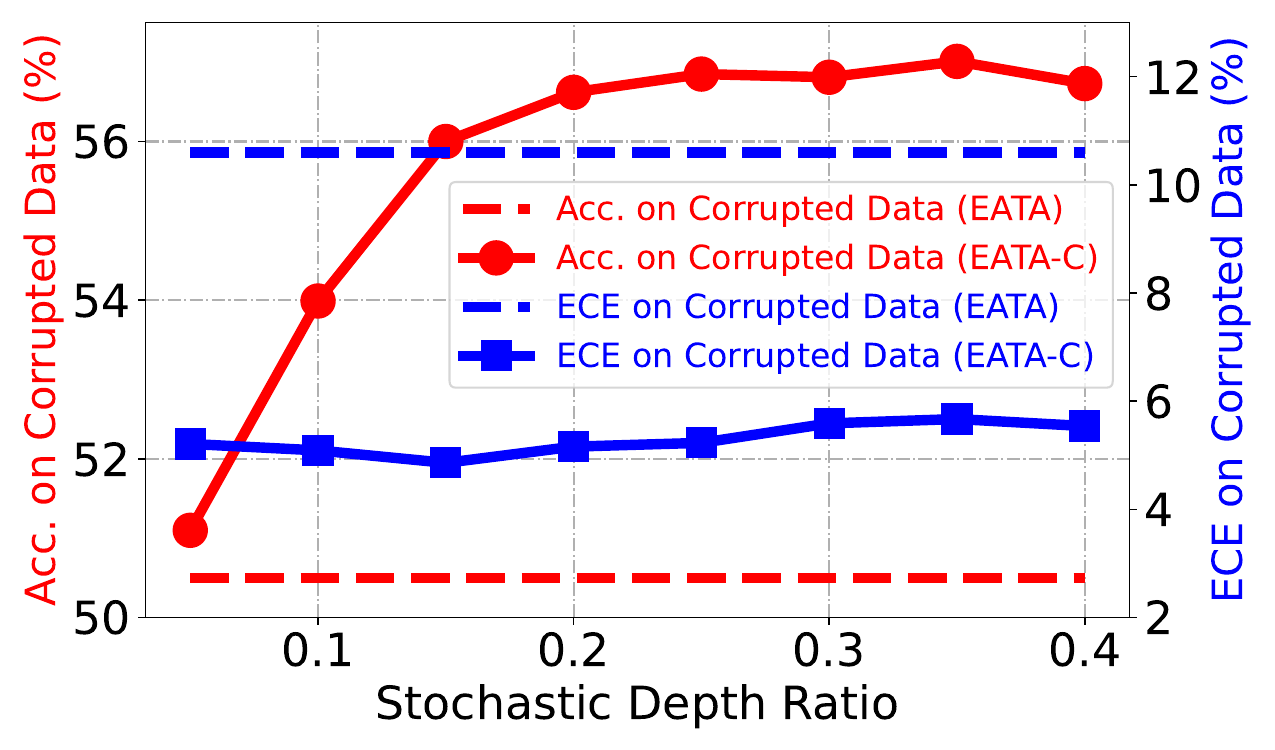}
        \vspace{-0.3in}
        \caption{Effect of different stochastic depth ratios in \myeataC. Results obtained on ImageNet-C(Gaussian, level 5) with ViT-Base.}
        \label{fig:stochastic_depth_influence}
    \end{minipage}
\end{figure*}

\begin{table}[t]
\vspace{-0.1in}
	\caption{
	Effectiveness of components in sample-adaptive weight $S(\bx)$ in \myeata on ImageNet-C (Gaussian noise) with ResNet-50.}
 \vspace{-0.15in}
	\label{tab:effects_diff_components}
\newcommand{\tabincell}[2]{\begin{tabular}{@{}#1@{}}#2\end{tabular}}
 \begin{center}
 \begin{threeparttable}
    \resizebox{1.0\linewidth}{!}{
 	\begin{tabular}{l|cccc}\toprule
        \multicolumn{1}{c|}{\multirow{2}[0]{*}{Method}} & \multicolumn{2}{c}{Level 3} & \multicolumn{2}{c}{Level 5} \\
         & \multicolumn{1}{c}{Acc. (\%)} & \multicolumn{1}{c}{\#Backwards} & \multicolumn{1}{c}{Acc. (\%)} & \multicolumn{1}{c}{\#Backwards} \\
        \midrule
         Baseline ($S(\bx)\small{=}1$) & 68.8  & 50,000	& 33.4  & 50,000      \\
         ~~$+S^{ent}{(\bx)}$ (Eqn.~\ref{eq:lambda_1st}) & 70.7 & 45,302	     & 49.6 & 37,636        \\
         ~~$+S{(\bx)}$ (Eqn.~\ref{eq:lambda_2nd})  &\textbf{70.8} &\textbf{36,057}       & \textbf{50.4} & \textbf{28,168}                 \\
        \bottomrule
	\end{tabular}}
	 \end{threeparttable}
	 \end{center}
	
\vspace{-0.1in}
\end{table}

\noindent\textbf{Entropy Constant $E_0$ in Eqn.~(\ref{eq:lambda_1st}).}
We evaluate our \myeata with different $E_0$, selected from \{0.20, 0.25, 0.30, 0.35, 0.40, 0.45, 0.50, 0.55\}$\times\ln 10^3$, where $10^3$ is the class number of ImageNet. From Figure~\ref{fig:entropy0_effects}, our \myeata achieves excellent performance when $E_0$ belongs to $[0.4, 0.5]$. Either a smaller or larger $E_0$ would hamper the performance. The reasons are mainly as follows. When $E_0$ is small, \myeata removes too many samples during adaptation and thus is unable to learn enough knowledge from the remaining samples. When $E_0$ is too large, some high-entropy samples would take part but contribute unreliable and harmful gradients, resulting in performance degradation. As larger $E_0$ leads to more backward passes, we set $E_0$ to 0.4$\times\ln 10^3$ for efficiency-performance trade-off and fix the proportion of 0.4 for all other ImageNet experiments.

\begin{table}[t]
    \caption{Effects of components in \myeataC. Results obtained on 15 corruptions of ImageNet-C (level 5) with ResNet-50 in single-domain TTA scenario, \ie, \textbf{the model parameters are reset before adapting to a new domain}. \textbf{CL} denotes consistency loss. \textbf{ER} denotes min-max entropy regularizer. \textbf{FR} denotes Fisher regularizer. \textbf{SS} denotes active sample selection.}
    \vspace{-0.15in}
    \label{tab:eata-c_ablation}
\newcommand{\tabincell}[2]{\begin{tabular}{@{}#1@{}}#2\end{tabular}}
 \begin{center}
 \begin{threeparttable}
    \resizebox{1.0\linewidth}{!}{
 	\begin{tabular}{l|cccc|cccc}
 	\toprule
 	  \multirow{2}{*}{Experiment} & \multirow{2}{*}{CL} & \multirow{2}{*}{ER} & \multirow{2}{*}{FR} &  \multirow{2}{*}{SS} & \multicolumn{4}{c}{Average}  \\
        ~ & ~ & ~ & ~ & ~ & Acc. & ECE & \#Forwards & \#Backwards \\
        \midrule
        Source & & & & & 39.8 & 7.5 & 50,000 & 0 \\
        1 & \checkmark & & & & 47.7 & 4.3 & 50,000$\times$2 & 50,000\\
        2 & & \checkmark & & & 44.0 & 5.2 & 50,000$\times$2 & 50,000\\
        3 & \checkmark & \checkmark & & & 48.6 & 4.3 & 50,000$\times$2 & 50,000 \\
        \midrule
        4 & \checkmark & & \checkmark & &  47.5 & 3.7 & 50,000$\times$2 & 50,000\\
        5 & & \checkmark & \checkmark & & 43.4 & 4.2 & 50,000$\times$2 & 50,000\\
        6 & \checkmark & \checkmark & \checkmark & & 48.5 & 3.8 & 50,000$\times$2 & 50,000\\
        \midrule
        7 & \checkmark & & & \checkmark & 48.9 & 5.9 & 83,583 & 33,583\\
        8 & & \checkmark & & \checkmark & 42.9 & 4.4 & 74,705 & 24,705\\
        9 (ETA-C) & \checkmark & \checkmark & & \checkmark & 48.9 & 5.4 & 82,882 & 32,882 \\
        10 & \checkmark & & \checkmark & \checkmark & 48.8 & 5.1 & 82,952 & 32,952\\
        11 & & \checkmark & \checkmark & \checkmark & 42.3 & 3.8 & 74,226 & 24,226\\
        12 (EATA-C) & \checkmark & \checkmark & \checkmark & \checkmark & 49.0 & 4.6 & 82,492 & 32,492\\
        \bottomrule
    	\end{tabular}
	}
	 \end{threeparttable}
	 \end{center}
  \vspace{-0.1in}
\end{table}

\noindent \textbf{Similarity Threshold $\epsilon$ in Eqn.~\ref{eq:lambda_div}.} We use $\epsilon$ to select diverse samples for TTA. From Figure~\ref{fig:d_influence}, \myeata maintains stable accuracy across a wide range of $\epsilon \in [0.01, 0.08]$, showcasing insensitivity, while a smaller $\epsilon$ removes significantly more samples and improves computational efficiency. We set $\epsilon{=}0.05$ without careful tuning.
More results on efficiency (\ie, time and memory usage) of \myeata and \myeataC with varying $\epsilon$ are provided in Table~\ref{tab:efficiency_comparison}.

\noindent\textbf{Number of Samples for Calculating Fisher in Eqn.~(\ref{eq:fisher_information}).}
As described in Section~\ref{sec:adaptation_wo_forgetting}, the calculation of Fisher information involves a small set of unlabeled ID samples, which can be collected via existing OOD detection techniques~\cite{berger2021confidence}. Here, we investigate the effect of \#samples $Q$, selected from \{200, 300, 500, 700, 1000, 1500, 2000, 3000\}. From Figure~\ref{fig:number_samples_fisher}, our \myeata achieves stable performance with $Q\geq 300$, \ie, compared with \myeta, the OOD performance is comparable and the clean accuracy is much higher. These results show that our \myeata does not need to collect too many ID samples, which are easy to obtain in practice.

\noindent \textbf{Factor $\alpha$ for Entropy Regularizer in Eqn.~\ref{eq:eata-c objective}.} We directly set $\alpha{=}0.1$ to align the magnitudes of consistency loss and entropy regularization loss for \myeataC without careful tuning. From Figure~\ref{fig:alpha_influence}, increasing $\alpha$ within $[0,0.1]$ effectively reduces more ECE while maintaining stable accuracy, verifying its efficacy. However, when $\alpha$ exceeds 0.1, the entropy regularization loss dominates the adaptation, which leads to a gradual decline in accuracy.

\noindent \textbf{Factor $\beta$ for Fisher Regularizer in Eqn.~\ref{eq:eata-c objective}.} From Figure~\ref{fig:beta_influence}, compared to ETA-C (\ie, setting $\beta{=}0$), introducing the fisher regularizer consistently achieves better accuracy. Moreover, once activated, the performance of \myeata becomes largely insensitive to $\beta$ within the tested range of $[20, 140]$, highlighting its robustness.

\noindent\textbf{Stochastic Depth Ratio for Obtaining Sub-Network.} In Eqn.~(\ref{eq:consistency loss}), We generate an extra prediction from the sub-network to measure model uncertainty, where the sub-network is obtained via stochastic depth~\cite{huang2016depth} throughout the experiments. We evaluate the effect of stochastic depth ratio selected from \{0.05, 0.1, 0.15, 0.2, 0.25, 0.3, 0.35 ,0.4\}. As shown in Figure~\ref{fig:stochastic_depth_influence}, our \myeataC achieves a satisfying performance-calibration trade-off when the ratio belongs to $[0.15,0.25]$, \black{where the full network consistently outperforms the sub-network while the sub-network retains sufficient capacity for learning}. We fix the ratio to 0.2 for all other ImageNet experiments.

\subsection{\camera{More Discussions}}

\noindent \textbf{Efficiency Analysis of \myeata and \myeataC.} We evaluate the efficiency of our methods by including a more comprehensive comparison of time and memory usage for TTA, as in Table~\ref{tab:efficiency_comparison}. The results reveal the following: 1) \textit{\textbf{Adaptation Time}}: \myeata and \myeataC require significantly less adaptation time than most baseline methods, while maintaining competitive or superior accuracy. For example, on ViT-Base, \myeataC improves accuracy from 48.8\% (ROID) to 56.8\% with a reduction in adaptation time from 181.6 seconds to 114.9 seconds. Moreover, by setting a stricter threshold $\epsilon$ to filter redundant test samples, \myeata and \myeataC can be further accelerated while maintaining performance. For example, using $\epsilon{=}0.02$, \myeata on ViT-Base increases accuracy from 33.4\% (Tent) to 50.7\% while reducing adaptation time from 106.1 seconds to 85.1 seconds; 2) \textit{\textbf{Memory Usage}}: Our methods also demonstrate efficient memory utilization, where \myeata and \myeataC consume substantially less memory compared to all competing TTA methods, \eg, on ResNet-50, memory usage decreases from 5417.7MB (SAR) to 2205.5MB (\myeataC, $\epsilon{=}0.02$) while accuracy increases from 29.6\% to 35.9\%. This memory reduction is achieved through our active sample selection strategy, which reduces the number of samples involved in backpropagation. Note that the current Pytorch implementation does not support instance-wise gradient computation, thus an ideal implementation should further speed up both \myeata and \myeataC. See more discussions on our implementation details in Appendix~\ref{supp:sec:more_impl}.

\begin{table}[t]
        \caption{Efficiency comparison regarding wall-clock time and peak memory usage on ImageNet-C (Gaussian, severity level 5) with an A100 GPU.}
        \setlength{\tabcolsep}{3pt}
         \vspace{-0.15in}
        	\label{tab:efficiency_comparison}
        \newcommand{\tabincell}[2]{\begin{tabular}{@{}#1@{}}#2\end{tabular}}
         \begin{center}
         \begin{threeparttable}
            \resizebox{1\linewidth}{!}{
         	\begin{tabular}{l|ccc|ccc}\toprule
                \multicolumn{1}{l|}{\multirow{2}[0]{*}{Method}} & \multicolumn{3}{c|}{ResNet-50} & \multicolumn{3}{c}{ViT-Base} \\
                 & \multicolumn{1}{c}{Acc. (\%)} & \multicolumn{1}{c}{Time (s)} & \multicolumn{1}{c|}{Mem. (MB)} & \multicolumn{1}{c}{Acc. (\%)} & \multicolumn{1}{c}{Time (s)} & \multicolumn{1}{c}{Mem. (MB)} \\
                \midrule
                 Source & 1.8 & 54.6 & 771.7 & 12.9 & 55.7 & 816.6 \\
                 Tent~\cite{wang2021tent} & 28.0 & 99.9 & 5417.6 & 33.4 & 106.1 & 7433.2 \\
                 SAR~\cite{niu2023sar} & 29.6 & 149.0 & 5417.7 & 43.1 & 167.3 & 7433.2 \\
                 ROID~\cite{marsden2024universal} & 36.7 & 190.2 & 9315.3 & 48.8 & 181.6 & 12321.4 \\
                 CoTTA~\cite{wang2022continual} & 19.9 & 241.0 & 12196.7 & 45.6 & 229.9 & 22296.5 \\
                 TEA~\cite{yuan2024tea} & 18.5 & 2266.1 & 15942.0 & 46.9 & - & - \\
                 MEMO~\cite{zhang2021memo} & 6.8 & 36329.2 & 8154.3 & 32.2 & 39918.8 & 11061.1 \\
                 \midrule
                 EATA ($\epsilon{=}0.02$) & 35.2 & \textbf{93.0} & 2285.2 & 50.7 & \textbf{85.1} & 4234.6 \\
                 EATA-C ($\epsilon{=}0.02$) & 35.9 & 95.2 & \textbf{2205.5} & 55.0 & 88.8 & \textbf{4014.4} \\
                 EATA (Ours) & 35.6 & 106.5 & 3693.0 & 50.5 & 108.7 & 5887.6 \\
                 EATA-C (Ours) & \textbf{37.2} & 122.4 & 3358.7 & \textbf{56.8} & 114.9 & 5786.6 \\
                \bottomrule
        	\end{tabular}}
        	 \end{threeparttable}
        	 \end{center}
        \end{table}

\begin{table}[t]
    \vspace{-0.1in}
    \caption{Reliability of data uncertainty indicator. We report sub-model Acc. (\%) on ImageNet-C(Gaussian, level 5) after adapting to $B$ batches (batch size 64) with ViT-Base. ``\#Uncertain'' are samples with disagreed predictions. ``Indicator Acc.'' is the ratio of these samples misclassified by sub-model.}
    \setlength{\tabcolsep}{2.5pt}
     \vspace{-0.15in}
	\label{tab:uncertain_indicator_accuracy}
\newcommand{\tabincell}[2]{\begin{tabular}{@{}#1@{}}#2\end{tabular}}
 \begin{center}
 \begin{threeparttable}
    \resizebox{1.\linewidth}{!}{
 	\begin{tabular}{l|cccccc}
 	\toprule
        \multirow{2}{*}{\tabincell{c}{Metric}}&\multirow{2}{*}{\tabincell{c}{Source}}& \multirow{1}{*}{\tabincell{c}{EATA-C}} & \multirow{1}{*}{\tabincell{c}{EATA-C}} & \multirow{1}{*}{\tabincell{c}{EATA-C}} & \multirow{1}{*}{\tabincell{c}{EATA-C}} & \multirow{1}{*}{\tabincell{c}{EATA-C}}\\
        ~ & ~ & ($B{=}150$) & ($B{=}300$) & ($B{=}450$) & ($B{=}600$) & ($B{=}750$) \\
        \midrule
        Model Acc. (\%) & 9.3 & 47.0 & 50.2 & 50.7 & 51.9 & 52.0 \\
        \#Uncertain & 26144 & 15697 & 14682 & 14149 & 13648 & 13722 \\
        Indicator Acc. (\%) & 97.1 & 90.0 & 88.9 & 89.3 & 88.6 & 89.1 \\
        \bottomrule
	\end{tabular}}
	 \end{threeparttable}
	 \end{center}
     \vspace{-0.1in}
\end{table}

\begin{figure}[t]
    \centering
    \begin{minipage}[t]{0.48\linewidth}
        \centering
        \includegraphics[width=1.0\linewidth]{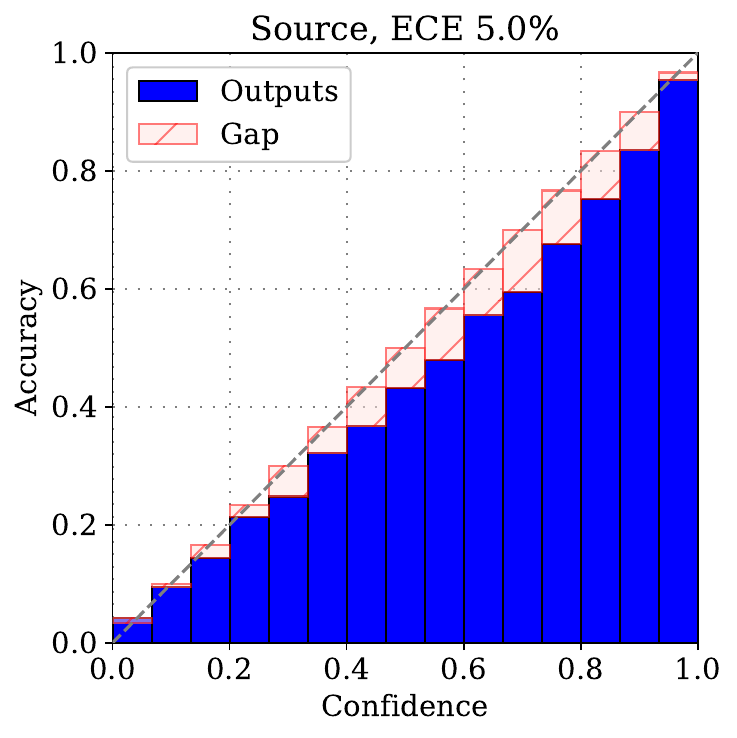}
    \end{minipage}
    \hfill
    \begin{minipage}[t]{0.48\linewidth}
        \centering
        \includegraphics[width=1.0\linewidth]{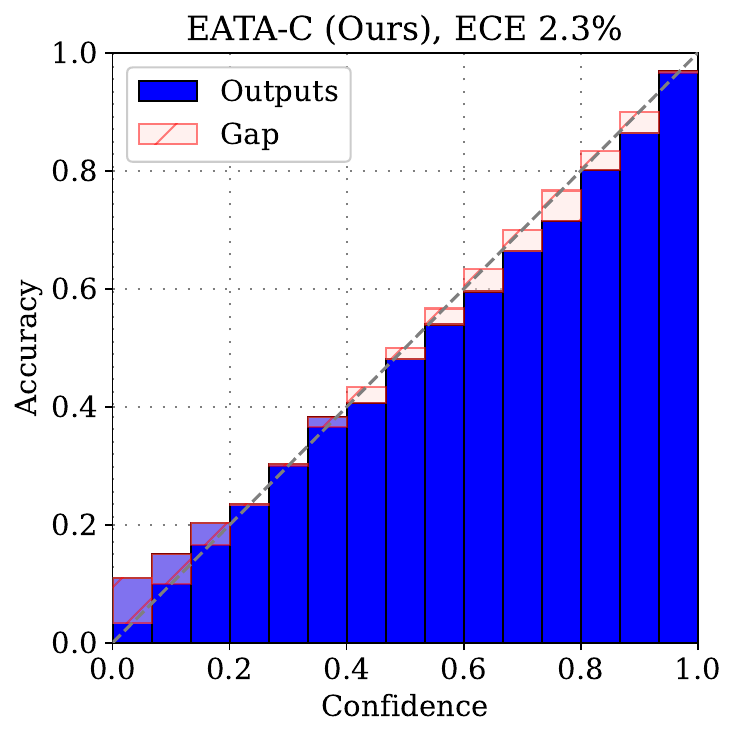}
    \end{minipage}
    \vspace{-0.1in}
    \caption{Calibration comparison of each confidence interval on ImageNet-C(Fog, level 5) with ViT-Base. Results are visualized following~\cite{guo2017calibration}.}
    \label{fig:ece_interval}
\end{figure}

\noindent \textbf{Effectiveness of Data Uncertainty Indicator.} We evaluate the effectiveness of our data uncertainty indicator, \ie, Eqn.~(\ref{eq:complementary_certainty}), throughout TTA. The results are detailed in Table~\ref{tab:uncertain_indicator_accuracy}: 1)~\textit{Consistently High Indicator Accuracy}: Across various adaptation stages, our indicator continues to reliably identify uncertain samples, on which the sub-model’s prediction is likely to be incorrect. Specifically, the indicator maintains around 90\% accuracy, suggesting its effectiveness even in the early stage of adaptation. This reliability allows us to apply entropy maximization on these uncertain data to improve calibration without hindering adaptation; 2) \textit{Reduced Uncertain Samples Over Time}: The model's initial poor performance mainly leads to a higher number of uncertain samples, including inherently difficult data for discrimination and data the model has yet to learn. 
However, model uncertainty quickly explains away during TTA (\ie, within one-fifth of the data stream), leading to a stabilized number of uncertain samples that reflects irreducible data uncertainty.

\noindent\textbf{Additional Memory by Fisher Regularizer.} 
Since we only regularize the affine parameters of normalization layers, \myeata needs very little extra memory. For ResNet-50 on ImageNet-C, the extra GPU memory at run time is only 9.8 MB, which is much less than that of Tent with batch size 64 (5,675 MB).

\noindent\textbf{Performance under Mixed-and-Shifted Distributions.} We evaluate Tent and our \myeata/\myeataC on mixed ImageNet-C (level=3 or 5) that consists of 15 different corruption types/distribution shifts (totaling 750k images). Results in Table~\ref{tab:mixed_distribution} show the stability of \myeata and \myeataC in large-scale and complex TTA scenarios.

\begin{table}[t!]
    \caption{Comparison with Tent~\cite{wang2021tent} \wrt corruption accuracy (\%) with mixture of 15 corruption types on ImageNet-C with ViT-Base.}
    \vspace{-0.15in}
	\label{tab:mixed_distribution}
\newcommand{\tabincell}[2]{\begin{tabular}{@{}#1@{}}#2\end{tabular}}
 \begin{center}
 \begin{threeparttable}
    \resizebox{1.0\linewidth}{!}{
 	\begin{tabular}{l|c|c|c|c}
 	\toprule
        \multirow{1}{*}{\tabincell{c}{~Severity~~}}&\multirow{1}{*}{\tabincell{c}{Source}}&\multirow{1}{*}{\tabincell{c}{Tent}}&\multirow{1}{*}{\tabincell{c}{\myeata (ours)}}&\multirow{1}{*}{\tabincell{c}{\myeataC (ours)}}\\
        \midrule
         Level=3 &63.0	&69.8$_{(+6.8)}$ &71.1$_{(+8.1)}$ & \textbf{72.4}$_{(+9.4)}$     \\
         Level=5 &39.8	&47.0$_{(+7.2)}$ &58.2$_{(+18.4)}$ & \textbf{60.4}$_{(+20.6)}$     \\
        \bottomrule
	\end{tabular}}
	 \end{threeparttable}
	 \end{center}
  \vspace{-0.1in}
\end{table}

\begin{figure*}[t]
    \centering
    \subfigure[Performance and calibration throughout adaptation]{\label{fig:consistency_entropy_comparison}\includegraphics[width=71mm]{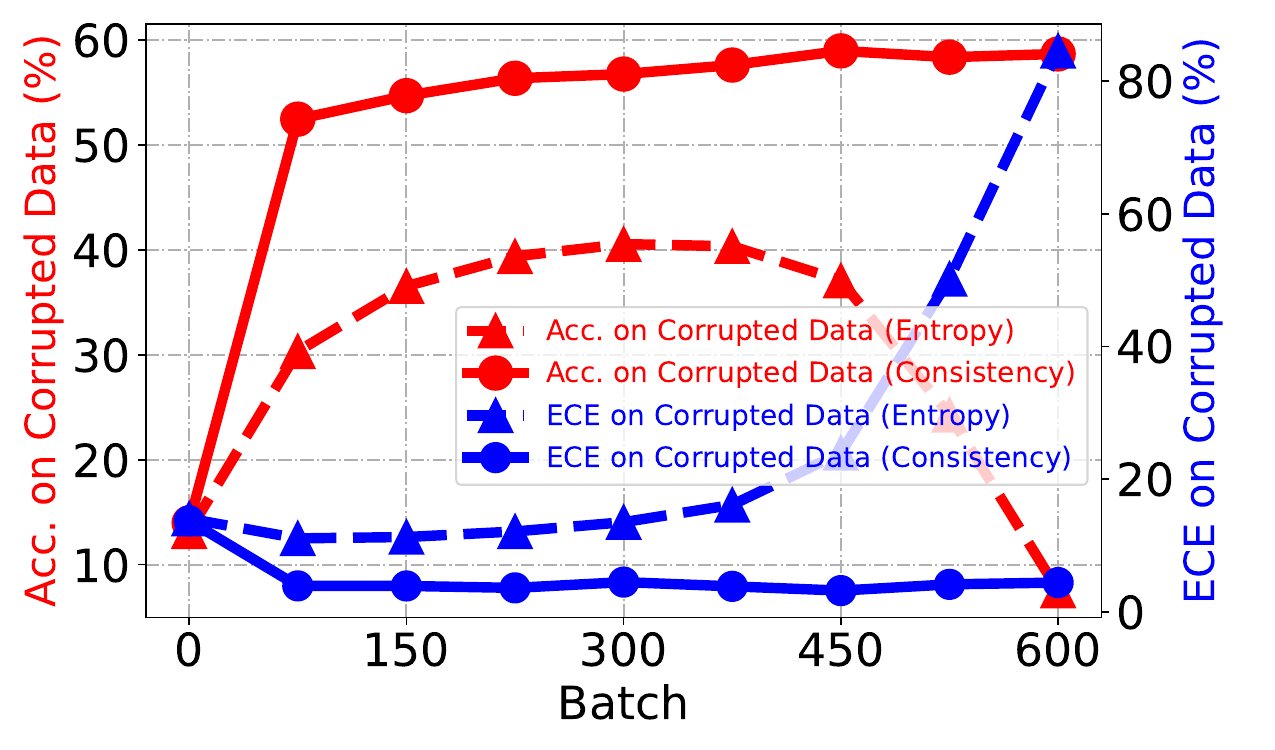}}
    \subfigure[Consistency maximization via Eqn.~(\ref{eq:consistency loss})]{\label{fig:consistent_heatmap}\includegraphics[width=52mm]{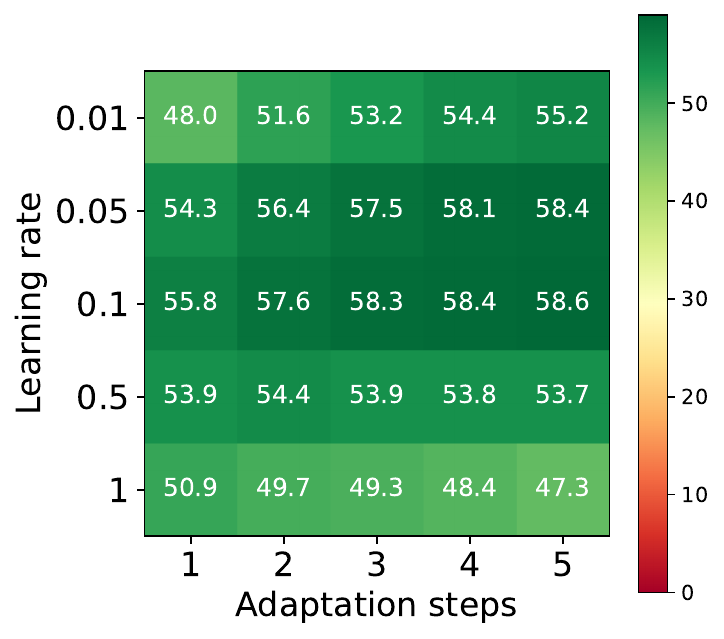}}
    \subfigure[Entropy minimization (Tent~\cite{wang2021tent})]{\label{fig:tent_heatmap}\includegraphics[width=52mm]{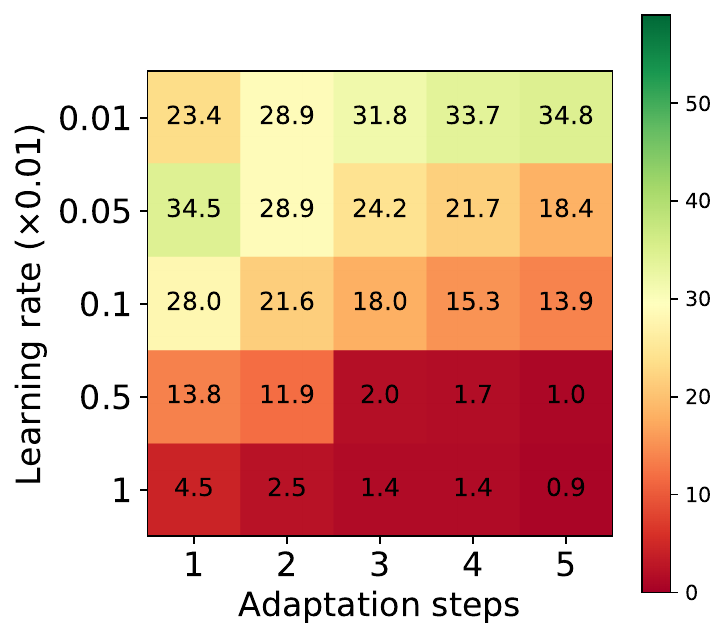}}
    \vspace{-0.05in}
    \caption{Comparisons of performance, calibration and stability using consistency loss in Eqn.~(\ref{eq:consistency loss}) or entropy minimization loss (\ie, Tent) on ImageNet-C (Gaussian noise, severity level=5) with ViT-Base. In \textbf{Left}, we split the dataset into 80\% and 20\% slices to conduct adaptation and to evaluate the adapted model. We report the performance and the calibration of the adapted model every 75 batches with a batch size of 64. In the Middle and Right, we evaluate the stability of consistency loss and Tent under various combinations of learning rates and adaptation steps per batch. Consistency loss achieves substantially higher OOD performance and better stability while maintaining lower ECE.}
    \label{fig:imageC-consistency-entropy}
\end{figure*}

\noindent \textbf{Calibration Across Confidence Intervals.} \myeataC aims to achieve a favorable balance among accuracy, calibration, and efficiency. In \myeataC, we discard high-entropy samples (termed active sample selection) mainly to improve computational efficiency. While some high-entropy samples might benefit from further calibration, they typically yield unreliable pseudo-labels, which can negatively impact the stability and effectiveness of TTA (see Table~\ref{tab:eata-c_ablation}). Instead of directly calibrating on these samples, we show that focusing adaptation and calibration on only low-entropy samples can also improve the calibration on high-entropy ones, as in Figure~\ref{fig:ece_interval}, while improving TTA efficiency and effectiveness.

\noindent\textbf{Advantage of Consistency Maximization over Tent~\cite{wang2021tent}.} 
We conduct a comprehensive comparison w.r.t. performance, calibration, and stability between the use of consistency loss, as defined in Eqn.~(\ref{eq:consistency loss}), and Tent~\cite{wang2021tent} to reduce uncertainty during testing. 
From Figure~\ref{fig:consistency_entropy_comparison}, we have the following observations:
1)~Consistency loss consistently demonstrates superior performance and calibration throughout adaptation. 2)~Consistency loss is more sample-efficient, where adapting with as few as 75 batches can significantly outperform Tent~\cite{wang2021tent} that adapts with 300 batches. 3)~Tent~\cite{wang2021tent} shows rapid degradation in performance and calibration after convergence. In contrast, consistency loss maintains stable performance and calibration after convergence and continuously exhibits strong generalization throughout TTA.

We further evaluate the stability of consistency loss and Tent~\cite{wang2021tent} across various combinations of learning rates and adaptation steps in Figures~\ref{fig:consistent_heatmap} and~\ref{fig:tent_heatmap} following \cite{zhao2023ttapitfal}. The results highlight that our consistency loss demonstrates remarkable stability and benefits from increased adaptation steps (\textit{e.g.}, $55.8\% \rightarrow 58.6\%$, the best performance under 1 and 5 adaptation steps, respectively). In contrast, Tent~\cite{wang2021tent} is highly sensitive to the combination of learning rate and adaptation steps, where its performance may deteriorate to as low as 1\%, further indicating its tendency to overfit. These findings collectively underscore the superiority of our consistency loss regarding performance, calibration, and stability, making it a more robust choice for TTA.

\section{Conclusion}
In this paper, we have proposed an Efficient Anti-forgetting Test-time Adaptation method (EATA), to improve the performance of pre-trained models on a potentially shifted test domain. To be specific, we devise a sample-efficient entropy minimization strategy that selectively performs test-time optimization with reliable and non-redundant samples. This improves the adaptation efficiency and meanwhile boosts the out-of-distribution performance. In addition, we introduce a Fisher-based anti-forgetting regularizer into test-time adaptation. With this loss, a model can be adapted continually without performance degradation on in-distribution test samples. Moreover, we design EATA with Calibration (EATA-C) for test-time adapted model's confidence calibration. To this end, we present a consistency loss for calibrated model uncertainty reduction and a sample-aware min-max entropy regularization for confidence re-calibration, which improves the performance and calibration of test-time adaptation. Extensive experimental results on image classification and semantic segmentation demonstrate the effectiveness of our proposed methods.

\section*{Acknowledgments}
This work was partially supported by the Joint Funds of the National Natural Science Foundation of China (Grant No.U24A20327), and TCL Science and Technology Innovation Fund, China.

\bibliographystyle{IEEEtran}
{
\bibliography{main}

\begin{thebibliography}{10}
\providecommand{\url}[1]{#1}
\csname url@samestyle\endcsname
\providecommand{\newblock}{\relax}
\providecommand{\bibinfo}[2]{#2}
\providecommand{\BIBentrySTDinterwordspacing}{\spaceskip=0pt\relax}
\providecommand{\BIBentryALTinterwordstretchfactor}{4}
\providecommand{\BIBentryALTinterwordspacing}{\spaceskip=\fontdimen2\font plus
\BIBentryALTinterwordstretchfactor\fontdimen3\font minus \fontdimen4\font\relax}
\providecommand{\BIBforeignlanguage}[2]{{%
\expandafter\ifx\csname l@#1\endcsname\relax
\typeout{** WARNING: IEEEtran.bst: No hyphenation pattern has been}%
\typeout{** loaded for the language `#1'. Using the pattern for}%
\typeout{** the default language instead.}%
\else
\language=\csname l@#1\endcsname
\fi
#2}}
\providecommand{\BIBdecl}{\relax}
\BIBdecl

\bibitem{he2016deep}
K.~He, X.~Zhang, S.~Ren, and J.~Sun, ``Deep residual learning for image recognition,'' in \emph{IEEE Conference on Computer Vision and Pattern Recognition}, 2016, pp. 770--778.

\bibitem{wang2018nonlocal}
X.~Wang, R.~Girshick, A.~Gupta, and K.~He, ``Non-local neural networks,'' in \emph{IEEE Conference on Computer Vision and Pattern Recognition}, 2018, pp. 7794--7803.

\bibitem{zeng2020dense}
R.~Zeng, H.~Xu, W.~Huang, P.~Chen, M.~Tan, and C.~Gan, ``Dense regression network for video grounding,'' in \emph{IEEE Conference on Computer Vision and Pattern Recognition}, 2020, pp. 10\,284--10\,293.

\bibitem{zeng2021graph}
R.~Zeng, W.~Huang, M.~Tan, Y.~Rong, P.~Zhao, J.~Huang, and C.~Gan, ``Graph convolutional module for temporal action localization in videos,'' \emph{IEEE Transactions on Pattern Analysis and Machine Intelligence}, pp. 6209--6223, 2021.

\bibitem{chen2021rspnet}
P.~Chen, D.~Huang, D.~He, X.~Long, R.~Zeng, S.~Wen, M.~Tan, and C.~Gan, ``Rspnet: Relative speed perception for unsupervised video representation learning,'' in \emph{AAAI Conference on Artificial Intelligence}, vol.~1, 2021, pp. 1045--1053.

\bibitem{choi2018stargan}
Y.~Choi, M.~Choi, M.~Kim, J.-W. Ha, S.~Kim, and J.~Choo, ``Stargan: Unified generative adversarial networks for multi-domain image-to-image translation,'' in \emph{IEEE Conference on Computer Vision and Pattern Recognition}, 2018, pp. 8789--8797.

\bibitem{fan2020inf}
D.-P. Fan, T.~Zhou, G.-P. Ji, Y.~Zhou, G.~Chen, H.~Fu, J.~Shen, and L.~Shao, ``Inf-net: Automatic covid-19 lung infection segmentation from ct images,'' \emph{IEEE Transactions on Medical Imaging}, vol.~39, no.~8, pp. 2626--2637, 2020.

\bibitem{xu2021towards}
G.~Xu, S.~Niu, M.~Tan, Y.~Luo, Q.~Du, and Q.~Wu, ``Towards accurate text-based image captioning with content diversity exploration,'' in \emph{IEEE Conference on Computer Vision and Pattern Recognition}, 2021, pp. 12\,637--12\,646.

\bibitem{hendrycks2019benchmarking}
D.~Hendrycks and T.~Dietterich, ``Benchmarking neural network robustness to common corruptions and perturbations,'' in \emph{International Conference on Learning Representations}, 2019, pp. 1--11.

\bibitem{koh2021wilds}
P.~W. Koh, S.~Sagawa, H.~Marklund, S.~M. Xie, M.~Zhang, A.~Balsubramani, W.~Hu, M.~Yasunaga, R.~L. Phillips, I.~Gao \emph{et~al.}, ``Wilds: A benchmark of in-the-wild distribution shifts,'' in \emph{International Conference on Machine Learning}, 2021, pp. 5637--5664.

\bibitem{sun2020test}
Y.~Sun, X.~Wang, Z.~Liu, J.~Miller, A.~Efros, and M.~Hardt, ``Test-time training with self-supervision for generalization under distribution shifts,'' in \emph{International Conference on Machine Learning}, 2020, pp. 9229--9248.

\bibitem{wang2021tent}
D.~Wang, E.~Shelhamer, S.~Liu, B.~Olshausen, and T.~Darrell, ``Tent: Fully test-time adaptation by entropy minimization,'' in \emph{International Conference on Learning Representations}, 2021, pp. 1--12.

\bibitem{liu2021ttt++}
Y.~Liu, P.~Kothari, B.~van Delft, B.~Bellot-Gurlet, T.~Mordan, and A.~Alahi, ``Ttt++: When does self-supervised test-time training fail or thrive?'' in \emph{Advances in Neural Information Processing Systems}, vol.~34, 2021, pp. 21\,808--21\,820.

\bibitem{zhang2021memo}
M.~M. Zhang, S.~Levine, and C.~Finn, ``Memo: Test time robustness via adaptation and augmentation,'' in \emph{Advances in Neural Information Processing Systems}, 2022, pp. 38\,629--38\,642.

\bibitem{zhang2022Self}
Y.~Zhang, B.~Hooi, L.~Hong, and J.~Feng, ``Self-supervised aggregation of diverse experts for test-agnostic long-tailed recognition,'' in \emph{Advances in Neural Information Processing Systems}, 2022, pp. 34\,077--34\,090.

\bibitem{wang2022continual}
Q.~Wang, O.~Fink, L.~Van~Gool, and D.~Dai, ``Continual test-time domain adaptation,'' in \emph{IEEE Conference on Computer Vision and Pattern Recognition}, 2022, pp. 7201--7211.

\bibitem{gidaris2018unsupervised}
S.~Gidaris, P.~Singh, and N.~Komodakis, ``Unsupervised representation learning by predicting image rotations,'' in \emph{International Conference on Learning Representations}, 2018, pp. 1--14.

\bibitem{niu2023sar}
S.~Niu, J.~Wu, Y.~Zhang, Z.~Wen, Y.~Chen, P.~Zhao, and M.~Tan, ``Towards stable test-time adaptation in dynamic wild world,'' in \emph{Internetional Conference on Learning Representations}, 2023, pp. 1--14.

\bibitem{bojarski2016driving}
M.~Bojarski, D.~Del~Testa, D.~Dworakowski, B.~Firner, B.~Flepp, P.~Goyal, L.~D. Jackel, M.~Monfort, U.~Muller, J.~Zhang \emph{et~al.}, ``End to end learning for self-driving cars,'' \emph{arXiv preprint arXiv:1604.07316}, 2016.

\bibitem{anwar2018medical}
S.~M. Anwar, M.~Majid, A.~Qayyum, M.~Awais, M.~Alnowami, and M.~K. Khan, ``Medical image analysis using convolutional neural networks: a review,'' \emph{Journal of medical systems}, vol.~42, no.~11, pp. 1--13, 2018.

\bibitem{kirkpatrick2017overcoming}
J.~Kirkpatrick, R.~Pascanu, N.~Rabinowitz, J.~Veness, G.~Desjardins, A.~A. Rusu, K.~Milan, J.~Quan, T.~Ramalho, A.~Grabska-Barwinska \emph{et~al.}, ``Overcoming catastrophic forgetting in neural networks,'' \emph{Proceedings of the national academy of sciences}, vol. 114, no.~13, pp. 3521--3526, 2017.

\bibitem{gal2016dropout}
Y.~Gal and Z.~Ghahramani, ``Dropout as a bayesian approximation: Representing model uncertainty in deep learning,'' in \emph{international conference on machine learning}.\hskip 1em plus 0.5em minus 0.4em\relax PMLR, 2016, pp. 1050--1059.

\bibitem{kendall2017uncertainties}
A.~Kendall and Y.~Gal, ``What uncertainties do we need in bayesian deep learning for computer vision?'' in \emph{Advances in Neural Information Processing Systems}, vol.~30, 2017, pp. 5574--5584.

\bibitem{niu2022eata}
S.~Niu, J.~Wu, Y.~Zhang, Y.~Chen, S.~Zheng, P.~Zhao, and M.~Tan, ``Efficient test-time model adaptation without forgetting,'' in \emph{International conference on machine learning}.\hskip 1em plus 0.5em minus 0.4em\relax PMLR, 2022, pp. 16\,888--16\,905.

\bibitem{dosovitskiy2020vit}
A.~Dosovitskiy, L.~Beyer, A.~Kolesnikov, D.~Weissenborn, X.~Zhai, T.~Unterthiner, M.~Dehghani, M.~Minderer, G.~Heigold, S.~Gelly \emph{et~al.}, ``An image is worth 16x16 words: Transformers for image recognition at scale,'' in \emph{International Conference on Learning Representations}, 2021, pp. 1--12.

\bibitem{gandelsman2022tmae}
Y.~Gandelsman, Y.~Sun, X.~Chen, and A.~Efros, ``Test-time training with masked autoencoders,'' in \emph{Advances in Neural Information Processing Systems}, vol.~35, 2022, pp. 29\,374--29\,385.

\bibitem{bartler2022mt3}
A.~Bartler, A.~B{\"u}hler, F.~Wiewel, M.~D{\"o}bler, and B.~Yang, ``Mt3: Meta test-time training for self-supervised test-time adaption,'' in \emph{International Conference on Artificial Intelligence and Statistics}.\hskip 1em plus 0.5em minus 0.4em\relax PMLR, 2022, pp. 3080--3090.

\bibitem{nado2020evaluating}
Z.~Nado, S.~Padhy, D.~Sculley, A.~D'Amour, B.~Lakshminarayanan, and J.~Snoek, ``Evaluating prediction-time batch normalization for robustness under covariate shift,'' \emph{arXiv preprint arXiv:2006.10963}, 2020.

\bibitem{schneider2020improving}
S.~Schneider, E.~Rusak, L.~Eck, O.~Bringmann, W.~Brendel, and M.~Bethge, ``Improving robustness against common corruptions by covariate shift adaptation,'' in \emph{Advances in Neural Information Processing Systems}, vol.~33, 2020, pp. 11\,539--11\,551.

\bibitem{reddy2024towards}
N.~Reddy, M.~Baktashmotlagh, and C.~Arora, ``Towards domain-aware knowledge distillation for continual model generalization,'' in \emph{{Winter Conference on Applications of Computer Vision}}, 2024, pp. 696--707.

\bibitem{fleuret2021test}
F.~Fleuret \emph{et~al.}, ``Test time adaptation through perturbation robustness,'' in \emph{Advances in Neural Information Processing Systems Workshop}, 2021.

\bibitem{iwasawa2021test}
Y.~Iwasawa and Y.~Matsuo, ``Test-time classifier adjustment module for model-agnostic domain generalization,'' in \emph{Advances in Neural Information Processing Systems}, vol.~34, 2021, pp. 2427--2440.

\bibitem{li2017learning}
Z.~Li and D.~Hoiem, ``Learning without forgetting,'' \emph{IEEE transactions on pattern analysis and machine intelligence}, vol.~40, no.~12, pp. 2935--2947, 2017.

\bibitem{rolnick2019experience}
D.~Rolnick, A.~Ahuja, J.~Schwarz, T.~P. Lillicrap, and G.~Wayne, ``Experience replay for continual learning,'' in \emph{Advances in Neural Information Processing Systems}, 2019, pp. 348--358.

\bibitem{farajtabar2020orthogonal}
M.~Farajtabar, N.~Azizan, A.~Mott, and A.~Li, ``Orthogonal gradient descent for continual learning,'' in \emph{International Conference on Artificial Intelligence and Statistics}, 2020, pp. 3762--3773.

\bibitem{niu2021disturbance}
S.~Niu, J.~Wu, Y.~Zhang, Y.~Guo, P.~Zhao, J.~Huang, and M.~Tan, ``Disturbance-immune weight sharing for neural architecture search,'' \emph{Neural Networks}, vol. 144, pp. 553--564, 2021.

\bibitem{Mittal_2021_CVPR}
S.~Mittal, S.~Galesso, and T.~Brox, ``Essentials for class incremental learning,'' in \emph{IEEE Conference on Computer Vision and Pattern Recognition}, June 2021, pp. 3513--3522.

\bibitem{pei2018multi}
Z.~Pei, Z.~Cao, M.~Long, and J.~Wang, ``Multi-adversarial domain adaptation,'' in \emph{AAAI Conference on Artificial Intelligence}, 2018, pp. 3934--3941.

\bibitem{saito2018maximum}
K.~Saito, K.~Watanabe, Y.~Ushiku, and T.~Harada, ``Maximum classifier discrepancy for unsupervised domain adaptation,'' in \emph{IEEE Conference on Computer Vision and Pattern Recognition}, 2018, pp. 3723--3732.

\bibitem{zhang2020collaborative}
Y.~Zhang, Y.~Wei, Q.~Wu, P.~Zhao, S.~Niu, J.~Huang, and M.~Tan, ``Collaborative unsupervised domain adaptation for medical image diagnosis,'' \emph{IEEE Transactions on Image Processing}, vol.~29, pp. 7834--7844, 2020.

\bibitem{zhang2020covid}
Y.~Zhang, S.~Niu, Z.~Qiu, Y.~Wei, P.~Zhao, J.~Yao, J.~Huang, Q.~Wu, and M.~Tan, ``Covid-da: deep domain adaptation from typical pneumonia to covid-19,'' \emph{arXiv preprint arXiv:2005.01577}, 2020.

\bibitem{Qiu2021CPGA}
Z.~Qiu, Y.~Zhang, H.~Lin, S.~Niu, Y.~Liu, Q.~Du, and M.~Tan, ``Source-free domain adaptation via avatar prototype generation and adaptation,'' in \emph{International Joint Conference on Artificial Intelligence}, 2021, pp. 2921--2927.

\bibitem{liang2020we}
J.~Liang, D.~Hu, and J.~Feng, ``Do we really need to access the source data? source hypothesis transfer for unsupervised domain adaptation,'' in \emph{International Conference on Machine Learning}, 2020, pp. 6028--6039.

\bibitem{guo2017calibration}
C.~Guo, G.~Pleiss, Y.~Sun, and K.~Q. Weinberger, ``On calibration of modern neural networks,'' in \emph{International conference on machine learning}.\hskip 1em plus 0.5em minus 0.4em\relax PMLR, 2017, pp. 1321--1330.

\bibitem{naeini2015obtaining}
M.~P. Naeini, G.~Cooper, and M.~Hauskrecht, ``Obtaining well calibrated probabilities using bayesian binning,'' in \emph{Proceedings of the AAAI conference on artificial intelligence}, vol.~29, no.~1, 2015, pp. 2901--2907.

\bibitem{zhang2020mix}
J.~Zhang, B.~Kailkhura, and T.~Y.-J. Han, ``Mix-n-match: Ensemble and compositional methods for uncertainty calibration in deep learning,'' in \emph{International conference on machine learning}.\hskip 1em plus 0.5em minus 0.4em\relax PMLR, 2020, pp. 11\,117--11\,128.

\bibitem{kumar2018trainable}
A.~Kumar, S.~Sarawagi, and U.~Jain, ``Trainable calibration measures for neural networks from kernel mean embeddings,'' in \emph{International Conference on Machine Learning}.\hskip 1em plus 0.5em minus 0.4em\relax PMLR, 2018, pp. 2805--2814.

\bibitem{seo2019learning}
S.~Seo, P.~H. Seo, and B.~Han, ``Learning for single-shot confidence calibration in deep neural networks through stochastic inferences,'' in \emph{Proceedings of the IEEE/CVF conference on computer vision and pattern recognition}, 2019, pp. 9030--9038.

\bibitem{park2020calibrated}
S.~Park, O.~Bastani, J.~Weimer, and I.~Lee, ``Calibrated prediction with covariate shift via unsupervised domain adaptation,'' in \emph{International Conference on Artificial Intelligence and Statistics}.\hskip 1em plus 0.5em minus 0.4em\relax PMLR, 2020, pp. 3219--3229.

\bibitem{wang2020transferable}
X.~Wang, M.~Long, J.~Wang, and M.~Jordan, ``Transferable calibration with lower bias and variance in domain adaptation,'' in \emph{Advances in Neural Information Processing Systems}, vol.~33, 2020, pp. 19\,212--19\,223.

\bibitem{karandikar2021soft}
A.~Karandikar, N.~Cain, D.~Tran, B.~Lakshminarayanan, J.~Shlens, M.~C. Mozer, and B.~Roelofs, ``Soft calibration objectives for neural networks,'' in \emph{Advances in Neural Information Processing Systems}, vol.~34, 2021, pp. 29\,768--29\,779.

\bibitem{yoon2023esd}
H.~S. Yoon, J.~T.~J. Tee, E.~Yoon, S.~Yoon, G.~Kim, Y.~Li, and C.~D. Yoo, ``{ESD}: Expected squared difference as a tuning-free trainable calibration measure,'' in \emph{International Conference on Learning Representations}, 2023, pp. 1--12.

\bibitem{liu2020energy}
W.~Liu, X.~Wang, J.~Owens, and Y.~Li, ``Energy-based out-of-distribution detection,'' in \emph{Advances in Neural Information Processing Systems}, vol.~33, 2020, pp. 21\,464--21\,475.

\bibitem{berger2021confidence}
C.~Berger, M.~Paschali, B.~Glocker, and K.~Kamnitsas, ``Confidence-based out-of-distribution detection: A comparative study and analysis,'' in \emph{Uncertainty for Safe Utilization of Machine Learning in Medical Imaging, and Perinatal Imaging, Placental and Preterm Image Analysis}.\hskip 1em plus 0.5em minus 0.4em\relax Springer, 2021, pp. 122--132.

\bibitem{chapelle2005semi}
O.~Chapelle and A.~Zien, ``Semi-supervised classification by low density separation,'' in \emph{International workshop on artificial intelligence and statistics}.\hskip 1em plus 0.5em minus 0.4em\relax PMLR, 2005, pp. 57--64.

\bibitem{liang2023ttasurvey}
J.~Liang, R.~He, and T.~Tan, ``A comprehensive survey on test-time adaptation under distribution shifts,'' \emph{International Journal of Computer Vision}, pp. 1--34, 2024.

\bibitem{gal2016mcdropout}
Y.~Gal and Z.~Ghahramani, ``Dropout as a bayesian approximation: Representing model uncertainty in deep learning,'' in \emph{international conference on machine learning}.\hskip 1em plus 0.5em minus 0.4em\relax PMLR, 2016, pp. 1050--1059.

\bibitem{hinton2015kldivergence}
G.~Hinton, O.~Vinyals, and J.~Dean, ``Distilling the knowledge in a neural network,'' \emph{arXiv preprint arXiv:1503.02531}, 2015.

\bibitem{szegedy2016labelsm}
C.~Szegedy, V.~Vanhoucke, S.~Ioffe, J.~Shlens, and Z.~Wojna, ``Rethinking the inception architecture for computer vision,'' in \emph{Proceedings of the IEEE conference on computer vision and pattern recognition}, 2016, pp. 2818--2826.

\bibitem{huang2016depth}
G.~Huang, Y.~Sun, Z.~Liu, D.~Sedra, and K.~Q. Weinberger, ``Deep networks with stochastic depth,'' in \emph{Computer Vision--ECCV 2016: 14th European Conference, Amsterdam, The Netherlands, October 11--14, 2016, Proceedings, Part IV 14}.\hskip 1em plus 0.5em minus 0.4em\relax Springer, 2016, pp. 646--661.

\bibitem{balcan2007margin}
M.-F. Balcan, A.~Broder, and T.~Zhang, ``Margin based active learning,'' in \emph{International Conference on Computational Learning Theory}.\hskip 1em plus 0.5em minus 0.4em\relax Springer, 2007, pp. 35--50.

\bibitem{khan2019striking}
S.~Khan, M.~Hayat, S.~W. Zamir, J.~Shen, and L.~Shao, ``Striking the right balance with uncertainty,'' in \emph{Proceedings of the IEEE/CVF Conference on Computer Vision and Pattern Recognition}, 2019, pp. 103--112.

\bibitem{wang2022cotta}
Q.~Wang, O.~Fink, L.~Van~Gool, and D.~Dai, ``Continual test-time domain adaptation,'' in \emph{Proceedings of the IEEE/CVF Conference on Computer Vision and Pattern Recognition}, 2022, pp. 7201--7211.

\bibitem{marsden2024universal}
R.~A. Marsden, M.~D{\"o}bler, and B.~Yang, ``Universal test-time adaptation through weight ensembling, diversity weighting, and prior correction,'' in \emph{{Winter Conference on Applications of Computer Vision}}, 2024, pp. 2555--2565.

\bibitem{yuan2024tea}
Y.~Yuan, B.~Xu, L.~Hou, F.~Sun, H.~Shen, and X.~Cheng, ``Tea: Test-time energy adaptation,'' in \emph{IEEE Conference on Computer Vision and Pattern Recognition}, 2024, pp. 23\,901--23\,911.

\bibitem{hendrycks2021many}
D.~Hendrycks, S.~Basart, N.~Mu, S.~Kadavath, F.~Wang, E.~Dorundo, R.~Desai, T.~Zhu, S.~Parajuli, M.~Guo \emph{et~al.}, ``The many faces of robustness: A critical analysis of out-of-distribution generalization,'' in \emph{IEEE Conference on Computer Vision and Pattern Recognition}, 2021, pp. 8340--8349.

\bibitem{SDV21acdc}
C.~Sakaridis, D.~Dai, and L.~Van~Gool, ``{ACDC}: The adverse conditions dataset with correspondences for semantic driving scene understanding,'' in \emph{Proceedings of the IEEE/CVF International Conference on Computer Vision}, 2021, pp. 10\,765--10\,775.

\bibitem{xie2021segformer}
E.~Xie, W.~Wang, Z.~Yu, A.~Anandkumar, J.~M. Alvarez, and P.~Luo, ``Segformer: Simple and efficient design for semantic segmentation with transformers,'' in \emph{Advances in Neural Information Processing Systems}, vol.~34, 2021, pp. 12\,077--12\,090.

\bibitem{cordts2016cityscapes}
M.~Cordts, M.~Omran, S.~Ramos, T.~Rehfeld, M.~Enzweiler, R.~Benenson, U.~Franke, S.~Roth, and B.~Schiele, ``The cityscapes dataset for semantic urban scene understanding,'' in \emph{Proceedings of the IEEE conference on computer vision and pattern recognition}, 2016, pp. 3213--3223.

\bibitem{ni2024distribution}
J.~Ni, S.~Yang, R.~Xu, J.~Liu, X.~Li, W.~Jiao, Z.~Chen, Y.~Liu, and S.~Zhang, ``Distribution-aware continual test-time adaptation for semantic segmentation,'' in \emph{IEEE International Conference on Robotics and Automation}.\hskip 1em plus 0.5em minus 0.4em\relax IEEE, 2024, pp. 3044--3050.

\bibitem{press2024rdumb}
O.~Press, S.~Schneider, M.~K{\"u}mmerer, and M.~Bethge, ``Rdumb: A simple approach that questions our progress in continual test-time adaptation,'' in \emph{Advances in Neural Information Processing Systems}, vol.~36, 2024, pp. 39\,915--39\,935.

\bibitem{naeini2015ece}
M.~P. Naeini, G.~Cooper, and M.~Hauskrecht, ``Obtaining well calibrated probabilities using bayesian binning,'' in \emph{Proceedings of the AAAI conference on artificial intelligence}, vol.~29, no.~1, 2015, pp. 2901--2907.

\bibitem{zhao2023ttapitfal}
H.~Zhao, Y.~Liu, A.~Alahi, and T.~Lin, ``On pitfalls of test-time adaptation,'' in \emph{International Conference on Machine Learning (ICML)}, 2023, pp. 42\,058--42\,080.

\bibitem{wang2022semi}
Y.~Wang, H.~Wang, Y.~Shen, J.~Fei, W.~Li, G.~Jin, L.~Wu, R.~Zhao, and X.~Le, ``Semi-supervised semantic segmentation using unreliable pseudo-labels,'' in \emph{Proceedings of the IEEE/CVF Conference on Computer Vision and Pattern Recognition}, 2022, pp. 4248--4257.

\bibitem{pan2021model}
T.-Y. Pan, C.~Zhang, Y.~Li, H.~Hu, D.~Xuan, S.~Changpinyo, B.~Gong, and W.-L. Chao, ``On model calibration for long-tailed object detection and instance segmentation,'' in \emph{Advances in Neural Information Processing Systems}, vol.~34, 2021, pp. 2529--2542.

\end{thebibliography}
}

\vspace{-0.4in}
\begin{IEEEbiography}
[{\includegraphics[width=1in,height=1.25in,clip,keepaspectratio]{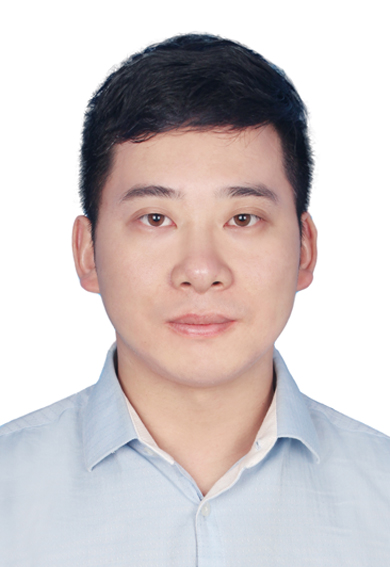}}]{Mingkui Tan}
is currently a Professor with the School of Software Engineering, South China University of Technology, Guangzhou, China. He received the Bachelor Degree in Environmental Science and Engineering in 2006 and the Master Degree in Control Science and Engineering in 2009, both from Hunan University in Changsha, China. He received the Ph.D. degree in Computer Science from Nanyang Technological University, Singapore, in 2014. From 2014-2016, he worked as a Senior Research Associate on computer vision in the School of Computer Science, University of Adelaide, Australia. His research interests include machine learning, sparse analysis, deep learning and large-scale optimization.
\end{IEEEbiography}

\vspace{-0.2in}
\begin{IEEEbiography}[{\includegraphics[width=1in,height=1.25in,clip,keepaspectratio]{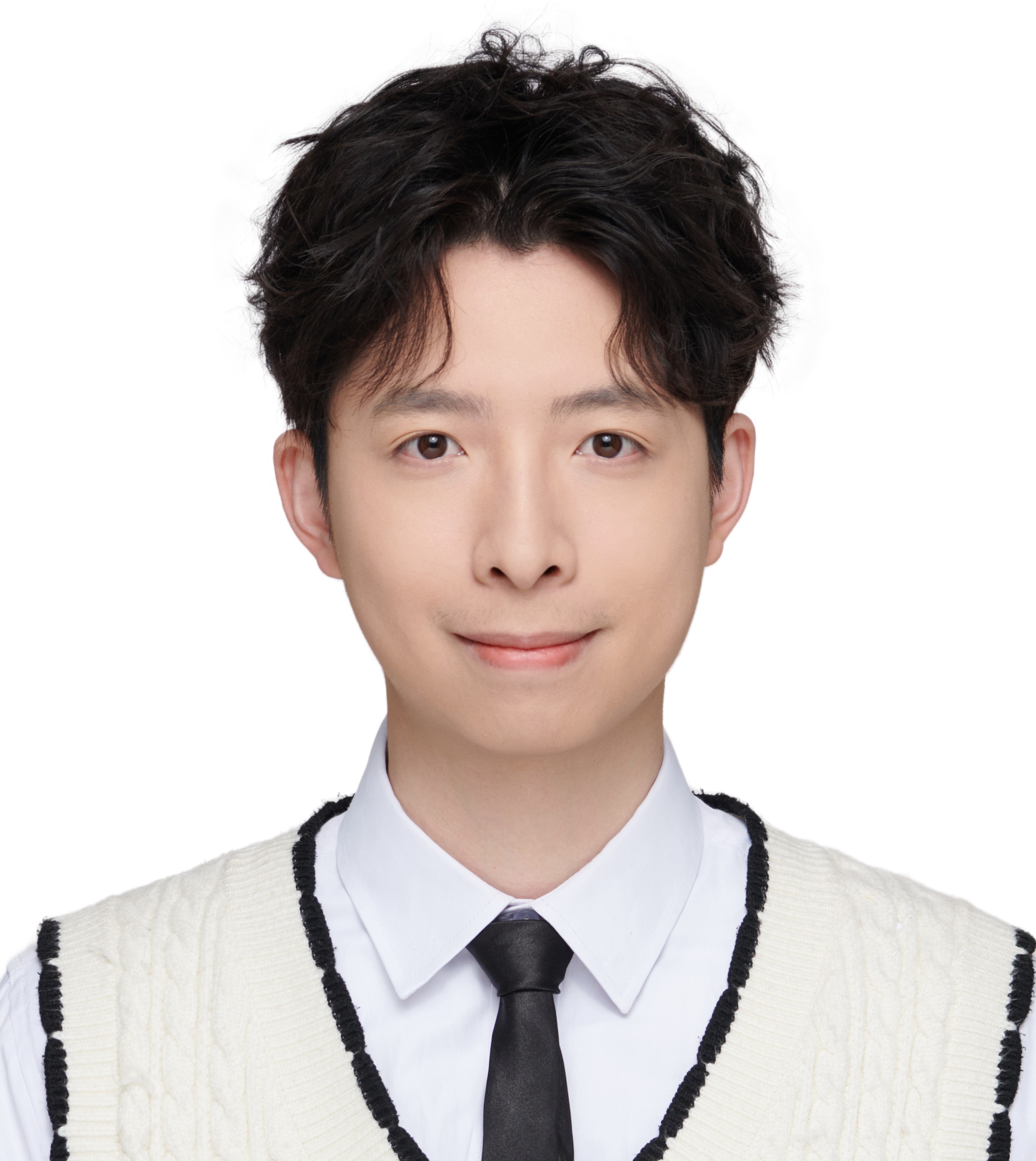}}]{Guohao Chen}
is a Master student in the School of Software Engineering at South China University of Technology. He received his Bachelor Degree in the School of Software Engineering in 2022 from South China University of Technology in Guangzhou, China. His research interests are broadly in machine learning and mainly focus on inference-time learning. He has published papers in top venues, including NeurIPS and ICML.
\end{IEEEbiography}

\begin{IEEEbiography}[{\includegraphics[width=1in,height=1.25in,clip,keepaspectratio]{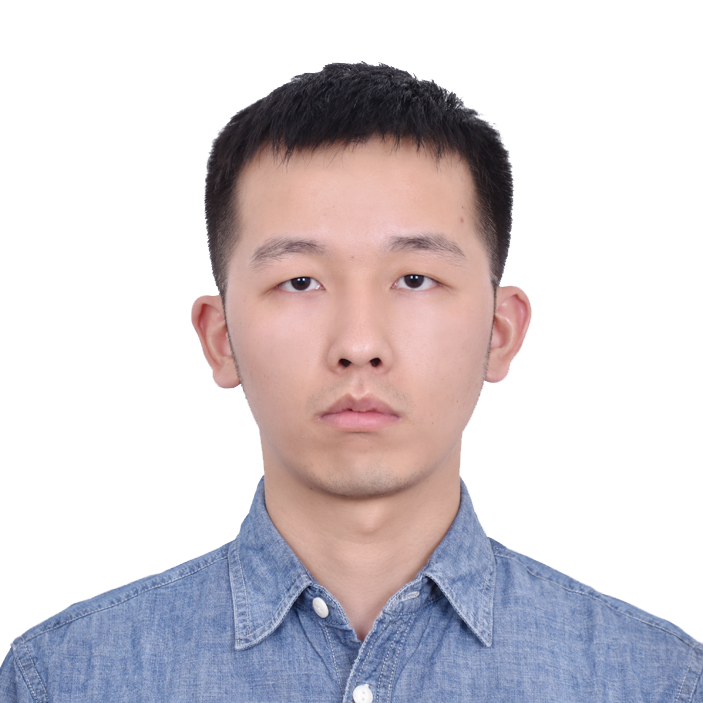}}]{Jiaxiang Wu} is currently a researcher at ByteDance, China. Previously, he has worked at Tencent AI Lab and XVERSE. He received the B.E. degree in automation from Beijing Institute of Technology, and the Ph.D. degree in computer science from Institute of Automation, Chinese Academy of Sciences. His research interests include model compression, neural architecture search, distributed optimization, and protein structure prediction. He has published papers in top venues, including JMLR, PNAS, ICML, NeurIPS, ICLR, CVPR, and AAAI. He has been invited as a reviewer for top-tier conferences and journals, including ICML, NeurIPS, ICLR, CVPR, AAAI, IJCAI, TPAMI, and TNNLS.
\end{IEEEbiography}

\vspace{-1in}
\begin{IEEEbiography}[{\includegraphics[width=1in,height=1.15in,clip,keepaspectratio]{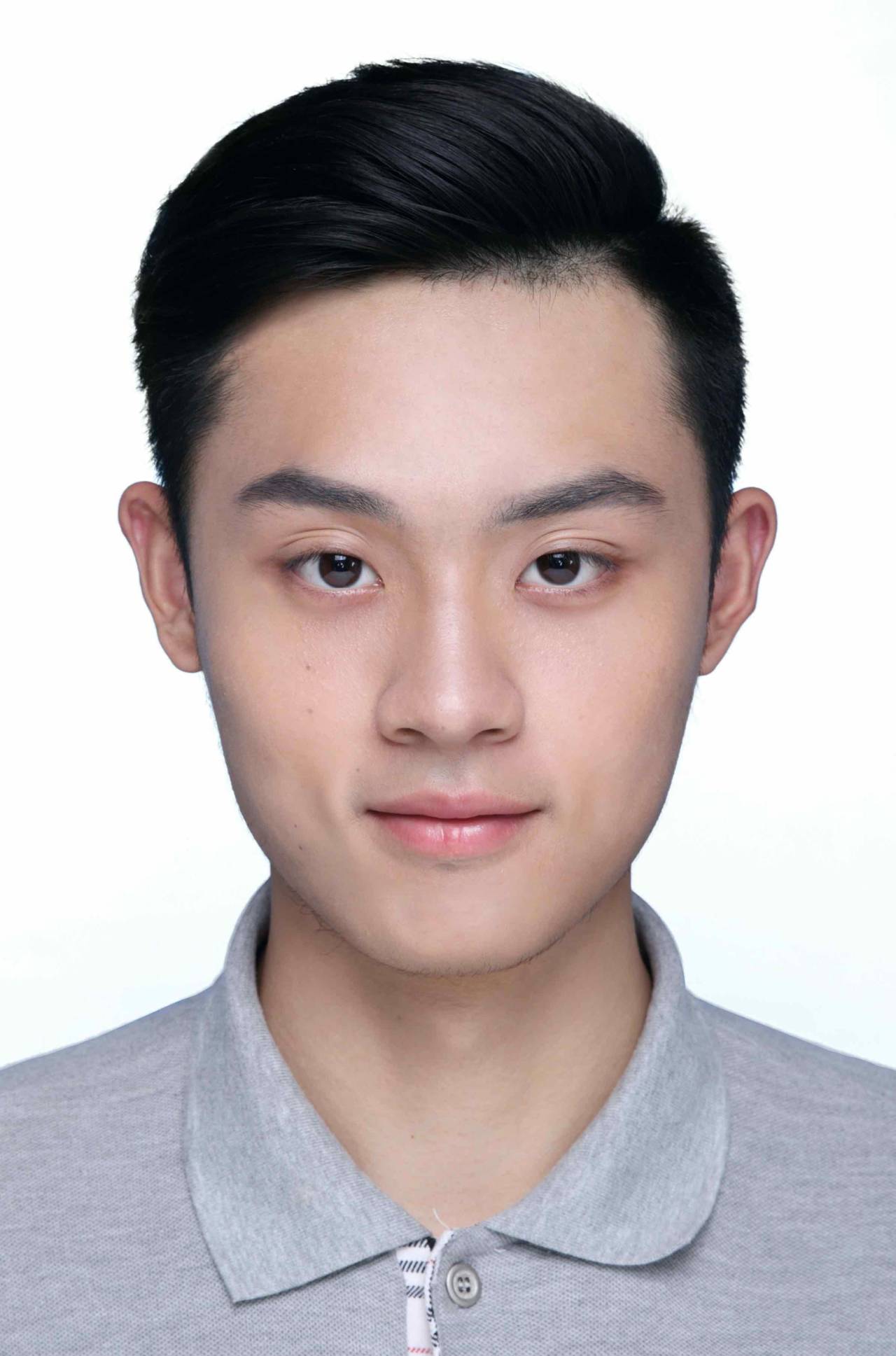}}]{Yifan Zhang}
obtained his Ph.D. degree in computer science at National University of Singapore. His research interests are broadly in machine learning, to solve domain shifts problems for deep learning. He has published papers in top venues, including NeurIPS, ICML, ICLR, CVPR, SIGKDD, ECCV, IJCAI, TIP, and TKDE. He has been invited as a reviewer for top-tier conferences and journals, including NeurIPS, ICML, ICLR, CVPR, ECCV, AAAI, IJCAI, TPAMI, TIP, IJCV, and TNNLS.
\end{IEEEbiography}

\vspace{-1in}
\begin{IEEEbiography}[{\includegraphics[width=1in,height=1.25in,clip,keepaspectratio]{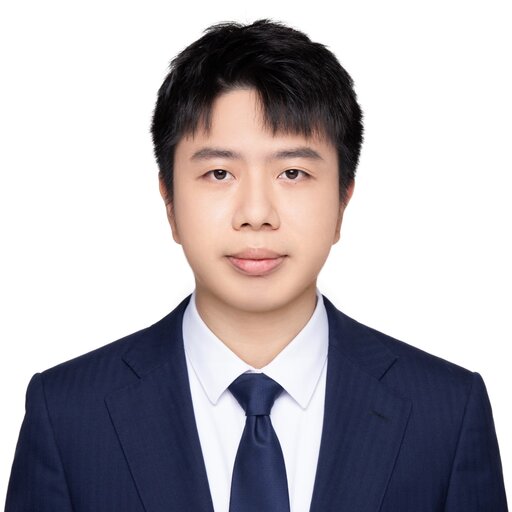}}]{Yaofo Chen} is recently a Post-doctoral Researcher with School of Future Technology at South China University of Technology. He received his Ph.D. degree in the School of Software Engineering in 2024 from South China University of Technology in Guangzhou, China. His research interests include neural architecture search and test-time adaptation. He has published papers in top venues, including ICML, ICLR, CVPR, AAAI, IEEE TCSVT and Neural Networks. He has been invited as a reviewer for top-tier conferences including ICLR, ICML, NeurIPS, CVPR, ICCV, ECCV and AAAI.
\end{IEEEbiography}

\vspace{-1in}
\begin{IEEEbiography}[{\includegraphics[width=1in,height=1.25in,clip,keepaspectratio]{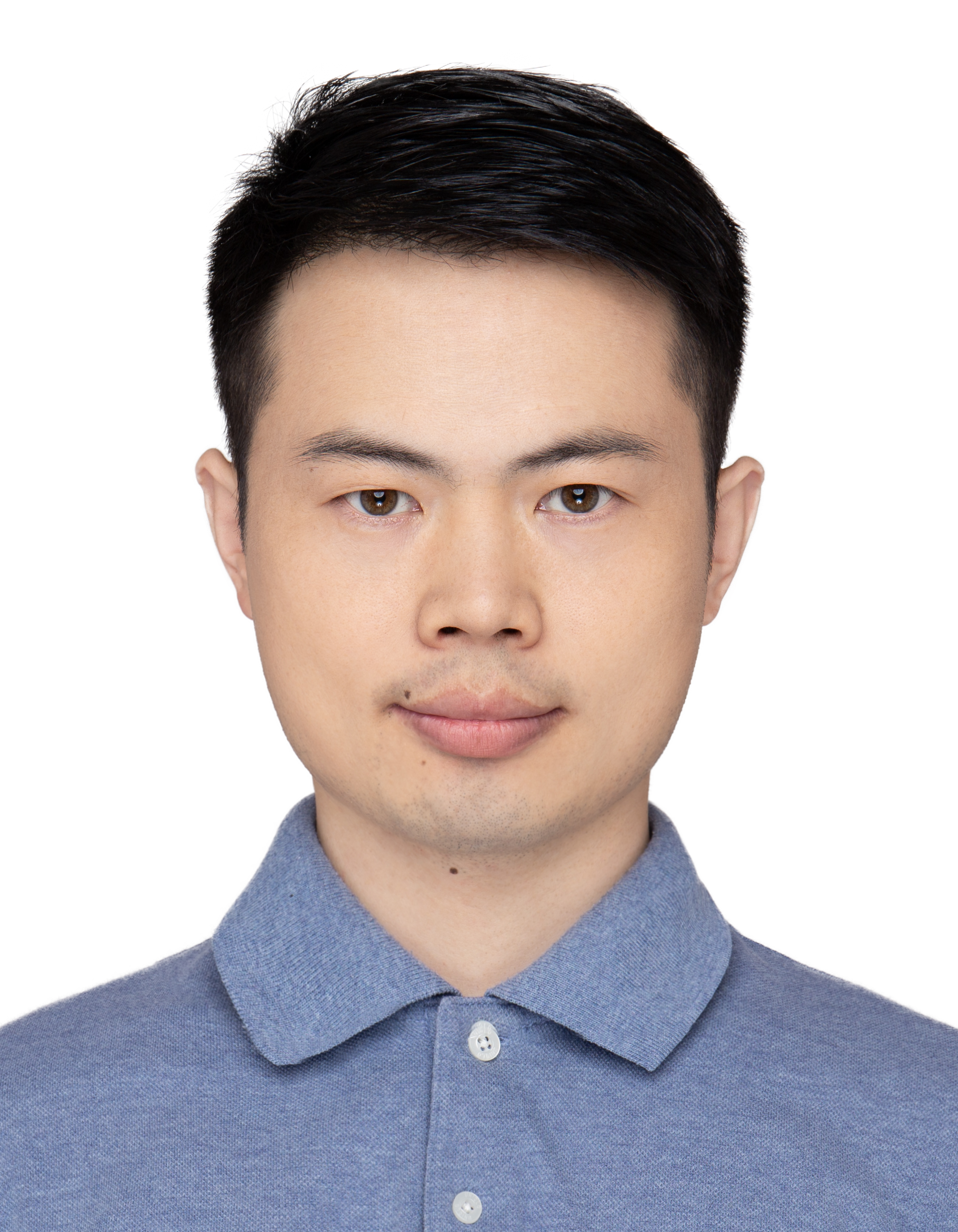}}]{Peilin Zhao} is currently a principal researcher at Tencent AI Lab in China. Previously, he worked at Rutgers University, A*STAR (Agency for Science, Technology and Research), and Ant Group. His research interests focused on machine learning and its applications. He has been invited to serve as area chair or associate editor at leading international conferences and journals such as ICML, TPAMI, etc. He received a bachelor's degree in mathematics from Zhejiang University, and a Ph.D. degree in computer science from Nanyang Technological University.
\end{IEEEbiography}

\vspace{-1in}
\begin{IEEEbiography}[{\includegraphics[width=1in,height=1.15in,clip,keepaspectratio]{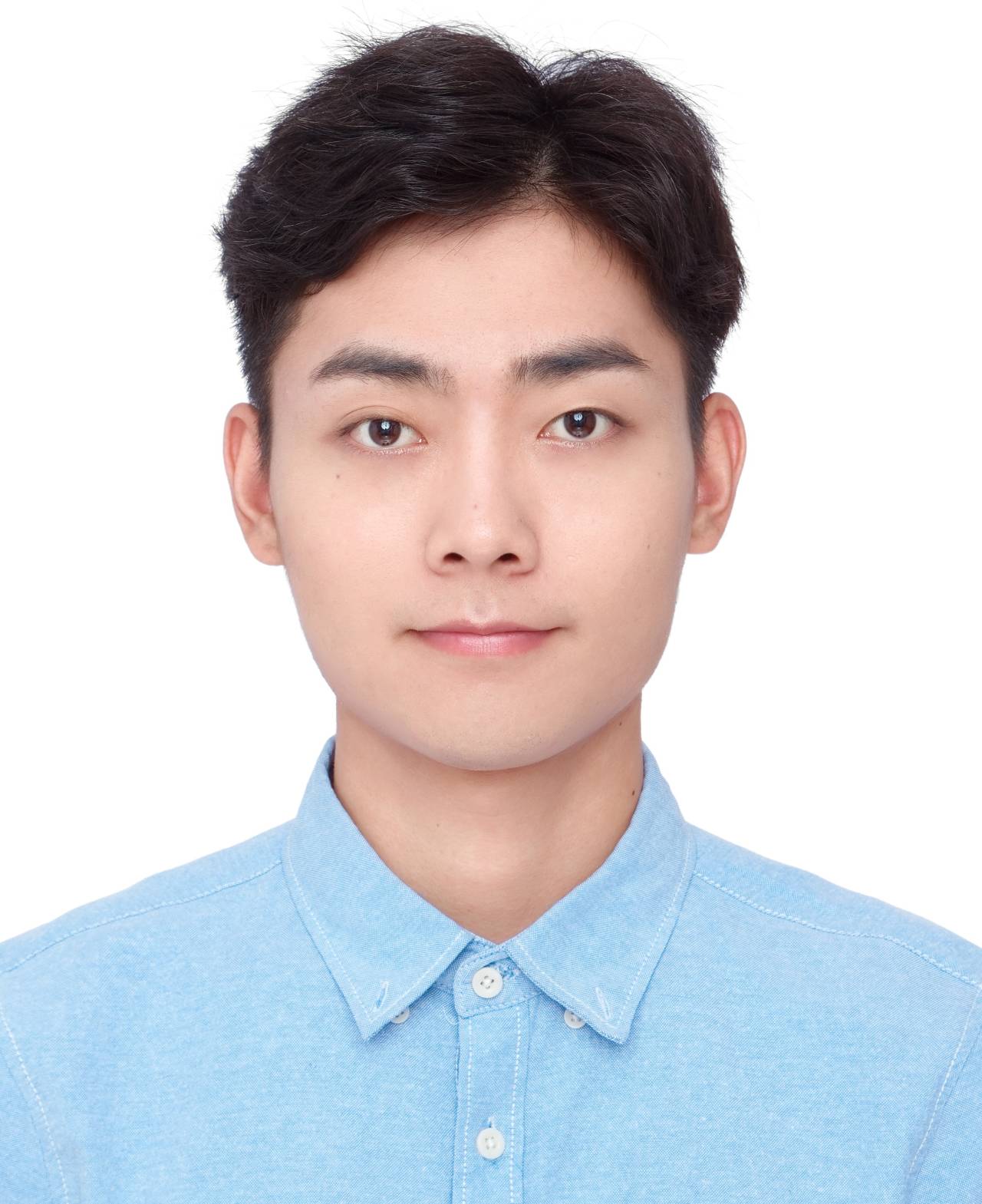}}]{Shuaicheng Niu}
 is currently a Research Fellow at Nanyang Technological University, Singapore. He received the Ph.D. degree from the South China University of Technology, China,  in 2023. His research interests are broadly in machine learning and mainly focus on test-time computing and automated machine learning. He has published papers in top venues, including ICML, ICLR, NeurIPS, CVPR, ECCV, IJCAI, AAAI, IEEE TPAMI, IEEE TIP, and IEEE TKDE. He has been invited as an area chair or reviewer for top-tier conferences and journals, including ICML, ICLR, NeurIPS, CVPR, ICCV, ECCV, IEEE TPAMI, IEEE TIP, IEEE TNNLS, and IJCV.
\end{IEEEbiography}

\newpage

\appendices
\onecolumn

\begin{LARGE}
~~~\vspace{1pt}
\begin{center}
    \bf Supplementary Materials for ``\mytitle''
\end{center}
\end{LARGE}
\vspace{2pt}

In the supplementary, we provide more implementation details and more experimental results of our \myeata.
We organize our supplementary as follows.

\begin{itemize}[leftmargin=*]
    \item In Section~\ref{supp:sec:more_impl}, we provide more details of our proposed \myeata and \myeataC.
    \item In Section~\ref{supp:sec:results_perf_efc}, we present additional experimental results and ablation studies to further demonstrate our superiority in out-of-distribution performance, calibration, efficiency, and stability.
    \item In Section~\ref{supp:sec:discussion}, we provide further insights, and empirical/visual evidence to validate the design choices in \myeataC.
\end{itemize}


\section{More Details of \myeata and \myeataC}\label{supp:sec:more_impl}
\subsection{Overall Design of \myeata}

\begin{figure*}[h]
\centering
\renewcommand\thefigure{A}
\includegraphics[width=1.0\linewidth]{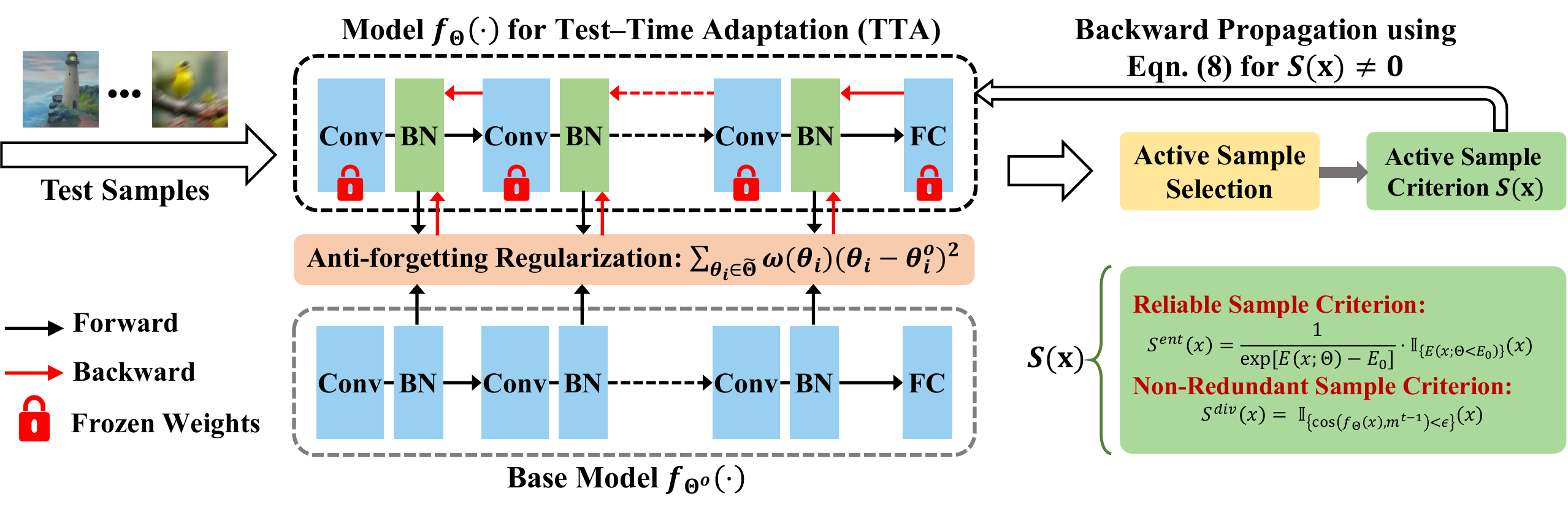}
\vspace{-0.15in}
\caption{An illustration of the proposed \myeata, which consists of a sample-efficient entropy minimization loss for test-time adaptation, and an anti-forgetting regularization to constrain important parameters from drastic change. }
\label{suppl:fig:eata overall}
\end{figure*}

We provide an overview of our \myeata. As shown in Figure~\ref{suppl:fig:eata overall}, given a trained base model $f_{\Theta^o}$, we perform test-time adaptation with a model $f_{\Theta}$ that is initialized from $\Theta^o$.
During the adaptation process, we only update the parameters of batch normalization layers in $f_{\Theta}$ and froze the rest parameters.
When a batch of test sample $\mX\small{=}\{\bx_b\}_{b=1}^{B}$ comes, we calculate a sample-adaptive weight $S(\bx)$ for each test sample to identify whether the sample is active for adaptation or not.
We only perform backward propagation with the samples whose $S(\bx) \neq 0$.
Moreover, we propose an anti-forgetting regularizer to prevent the model parameters $\Theta$ changing too much from $\Theta^o$.

\subsection{More Details on Datasets}\label{supp:sec:more_datasets}

Following the settings of Tent~\cite{wang2021tent} and MEMO~\cite{zhang2021memo}, we conduct experiments on three benchmark datasets for out-of-distribution generalization, \ie, ImageNet-C~\cite{hendrycks2019benchmarking}, ImageNet-R~\cite{hendrycks2021many}, and ACDC~\cite{SDV21acdc}.

\noindent\textbf{ImageNet-C} consists of various versions of corruption applied to 50,000 validation images from ImageNet. The dataset encompasses 15 distinct corruption types of 4 main categories, including Gaussian noise, shot noise, impulse noise, defocus blur, glass blur, motion blur, zoom blur, snow, frost, fog, brightness, contrast, elastic transformation, pixelation, and JPEG compression. Each corruption type is characterized by 5 different levels of severity, with higher severity levels indicating a more severe distribution shift.

\noindent\textbf{ImageNet-R} contains 30,000 images with various artistic renditions of 200 ImageNet classes. These images are primarily collected from Flickr and filtered by Amazon MTurk annotators.

\noindent\textbf{ACDC} contains four categories of images collected in adverse conditions, including 
\textit{fog}, \textit{night}, \textit{rain}, and \textit{snow}. Following CoTTA~\cite{wang2022cotta}, we use 400 unlabeled images from each adverse condition for continuous test-time adaptation.

\subsection{More Experimental Protocols on Evaluation}
Following Tent~\cite{wang2021tent} and CoTTA~\cite{wang2022cotta},  we use ResNet-50 and ViT-Base for ImageNet experiments, and Segformer-B5 for ACDC experiments. In classification experiments, the models are trained on the original ImageNet training set and then tested on clean or the aforementioned OOD test sets. In semantic segmentation experiments, the model is trained on Cityscapes and then tested on ACDC. For a fair comparison, the parameters of ViTBase and Segformer-B5 are directly obtained from the \textit{timm}\footnote{\href{https://github.com/pprp/timm}{https://github.com/pprp/timm}} and Segformer\footnote{\href{https://github.com/NVlabs/SegFormer}{https://github.com/NVlabs/SegFormer}}~\cite{xie2021segformer}, respectively. ResNet-50 is trained via the official code in the \textit{torchvision}\footnote{\href{https://github.com/pytorch/vision}{https://github.com/pytorch/vision}} library with stochastic depth.

\noindent\textbf{Our \myeata and \myeataC.} We employ SGD with a momentum of 0.9 and a batch size of 64 for test time adaptation. 
For \myeata, the learning rate is set to 0.00025/0.001 for ResNet-50/ViTBase on ImageNet, and 7.5$\times10^{-5}$ on ACDC, following SAR and CoTTA. The entropy constant $E_0$ in Eqn.~(\ref{eq:lambda_1st}) is set to $0.4\times\ln C$ in classification experiments and $0.1\times\ln C$ in segmentation, where $C$ is the number of classes. We fix the similarity threshold $\epsilon$ in Eqn.~(\ref{eq:lambda_2nd}) to 0.05 for all experiments. The trade-off parameter $\beta$ in Eqn.~(\ref{eq:overall_loss}) is set to 2,000/500 for ImageNet/ACDC to improve adaptation stability by mitigating forgetting;
For \myeataC, the learning rate is set to 0.005/0.1 for ResNet-50/ViTBase on ImageNet, and 0.0005 on ACDC. The sub-network for model uncertainty estimation is obtained via stochastic depth~\cite{huang2016depth} with a drop ratio of 0.2/0.6 for ImageNet/ACDC. The entropy constant $E_0$ in Eqn.~(\ref{eq:lambda_1st}) is set to 0.4/0.5$\times\ln C$, and the $\epsilon$ in Eqn.~(\ref{eq:lambda_2nd}) is set to 0.05/0.07 for ViTBase/ResNet-50 on ImageNet. In Eqn.~(\ref{eq:eata-c objective}), we set $\beta$ to 50/500 on ImageNet/ACDC and $\alpha$ to 0.1 on ImageNet. Note that for semantic segmentation, our \myeataC does not perform min-max entropy regularization and active sample selection due to the problem of long-tailed distribution~\cite{wang2022semi,pan2021model}. 
For both \myeata and \myeataC, the moving average factor $\alpha$ in Eqn.~(\ref{eq:moving_average}) is set to 0.1 following established practice, and we use 2,000/20 samples for ImageNet/ACDC to calculate $\omega(\theta_i)$ in Eqn.~(\ref{eq:fisher_information}).

\noindent\textbf{Compared Methods.} For BN adaptation~\cite{schneider2020improving}, MEMO~\cite{zhang2021memo}, CoTTA~\cite{wang2022cotta}, SAR~\cite{niu2023sar}, ROID~\cite{marsden2024universal}, Rdump~\cite{press2024rdumb}, TEA~\cite{yuan2024tea}, and DAT~\cite{ni2024distribution}, the hyper-parameters follow their original papers or MEMO.
Specifically, for BN adaptation~\cite{schneider2020improving}, both the batch size $B$ and prior strength $N$ are set to 64. The learning rate in CoTTA~\cite{wang2022cotta} for ViTBase is set to 0.005, and the augmentation threshold $p_{th}$ is set to 0.1. TEA~\cite{yuan2024tea} uses SGD with a learning rate of 0.00025/0.005 for ResNet-50/ViT-Base, respectively. Other hyper-parameter settings of CoTTA and TEA, as well as those for MEMO, SAR, ROID, Rdump, and DAT, can be found in their original paper.
For Tent~\cite{wang2021tent}, we use SGD with a momentum of 0.9. The batch size is 64 for both ImageNet experiments and the learning rate is set to 0.00025/0.001 for ResNet-50/ViTBase, respectively. Note that the hyper-parameters of Tent are the same as our \myeata for a fair comparison.

\section{More Experimental Results}\label{supp:sec:results_perf_efc}
\subsection{Results under Single-Domain Adaptation}
\vspace{-0.05in}
In Table~\ref{supp:tab:imagenet-c-level-5}, we provide additional results to evaluate the effectiveness of our \myeata and \myeataC under single-domain adaptation scenarios. From the results, while the compared methods exhibit stronger performance in the less challenging adaptation setting, our \myeata and \myeataC still demonstrate substantial superiority. Specifically, by filtering the unreliable and redundant samples from adaptation, our \myeata achieves substantial accuracy improvement while requiring significantly fewer backpropagations, \eg, on ViT-Base, average accuracy increases from 55.5\% (SAR) to 61.8\% (\myeata) with a reduction in backward passes from 72,446 to 32,524. Meanwhile, \myeataC further enhances performance and reduces calibration error on both ResNet-50 and ViTBase. These results are consistent with our findings in lifelong adaptation (Table~\ref{tab:imagenet-c-level-5}), verifying the effectiveness of our \myeata and \myeataC across diverse adaptation scenarios.

\vspace{-0.15in}
\subsection{More Results on Semantic Segmentation}
\vspace{-0.05in}
In Table~\ref{tab:suppl:acdc}, we present the complete results for lifelong test-time adaptation on the ACDC dataset. Although lifelong adaptation offers exposure to a broader range of data with the potential for improved performance, Tent and DAT tend to suffer from error accumulation and catastrophic forgetting, leading to a decline in subsequent adaptations. In contrast, our \myeata mitigates these issues by excluding unreliable gradients and constraining important weights from drastic changes, thereby maintaining a more stable performance. Moreover, our \myeataC is able to accumulate learned knowledge from prior adaptation and showcase consistent improvement over lifelong adaptation, increasing the average mIOU on four adverse datasets from 59.8\% (first round) to 62.3\% (tenth round). These results further highlight the long-term effectiveness of our methods in leveraging the potential of lifelong test-time adaptation.

\vspace{-0.15in}
\subsection{Results under Diverse Severity Levels}
\begin{figure*}[t]
\vspace{-0.1in}
\centering
\renewcommand\thefigure{D}
\includegraphics[width=.85\linewidth]{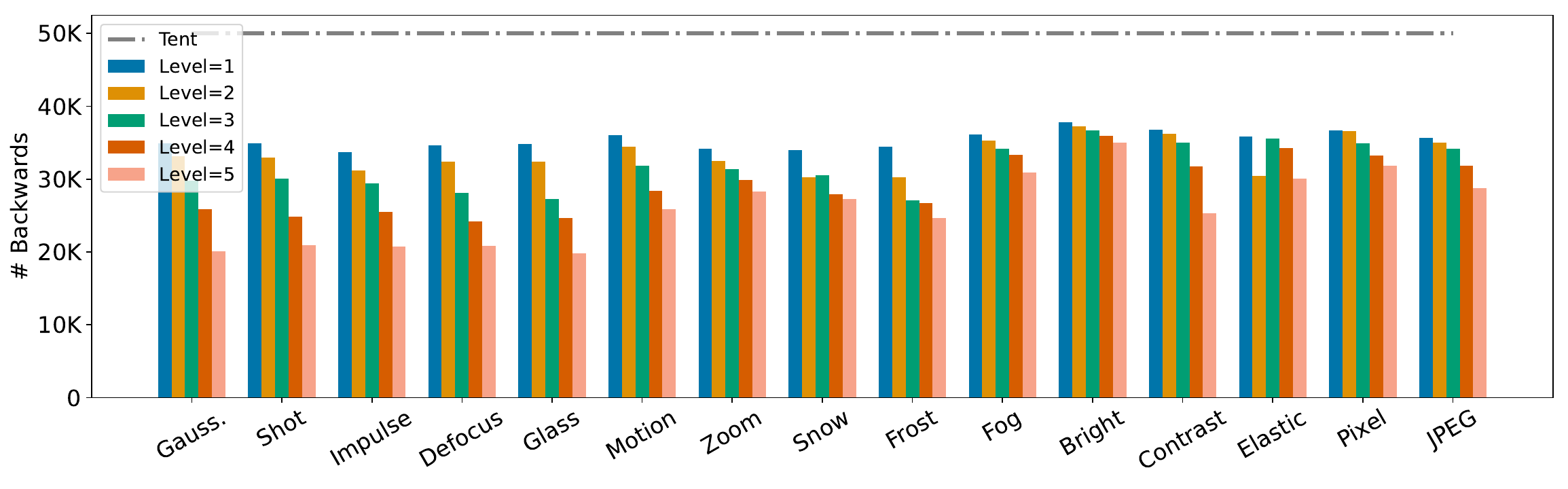}
\vspace{-0.2in}
\caption{Comparison between \myeta and Tent w.r.t. the number of backward passes on ImageNet-C with different corruption types and severity levels.}
\label{fig:imageC-num-backwards2}
\end{figure*}

\vspace{-0.05in}
In Figure~\ref{fig:imageC-num-backwards2}, we show the number of backward propagation of our \myeta on ImageNet-C with different corruption types and severity levels. Across various corruption types, our \myeta shows great superiority over existing methods in terms of adaptation efficiency. Compared with MEMO (50,000$\times$64) and Tent (50,000), our \myeta only requires 31,741 backward passes (averaged over 15 corruption types) when the severity level is set to 3. The reason is that we exclude some unreliable and redundant test samples out of test-time optimization. In this case, we only need to perform backward computation on those remaining test samples, leading to improved efficiency.

\vspace{-0.15in}
\subsection{More Results on Prevent Forgetting}\label{supp:sec:results_prev_forget}
\begin{figure}[t]
\centering     
\renewcommand\thefigure{F}
\subfigure{\label{fig:imageC-forgetting-level-1}\includegraphics[width=1.0\columnwidth]{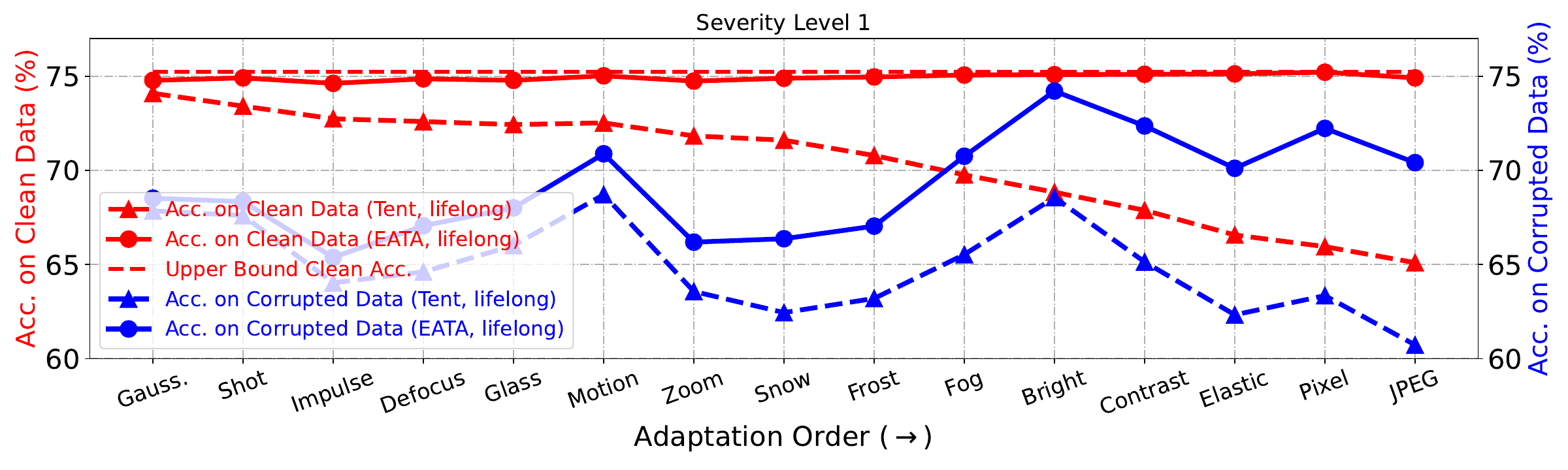}}
\subfigure{\label{fig:imageC-forgetting-level-2}\includegraphics[width=1.0\columnwidth]{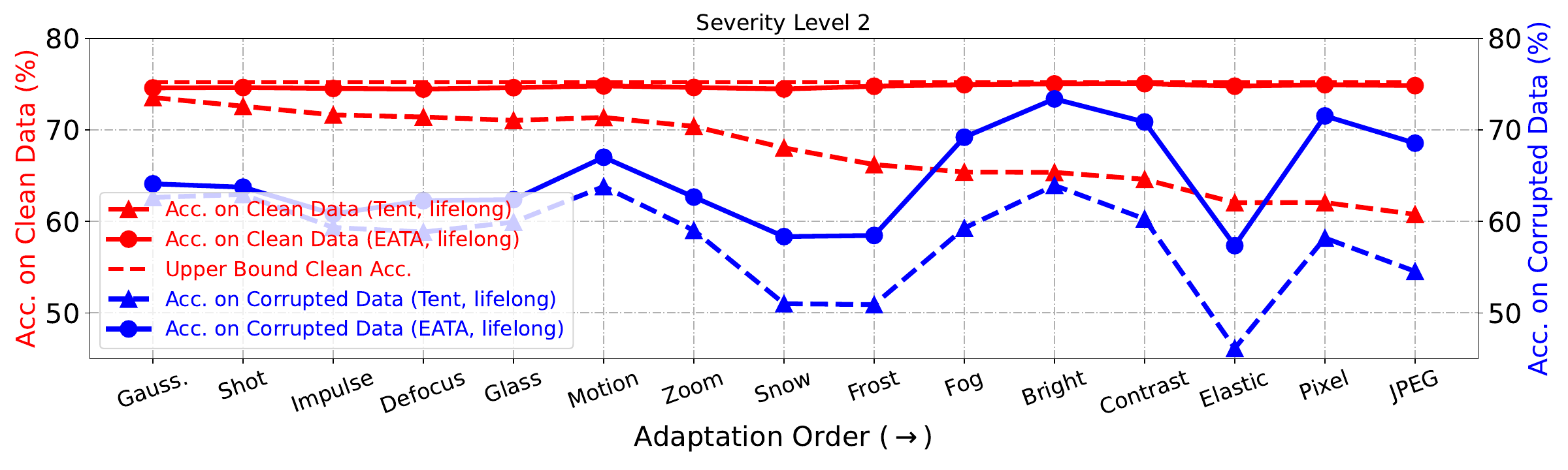}}
\subfigure{\label{fig:imageC-forgetting-level-3}\includegraphics[width=1.0\columnwidth]{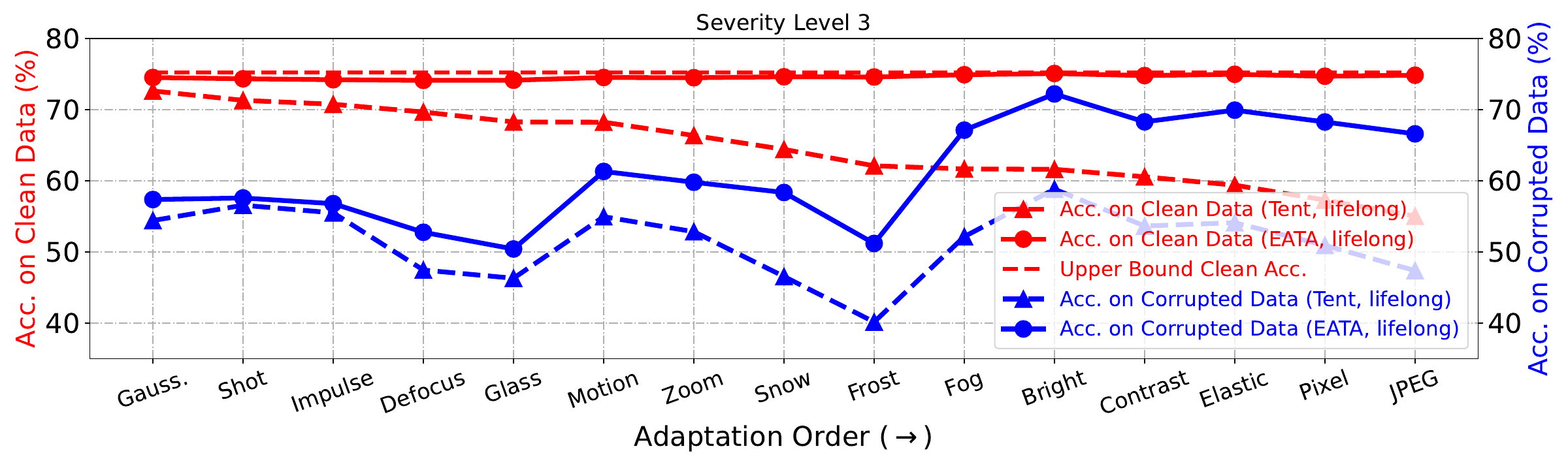}}
\subfigure{\label{fig:imageC-forgetting-level-4}\includegraphics[width=1.0\columnwidth]{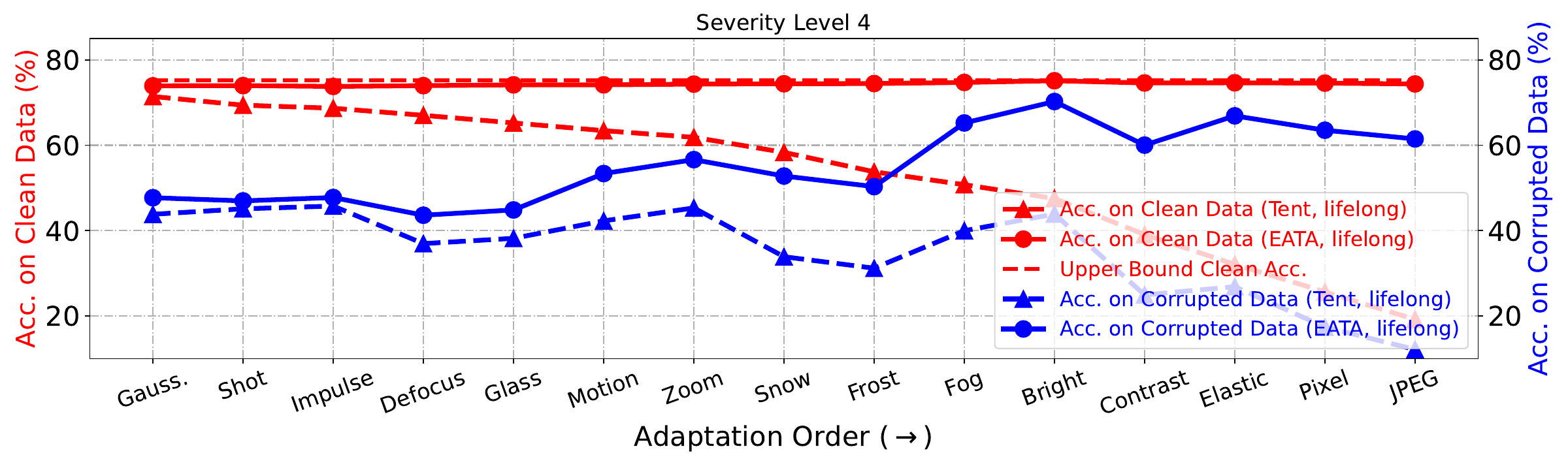}}
\vspace{-0.2in}
\caption{Comparison of preventing forgetting on ImageNet-C (severity levels 1-4) with ResNet-50. We record the OOD corruption accuracy on each corrupted test set and the associated ID clean accuracy (after OOD adaptation).  The model performs \textbf{lifelong adaptation}, in which the model parameters will never be reset.}
\label{fig:imageC-forgetting-levels}
\end{figure}

\begin{figure}[t]
\centering     
\renewcommand\thefigure{G}
\subfigure{\label{fig:imageC-forgetting-level-1-each-reset}\includegraphics[width=1.0\columnwidth]{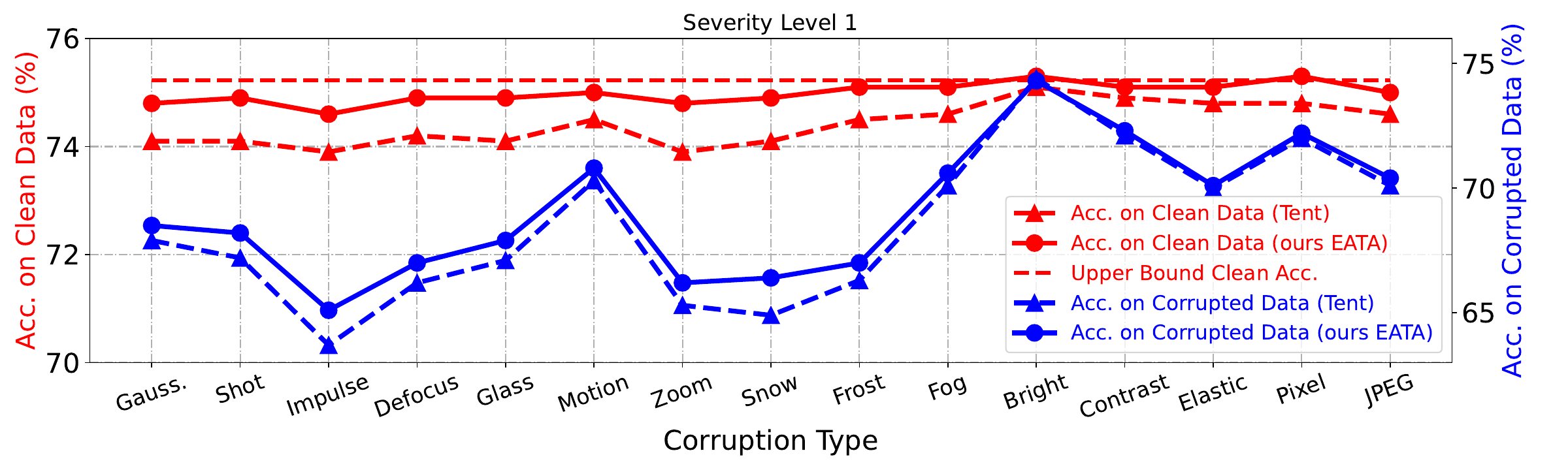}}
\subfigure{\label{fig:imageC-forgetting-level-2-each-reset}\includegraphics[width=1.0\columnwidth]{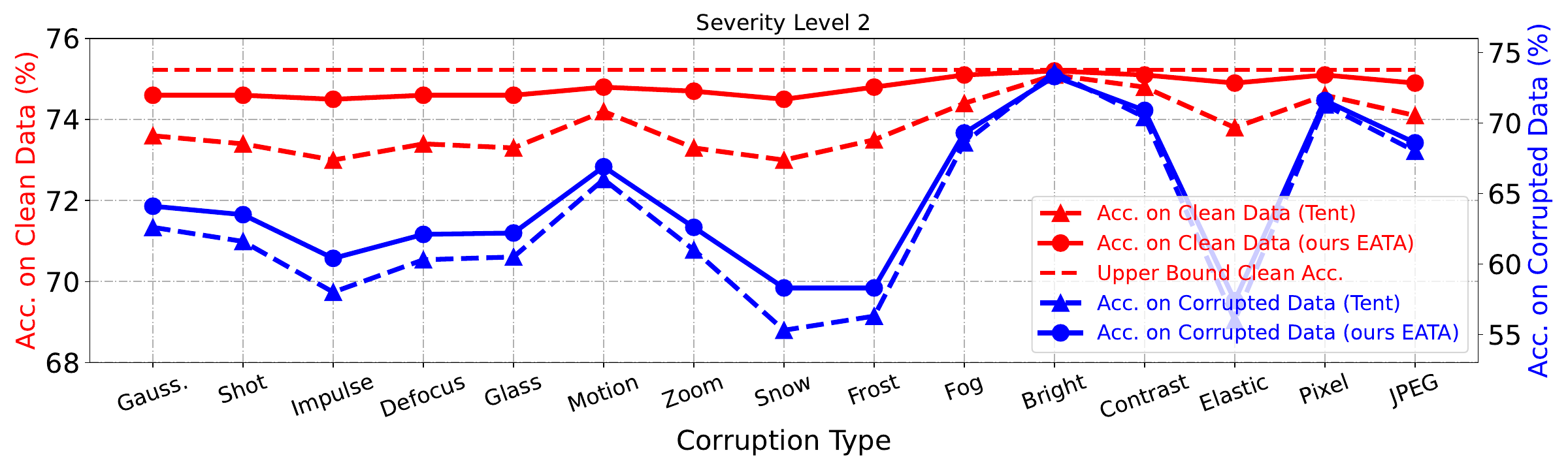}}
\subfigure{\label{fig:imageC-forgetting-level-3-each-reset}\includegraphics[width=1.0\columnwidth]{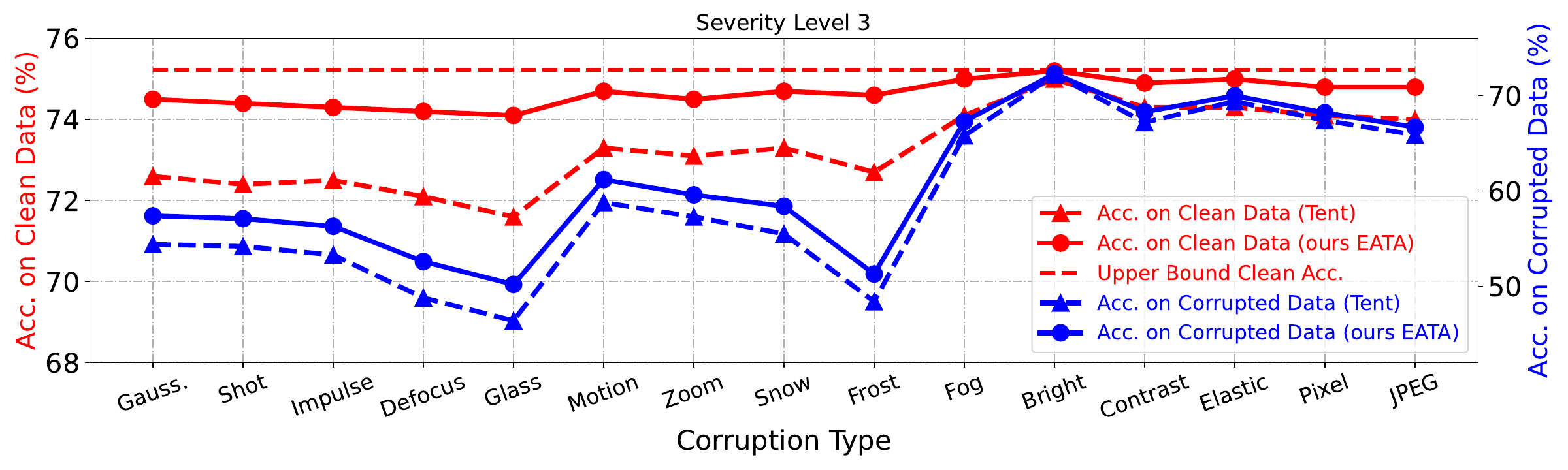}}
\subfigure{\label{fig:imageC-forgetting-level-4-each-reset}\includegraphics[width=1.0\columnwidth]{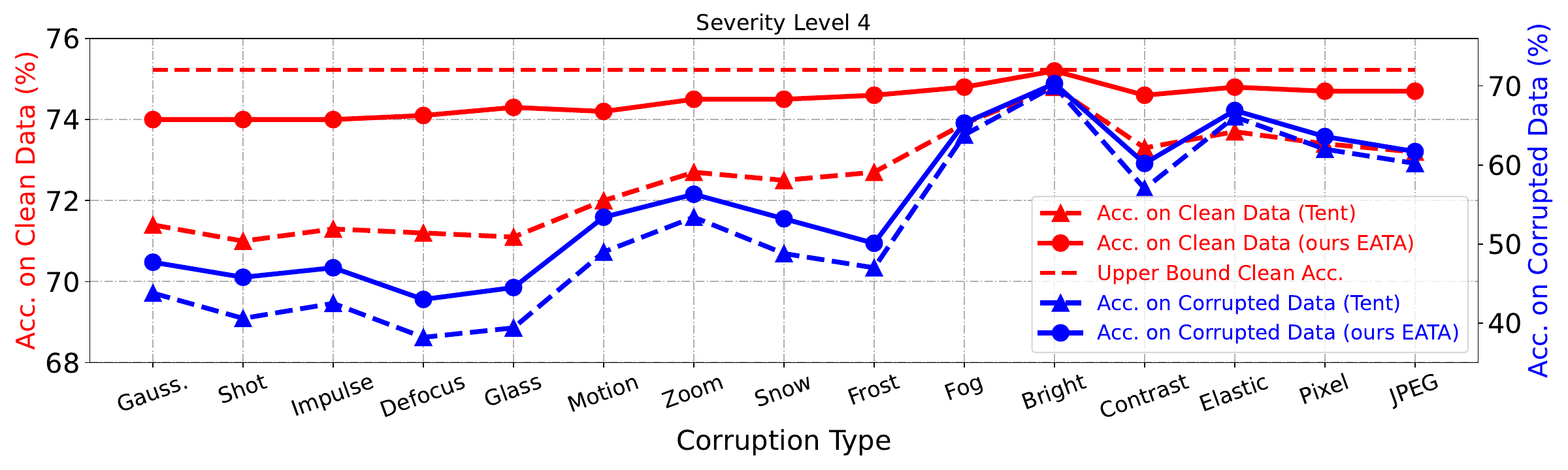}}
\vspace{-0.2in}
\caption{Comparisons of preventing forgetting on ImageNet-C (severity levels 1-4) with ResNet-50. We record the OOD corruption accuracy on each corrupted test set and the associated ID clean accuracy (after OOD adaptation). The model performs \textbf{single-domain adaptation}, where model parameters of both Tent and our \myeata are reset before adapting to a new corruption type.}
\label{fig:imageC-forgetting-levels-each-reset}
\end{figure}

We provide more results to demonstrate the effectiveness of our \myeata in preventing forgetting. We report the comparison results of \myeata (lifelong) \vs~Tent (lifelong) and \myeata \vs~Tent in Figures~\ref{fig:imageC-forgetting-levels} and~\ref{fig:imageC-forgetting-levels-each-reset}, respectively.  In the lifelong adaptation scenario, Tent suffers more severe ID performance degradation than that of reset adaptation (\ie, Figure~\ref{fig:imageC-forgetting-levels-each-reset}), showing that the more optimization steps, the more severe forgetting. Moreover, with the increase of the severity level, the ID clean accuracy degradation of Tent increases accordingly. This result indicates that the OOD adaptation with more severe distribution shifts will result in more severe forgetting. In contrast, our methods achieve higher OOD corruption accuracy and, meanwhile maintain the ID clean accuracy (competitive to the original accuracy that is tested before any OOD adaptation) in both two adaptation scenarios (reset and lifelong). These results are consistent with those in the main paper and further demonstrate the effectiveness of our proposed anti-forgetting weight regularization.

\vspace{-0.15in}
\subsection{More Ablation Studies on \myeataC's Components}\label{supp:sec:ablation}
\myeataC aims to achieve a favorable balance among accuracy, calibration, and efficiency. We supplement more ablation studies on different datasets and architectures to verify the effectiveness of each component, as in Table~\ref{suppl:tab:ablation-RC}. Our findings on ImageNet-R with ResNet-50 are consistent with those in Table~\ref{tab:eata-c_ablation}, including 1)~Using the consistency loss alone already yields a strong baseline; 2)~Adding the entropy regularizer and Fisher regularizer further improves TTA accuracy while reducing ECE; 3)~By filtering out unreliable and redundant samples, active sample selection significantly boosts computational efficiency while maintaining or improving accuracy.

Moreover, we provide further explanations regarding \textbf{the performance variations between ViT-Base and ResNet-50} when incorporating active sample selection. This is primarily attributed to the quality of pseudo labels for TTA. ViT-Base generally yields higher-quality pseudo labels, with consistently higher accuracy and lower ECE. Considering adaptation efficiency, the active sample selection strategy may remove a few instances that could have been beneficial to adaptation, resulting in a small drop in accuracy (\eg, from 66.4\% to 65.9\%), while reducing the required backward passes by over 35\%. In contrast, ResNet-50 tends to produce less stable pseudo labels, which could adversely affect the performance if all samples are used indiscriminately, as evidenced by the collapse of TEA~\cite{yuan2024tea} on ResNet in Table~\ref{tab:imagenet-c-level-5}. In this case, active sample selection not only reduces unnecessary backward passes, but also enhances accuracy by focusing test-time adaptation on more trustworthy samples.

\vspace{-0.05in}

\section{Additional Discussions}\label{supp:sec:discussion}

\subsection{Foundations of Disagreement-Based Data Uncertainty Indicator}

\begin{figure*}[t]
    \centering
        \begin{minipage}[t]{0.45\linewidth}
        \centering
        \renewcommand\thefigure{B}
        \includegraphics[width=.83\linewidth]{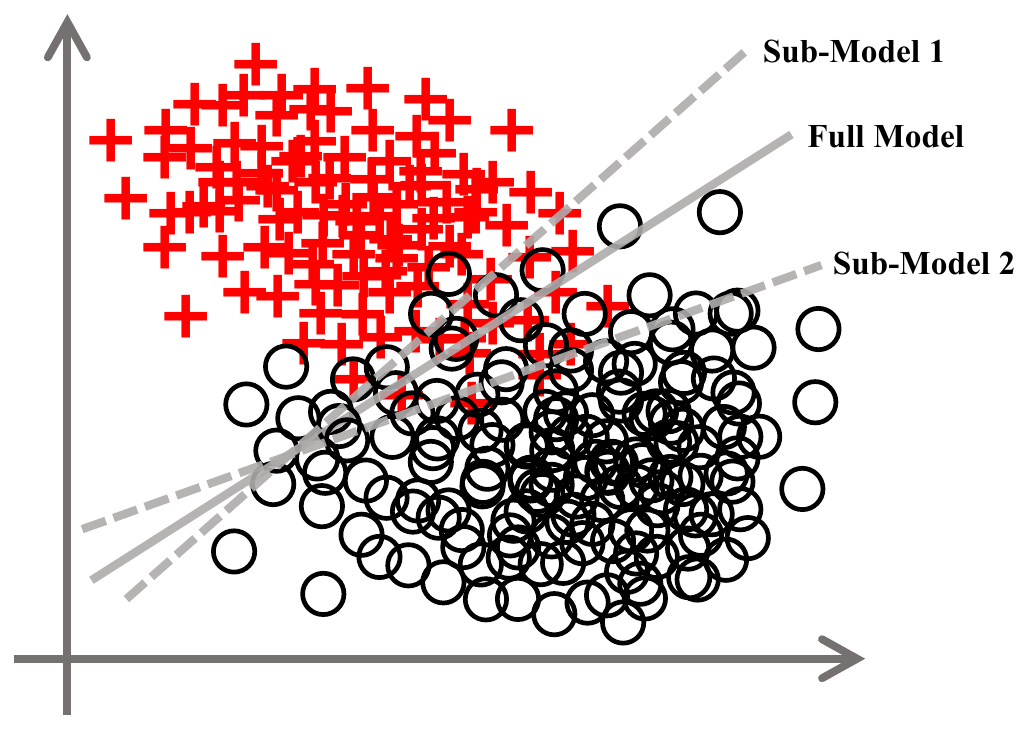}
        \caption{Toy illustration. Uncertain data near the decision boundary tend to cause disagreed predictions among the sub- and full model.}
        \label{fig:toy_figure}
    \end{minipage}
    \hfill
    \begin{minipage}[t]{0.48\linewidth}
        \centering
        \renewcommand\thefigure{C}
        \includegraphics[width=.68\linewidth]{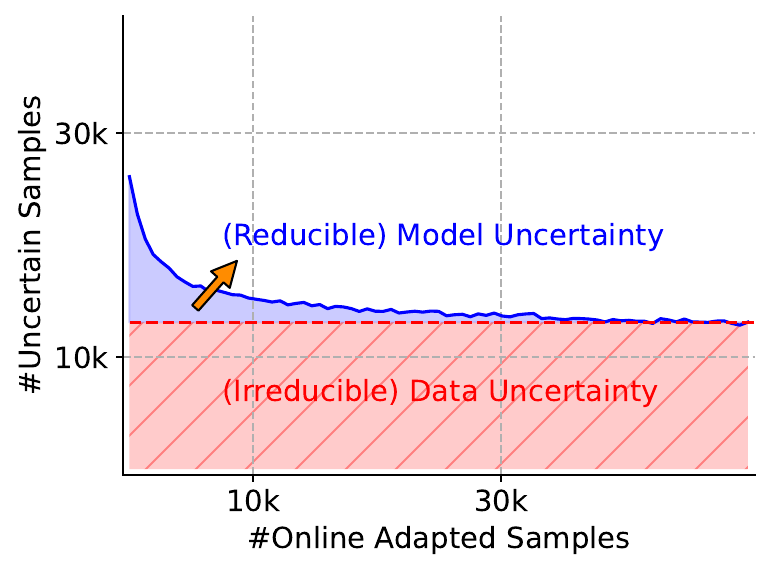}
        \caption{Evolution of uncertain samples (\ie, with disagreed predictions) during adaptation on ImageNet-C (Gauss, level 5) using Eqn.~(10).}
        \label{fig:uncertainty_comparison}
    \end{minipage}
    \vspace{-0.05in}
\end{figure*}

In EATA-C, we devise a disagreement-based indicator to identify uncertain samples. This is inspired by \textbf{margin-based theories in machine learning}~\cite{balcan2007margin,khan2019striking}, which indicate that samples near decision boundaries are inherently more uncertain and have been well justified with theoretical guarantees. To be specific, we leverage prediction disagreement between the full network and sub-network to detect these uncertain boundary samples, where conflicting predictions are more likely to occur. Figure~\ref{fig:toy_figure} provides a toy illustration of this principle and we depict our model and data uncertainty based on Figure~\ref{fig:toy_figure} below. 
\textbf{\textit{1) Model uncertainty.}} In our method, activating only a subset of parameters results in a new decision boundary for the sub-model, \eg, Sub-Model 1/2. Our consistency loss aligns the sub-model boundary with that of the full model, thereby aiming to minimize the model uncertainty and obtain an optimal decision boundary. 
\textbf{\textit{2) Data uncertainty.}} Unlike our consistency loss that measures model uncertainty, in which samples across the entire data space may yield low consistency loss, data uncertainty is reflected more prominently through the disagreement of predictions near the decision boundary.

Moreover, we provide \textbf{empirical evidence} of using prediction disagreement as the data uncertainty indicator in Table~\ref{tab:uncertain_indicator_accuracy} and Figure~\ref{fig:uncertainty_comparison}. From the results, we reveal that a substantial number of samples (over 13,000) exhibit persistent prediction disagreement even after extensive adaptation. Notably, about 90\% of these samples are misclassified by the well-trained model, suggesting intrinsic noise or ambiguity near decision boundaries, aligned with the definition of data uncertainty. Collectively, the margin-based theories and our empirical results provide a sound basis for adopting disagreement as a practical indicator of data uncertainty in TTA.

\subsection{Parameter Update Strategies: Norm-Only \textit{vs.} Full Updates}
In our main experiments, \myeataC updates only the normalization layers by following established practices of existing TTA work~\cite{wang2021tent, yuan2024tea}. Here, we further validate this design choice by comparing it against a variant that updates all model parameters, namely \myeataC$^\dagger$. The results in Table~\ref{tab:norm_or_all_parameters} highlight the following. 1)~\textbf{\textit{Reduced Network Divergence}}: Updating the normalization layer effectively decreases the divergence loss between the full network and the sub-network, from 0.565 (source model) to 0.451; 2)~\textbf{\textit{Improved Accuracy and Calibration}}: While updating all model parameters yields a marginally lower divergence, our \myeataC that updates only the norm layers leads to higher accuracy (56.8\% \textit{vs.} 56.0\%) and lower ECE (5.2\% \textit{vs.} 5.9\%). This improvement is mainly attributed to the increased stability of TTA, where restricting updates to the normalization layers acts as a regularizer for online adaptation and reduces the risk of forgetting and overfitting by keeping the majority of parameters frozen, also known as parameter-efficient tuning. 3)~\textbf{\textit{Lower Overhead}}: Our \myeataC also demonstrates reduced adaptation time (114.9s vs.\ 140.9s) and memory usage (5786.6MB vs.\ 9920.5MB) compared to EATA-C$^\dagger$, making it more practical for real-world applications.

\subsection{Visualization of Reliable and Unreliable Test Samples}
\begin{figure*}[t]
    \centering
    \renewcommand\thefigure{E}
    \includegraphics[width=.8\linewidth]{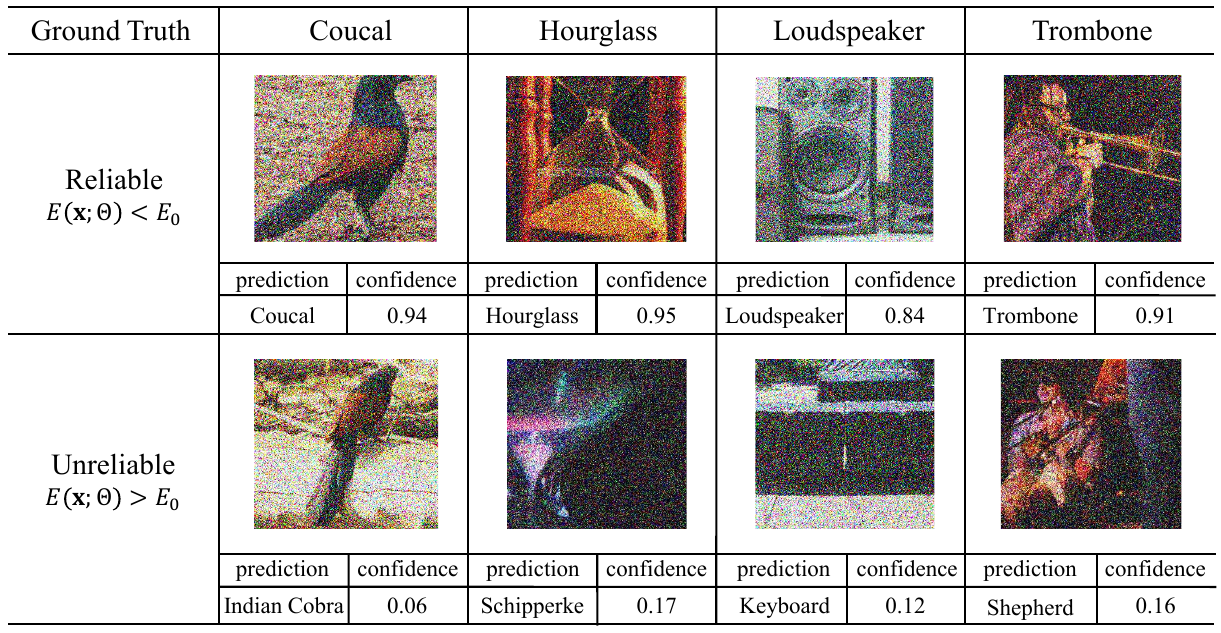}
    \vspace{-0.1in}
    \caption{Visualization of selected (reliable) and removed (unreliable) test samples for adaptation according to Eqn.~(\ref{eq:lambda_1st}). Predicition confidence is the highest class probability. Results are obtained on ImageNet-C(Gaussian, level 5) using the source ViT-Base model.}
    \label{suppl:fig:selected_removed_samples}
\end{figure*}

We show some selected and removed test samples for TTA according to our entropy-based indicator in Eqn.~(\ref{eq:lambda_1st}). From Figure~\ref{suppl:fig:selected_removed_samples}, the selected samples generally exhibit higher image quality (\ie, with clear and distinguishable class features), where the model can generate more reliable pseudo labels for adaptation. In contrast, the removed samples are often ambiguous even for human interpretation, leading to potentially incorrect predictions that would introduce noisy learning signals when used for adaptation. This highlights the effectiveness of our entropy-based identification in improving the robustness of TTA.

\begin{table*}[t]
    \renewcommand\thetable{A}
    \caption{Comparison with state-of-the-art methods on ImageNet-C with the highest severity level 5 regarding \textbf{Corruption Accuracy(\%, $\uparrow$)} and \textbf{Expected Calibration Error(\%,$\downarrow$)}. ``BN" and ``LN" denote batch and layer normalization. All results are evaluated in the \textbf{single-domain adaptation scenario} (\ie, the model parameters are reset before adapting to a new corruption type) except for MEMO~\cite{zhang2021memo}. We use * to denote episodic adaptation.}
    \label{supp:tab:imagenet-c-level-5}
\newcommand{\tabincell}[2]{\begin{tabular}{@{}#1@{}}#2\end{tabular}}
 \begin{center}
 \begin{threeparttable}
    \resizebox{1.0\linewidth}{!}{
        \begin{tabular}{c|c|c|ccc|cccc|cccc|cccc|c|cc}
        \multicolumn{3}{c}{} & \multicolumn{3}{c}{Noise} & \multicolumn{4}{c}{Blur} & \multicolumn{4}{c}{Weather} & \multicolumn{4}{c}{Digital} & \multicolumn{1}{c}{} & \multicolumn{2}{c}{Average} \\
        \toprule
         Model & Method & Metric & Gauss. & Shot & Impul. & Defoc. & Glass & Motion & Zoom & Snow & Frost & Fog & Brit. & Contr. & Elastic & Pixel & JPEG & Avg. & \#Forwards & \#Backards \\
        \midrule
        \multirow{16}{*}[-4.5ex]{R-50 (BN)} & \multirow{2}{*}{Source} & Acc. & 1.8 & 3.0 & 1.7 & 18.2 & 10.1 & 13.4 & 20.8 & 14.0 & 22.1 & 21.9 & 58.7 & 5.3 & 17.6 & 22.1 & 37.5 & 17.88 & \multirow{2}{*}{50,000} & \multirow{2}{*}{0} \\ & & ECE & 16.6 & 16.1 & 15.9 & 1.8 & 10.7 & 10.7 & 14.7 & 25.3 & 12.9 & 16.7 & 2.3 & 6.7 & 22.9 & 10.5 & 6.0 & 12.64 & & \\
                \cmidrule{2-21}
        & \multirow{2}{*}{BN Adapt}  & Acc. & 15.8 & 16.7 & 15.3 & 18.7 & 19.3 & 29.8 & 41.7 & 35.8 & 35.0 & 50.5 & 65.9 & 18.1 & 49.3 & 51.7 & 42.0 & 33.70 & \multirow{2}{*}{50,000} & \multirow{2}{*}{0} \\ & & ECE & 1.1 & 0.8 & 1.0 & 3.0 & 1.3 & 0.8 & 3.4 & 1.1 & 1.0 & 5.4 & 1.4 & 7.6 & 4.3 & 3.8 & 4.8 & 2.72 & &\\
        \cmidrule{2-21}
        & \multirow{2}{*}{Tent}   & Acc. & 28.0 & 30.1 & 28.1 & 29.9 & 29.5 & 42.2 & 49.7 & 46.2 & 41.5 & 57.7 & 67.1 & 30.0 & 55.7 & 58.3 & 52.5 & 43.11 & \multirow{2}{*}{50,000} & \multirow{2}{*}{50,000} \\ & & ECE & 11.7 & 11.2 & 11.1 & 12.6 & 12.3 & 7.7 & 5.4 & 6.5 & 8.8 & 3.4 & 2.9 & 21.9 & 3.5 & 3.7 & 4.0 & 8.46 & &\\
        \cmidrule{2-21}
        & \multirow{2}{*}{MEMO*}   & Acc. & 6.8 & 8.5 & 7.5 & 20.5 & 13.4 & 19.8 & 25.8 & 22.1 & 27.7 & 27.6 & 60.9 & 11.3 & 24.4 & 32.2 & 37.9 & 23.09 & \multirow{2}{*}{50,000$\times$65} & \multirow{2}{*}{50,000$\times$64} \\ & & ECE & 24.1 & 24.2 & 22.9 & 5.3 & 19.3 & 14.8 & 23.4 & 30.4 & 18.7 & 24.6 & 7.2 & 14.9 & 29.4 & 19.3 & 13.6 & 19.47 & &\\
        \cmidrule{2-21}
        & \multirow{2}{*}{CoTTA}   & Acc. & 19.9 & 20.8 & 19.4 & 17.4 & 22.1 & 36.3 & 47.9 & 41.7 & 39.9 & 55.8 & 66.9 & 22.1 & 54.6 & 57.5 & 47.2 & 37.97 & \multirow{2}{*}{151,753} & \multirow{2}{*}{50,000} \\ & & ECE & 3.6 & 4.5 & 3.5 & 10.0 & 7.7 & 5.2 & 3.2 & 3.8 & 4.9 & 1.8 & 3.8 & 16.7 & 2.4 & 2.4 & 1.4 & 5.00 & &\\
        \cmidrule{2-21}
        & \multirow{2}{*}{SAR}   & Acc. & 29.6 & 31.9 & 28.3 & 30.5 & 30.5 & 43.2 & 50.3 & 47.3 & 43.2 & 58.3 & 67.2 & 38.4 & 55.7 & 58.6 & 53.2 & 44.42 & \multirow{2}{*}{78,661} & \multirow{2}{*}{52,660} \\ & & ECE & 3.7 & 4.0 & 3.3 & 7.4 & 6.9 & 3.2 & 1.7 & 1.8 & 2.9 & 1.3 & 1.6 & 8.7 & 1.3 & 1.3 & 1.3 & 3.37 & &\\
        \cmidrule{2-21}
        &  & Acc. & 35.6 & 37.8 & 36.1 & 35.3 & 35.4 & 47.6 & 53.1 & 51.4 & 45.9 & 60.1 & 67.6 & 45.8 & 58.2 & 60.3 & 55.4 & 48.36 & & \\ & \multirow{-2}{*}{EATA (Ours)} & ECE & 10.5 & 10.2 & 10.3 & 12.9 & 12.3 & 8.3 & 6.6 & 6.7 & 8.2 & 4.7 & 4.5 & 9.9 & 5.1 & 5.3 & 5.6 & 8.08 & \multirow{-2}{*}{50,000} & \multirow{-2}{*}{26,188}\\
        \cmidrule{2-21}
        & & Acc. & 37.2 & 39.2 & 38.0 & 35.5 & 35.9 & 48.2 & 52.3 & 51.9 & 46.5 & 60.4 & 67.3 & 48.7 & 58.0 & 60.5 & 56.0 & 49.04 & &\\ & \multirow{-2}{*}{EATA-C (Ours)} & ECE & 7.1 & 6.8 & 6.8 & 9.3 & 8.9 & 4.3 & 3.6 & 3.0 & 4.5 & 1.8 & 1.9 & 4.5 & 2.3 & 2.4 & 2.4 & 4.63 & \multirow{-2}{*}{82,492} & \multirow{-2}{*}{32,492}\\
        \midrule
        \multirow{14}{*}[-4ex]{ViT (LN)} & \multirow{2}{*}{Source} & Acc. & 12.9 & 17.6 & 11.7 & 34.4 & 27.7 & 43.7 & 36.2 & 43.4 & 45.4 & 52.8 & 73.3 & 45.5 & 37.9 & 54.7 & 60.2 & 39.84  & \multirow{2}{*}{50,000} & \multirow{2}{*}{0}\\ & & ECE & 14.2 & 11.3 & 15.2 & 2.2 & 8.2 & 6.0 & 10.7 & 7.5 & 8.3 & 5.0 & 2.1 & 2.3 & 12.2 & 4.1 & 2.9 & 7.48 & &\\
        \cmidrule{2-21}
        & \multirow{2}{*}{Tent} & Acc. & 33.4 & 42.1 & 41.4 & 48.8 & 45.2 & 54.9 & 48.2 & 55.6 & 55.1 & 64.4 & 75.2 & 62.5 & 51.6 & 65.5 & 65.0 & 53.92 & \multirow{2}{*}{50,000} & \multirow{2}{*}{50,000} \\ & & ECE & 24.1 & 10.1 & 11.7 & 8.6 & 10.4 & 7.5 & 12.4 & 8.0 & 8.0 & 4.8 & 2.4 & 5.2 & 10.0 & 4.3 & 4.1 & 8.78 & &\\
        \cmidrule{2-21}
        & \multirow{2}{*}{MEMO*}   & Acc. & 32.2 & 35.1 & 32.6 & 37.5 & 28.5 & 43.3 & 40.1 & 45.2 & 47.0 & 53.9 & 73.3 & 53.3 & 39.6 & 59.5 & 62.8 & 45.60 & \multirow{2}{*}{50,000$\times$65} & \multirow{2}{*}{50,000$\times$64} \\ & & ECE & 33.0 & 32.5 & 33.2 & 36.7 & 45.7 & 41.0 & 47.2 & 39.8 & 38.8 & 33.7 & 20.0 & 26.7 & 48.5 & 30.4 & 27.2 & 35.63 & &\\
        \cmidrule{2-21}
        & \multirow{2}{*}{CoTTA}   & Acc. & 45.6 & 46.8 & 28.1 & 45.3 & 42.0 & 57.6 & 49.8 & 59.5 & 57.6 & 65.2 & 74.9 & 58.2 & 55.5 & 66.3 & 65.7 & 54.54 & \multirow{2}{*}{50,000$\times$3} & \multirow{2}{*}{50,000}\\ & & ECE & 9.0 & 8.9 & 9.9 & 7.9 & 12.9 & 8.1 & 12.1 & 7.1 & 9.0 & 6.4 & 4.3 & 6.6 & 10.7 & 6.7 & 6.6 & 8.42 & &\\
        \cmidrule{2-21}
        & \multirow{2}{*}{SAR}   & Acc. & 43.1 & 44.0 & 44.3 & 50.4 & 48.2 & 55.1 & 49.0 & 56.9 & 56.0 & 64.7 & 75.3 & 62.4 & 53.5 & 65.3 & 64.9 & 55.53 & \multirow{2}{*}{86,684} & \multirow{2}{*}{72,446} \\ & & ECE & 8.0 & 7.0 & 7.6 & 6.1 & 6.7 & 5.9 & 9.2 & 5.5 & 5.9 & 3.6 & 2.1 & 4.0 & 7.2 & 3.6 & 3.4 & 5.71 & &\\
        \cmidrule{2-21}
        &  & Acc. & 50.5 & 51.3 & 51.3 & 55.9 & 56.0 & 60.7 & 58.0 & 64.2 & 62.2 & 70.1 & 77.4 & 66.7 & 63.9 & 70.5 & 68.9 & 61.84 &  & \\ & \multirow{-2}{*}{EATA (Ours)} & ECE & 10.6 & 10.1 & 10.3 & 9.4 & 8.8 & 7.8 & 9.1 & 6.4 & 7.3 & 5.4 & 2.8 & 6.1 & 6.6 & 4.5 & 4.8 & 7.34 & \multirow{-2}{*}{50,000} & \multirow{-2}{*}{32,524}\\
        
        \cmidrule{2-21}
        &  & Acc. & 56.6 & 57.6 & 57.2 & 57.8 & 59.8 & 64.6 & 64.4 & 69.1 & 67.2 & 73.3 & 78.6 & 67.0 & 70.2 & 73.9 & 71.2 & 65.90 &  & \\ & \multirow{-2}{*}{EATA-C (Ours)} & ECE & 5.3 & 4.8 & 5.2 & 5.3 & 4.1 & 3.5 & 4.0 & 2.8 & 2.7 & 2.1 & 1.4 & 2.9 & 2.2 & 1.7 & 2.2 & 3.34 & \multirow{-2}{*}{82,364} & \multirow{-2}{*}{32,364} \\
    \bottomrule
        \end{tabular}
        }
         \end{threeparttable}
         \end{center}
\end{table*}

\begin{table*}[t]
        \renewcommand\thetable{B}
        \caption{
        Semantic segmentation results (mIoU in \%) on the Cityscapes-to-ACDC in lifelong TTA scenario. The model is continually adapted to the four adverse conditions for ten rounds without resetting model parameters, \ie, \textbf{lifelong adaptation}. All results are evaluated based on the Segformer-B5 model.}
    \vspace{-0.05in}
        \label{tab:suppl:acdc}
\newcommand{\tabincell}[2]{\begin{tabular}{@{}#1@{}}#2\end{tabular}}
 \begin{center}
 \begin{threeparttable}
    \resizebox{1.0\linewidth}{!}{
        \begin{tabular}{l|llll|llll|llll|llll|llll|l}
        \toprule
        Condition & Fog & Night & Rain & Snow & Fog & Night & Rain & Snow & Fog & Night & Rain & Snow & Fog & Night & Rain & Snow & Fog & Night & Rain & Snow & cont. \\
        \midrule
        Round & \multicolumn{4}{l}{1} & \multicolumn{4}{|l}{2} & \multicolumn{4}{|l}{3} & \multicolumn{4}{|l|}{4} & \multicolumn{4}{l|}{5} & cont. \\
        \midrule
        Source & 69.1 & 40.3 & 59.7 & 57.8 & 69.1 & 40.3 & 59.7 & 57.8 & 69.1 & 40.3 & 59.7 & 57.8 & 69.1 & 40.3 & 59.7 & 57.8 & 69.1 & 40.3 & 59.7 & 57.8 & cont. \\
        BN Stats Adapt & 62.3 & 38.0 & 54.6 & 53.0 & 62.3 & 38.0 & 54.6 & 53.0 & 62.3 & 38.0 & 54.6 & 53.0 & 62.3 & 38.0 & 54.6 & 53.0 & 62.3 & 38.0 & 54.6 & 53.0 & cont. \\
        TENT-continual & 69.0 & 40.2 & 60.0 & 57.3 & 68.4 & 39.1 & 60.0 & 56.4 & 67.6 & 37.9 & 59.7 & 55.3 & 66.6 & 36.6 & 58.9 & 54.2 & 65.9 & 35.3 & 57.9 & 53.3 & cont. \\
        CoTTA & 70.9 & 41.1 & 62.4 & 59.7 & 70.9 & 41.0 & 62.5 & 59.7 & 70.9 & 40.8 & 62.6 & 59.7 & 70.8 & 40.6 & 62.6 & 59.7 & 70.8 & 40.6 & 62.6 & 59.7 & cont. \\
        DAT & 71.7 & 44.6 & 63.8 & 62.2 & 70.3 & 44.3 & 62.5 & 61.3 & 69.1 & 43.4 & 61.4 & 60.3 & 68.0 & 42.0 & 60.9 & 59.4 & 66.8 & 40.9 & 60.4 & 59.0 & cont. \\
        \myeata & 69.1 & 40.5 & 59.8 & 58.1 & 69.3 & 41.1 & 60.0 & 58.4 & 69.3 & 41.5 & 60.1 & 58.6 & 69.3 & 41.8 & 60.1 & 58.6 & 69.2 & 42.1 & 59.9 & 58.5 & cont. \\ 
        \myeataC & 71.0 & 44.3 & 63.1 & 61.1 & 71.9 & 44.7 & 63.9 & 62.5 & 72.0 & 46.1 & 64.2 & 63.1 & 72.0 & 47.3 & 64.9 & 63.8 & 71.8 & 46.2 & 64.3 & 64.0 & cont. \\
        \midrule
        Round & \multicolumn{4}{l}{6} & \multicolumn{4}{|l}{7} & \multicolumn{4}{|l}{8} & \multicolumn{4}{|l|}{9} & \multicolumn{4}{l|}{10} & Mean \\
        \midrule
        Source & 69.1 & 40.3 & 59.7 & 57.8 & 69.1 & 40.3 & 59.7 & 57.8 & 69.1 & 40.3 & 59.7 & 57.8 & 69.1 & 40.3 & 59.7 & 57.8 & 69.1 & 40.3 & 59.7 & 57.8 & 56.7 \\
        BN Stats Adapt & 62.3 & 38.0 & 54.6 & 53.0 & 62.3 & 38.0 & 54.6 & 53.0 & 62.3 & 38.0 & 54.6 & 53.0 & 62.3 & 38.0 & 54.6 & 53.0 & 62.3 & 38.0 & 54.6 & 53.0 & 52.0 \\
        TENT-continual & 65.2 & 34.3 & 56.9 & 52.4 & 64.6 & 33.4 & 55.9 & 51.6 & 63.9 & 32.4 & 54.7 & 50.6 & 63.2 & 31.5 & 53.7 & 49.6 & 52.7 & 30.4 & 52.6 & 48.7 & 52.4 \\
        CoTTA & 65.2 & 34.3 & 56.9 & 52.4 & 64.6 & 33.4 & 55.9 & 51.6 & 63.9 & 32.4 & 54.7 & 50.6 & 63.2 & 31.5 & 53.7 & 49.6 & 52.7 & 30.4 & 52.6 & 48.7 & 58.4\\
        DAT & 66.4 & 40.7 & 59.7 & 58.3 & 66.1 & 40.6 & 59.8 & 57.8 & 65.6 & 40.3 & 59.3 & 56.8 & 65.1 & 39.7 & 58.7 & 56.0 & 63.8 & 39.6 & 58.2 & 55.4 & 57.0 \\
        \myeata & 69.0 & 42.3 & 59.7 & 58.2 & 68.8 & 42.5 & 59.4 & 57.9 & 68.6 & 42.7 & 58.9 & 57.4 & 68.3 & 42.8 & 58.4 & 56.9 & 67.9 & 42.8 & 57.7 & 56.3 & 57.0 \\ 
        \myeataC & 72.7 & 47.2 & 64.5 & 63.9 & 71.8 & 48.2 & 64.2 & 64.2 & 71.5 & 48.1 & 64.6 & 64.6 & 71.8 & 48.4 & 64.5 & 64.4 & 72.0 & 48.7 & 64.3 & 64.1 & 61.6\\
        \bottomrule
        \end{tabular}}
         \end{threeparttable}
         \end{center}
\end{table*}

 \begin{table*}[t]
        \centering
        \renewcommand\thetable{C}
        \caption{Effects of components in EATA-C. Results obtained in \textbf{single-domain TTA scenario}, \ie, the model parameters are reset before adapting to a new corruption. \textbf{CL} denotes consistency loss. \textbf{ER} denotes entropy regularization. \textbf{FR} denotes Fisher regularization. \textbf{SS} denotes active sample selection.}
        \label{suppl:tab:ablation-RC}
        \begin{minipage}{0.49\textwidth}
            \centering
            \newcommand{\tabincell}[2]{\begin{tabular}{@{}#1@{}}#2\end{tabular}}
 \begin{center}
 \begin{threeparttable}
    \resizebox{1.0\linewidth}{!}{
        \begin{tabular}{l|cccc|cccc}
        \toprule
          \multirow{2}{*}{Experiment} & \multirow{2}{*}{CL} & \multirow{2}{*}{ER} & \multirow{2}{*}{FR} &  \multirow{2}{*}{SS} & \multicolumn{4}{c}{\textit{ResNet-50} over \textit{ImageNet-R}}  \\
        ~ & ~ & ~ & ~ & ~ & Acc. & ECE & \#Forwards & \#Backwards \\
        \midrule
        Source & & & & & 38.0 & 17.7 & 30,000 & 0 \\
        1 & \checkmark & & & & 46.2 & 13.1 & 30,000$\times$2 & 30,000 \\
        2 & & \checkmark & & & 44.3 & 12.0 & 30,000$\times$2 & 30,000\\
        3 & \checkmark & \checkmark & & & 47.0 & 12.1 & 30,000$\times$2 & 30,000 \\
        \midrule
        4 & \checkmark & & \checkmark & & 45.8 & 13.2 & 30,000$\times$2 & 30,000\\
        5 & & \checkmark & \checkmark & & 43.9 & 12.3 & 30,000$\times$2 & 30,000\\
        6 & \checkmark & \checkmark & \checkmark & & 46.7 & 12.3 & 30,000$\times$2 & 30,000\\
        \midrule
        7 & \checkmark & & & \checkmark & 46.9 & 14.4 & 35,147 & 5,147\\
        8 & & \checkmark & & \checkmark & 44.6 & 11.9 & 34,991 & 4,991\\
        9 (ETA-C) & \checkmark & \checkmark & & \checkmark & 46.8 & 13.4 & 35,137 & 5,137 \\
        10 & \checkmark & & \checkmark & \checkmark & 47.1 & 13.7 & 35,179 & 5,179\\
        11 & & \checkmark & \checkmark & \checkmark & 44.6 & 12.1 & 34,935 & 4,935\\
        12 (EATA-C) & \checkmark & \checkmark & \checkmark & \checkmark & 47.1 & 13.3 & 35,122 & 5,122\\
        \bottomrule
        \end{tabular}
        }
         \end{threeparttable}
         \end{center}
        \end{minipage}
        \hfill
        \begin{minipage}{0.49\textwidth}
            \centering
            \newcommand{\tabincell}[2]{\begin{tabular}{@{}#1@{}}#2\end{tabular}}
 \begin{center}
 \begin{threeparttable}
    \resizebox{1.0\linewidth}{!}{
        \begin{tabular}{l|cccc|cccc}
        \toprule
          \multirow{2}{*}{Experiment} & \multirow{2}{*}{CL} & \multirow{2}{*}{ER} & \multirow{2}{*}{FR} &  \multirow{2}{*}{SS} & \multicolumn{4}{c}{\textit{ViT-Base} over \textit{ImageNet-C} (level 5)}  \\
        ~ & ~ & ~ & ~ & ~ & Acc. & ECE & \#Forwards & \#Backwards \\
        \midrule
        Source & & & & & 39.8 & 7.5 & 50,000 & 0 \\
        1 & \checkmark & & & & 65.6  & 3.3  & 50,000$\times$2 & 50,000\\
        2 & & \checkmark & & & 53.4 & 4.1 & 50,000$\times$2 & 50,000\\
        3 & \checkmark & \checkmark & & & 65.9 & 2.5 & 50,000$\times$2 & 50,000 \\
        \midrule
        4 & \checkmark & & \checkmark & & 66.0 & 3.3 & 50,000$\times$2 & 50,000\\
        5 & & \checkmark & \checkmark & & 53.0 & 4.5 & 50,000$\times$2 & 50,000\\
        6 & \checkmark & \checkmark & \checkmark & & 66.4 & 2.6 & 50,000$\times$2 & 50,000\\
        \midrule
        7 & \checkmark & & & \checkmark & 65.6  & 4.4  & 83,182 & 33,182\\
        8 & & \checkmark & & \checkmark & 52.9  & 3.4  & 73,323 & 23,323\\
        9 (ETA-C) & \checkmark & \checkmark & & \checkmark & 65.3  & 3.4  & 82,015 & 32,015 \\
        10 & \checkmark & & \checkmark & \checkmark & 66.2  & 4.3  & 83,429 & 33,429\\
        11 & & \checkmark & \checkmark & \checkmark & 53.2  & 3.4  & 73,723 & 23,723\\
        12 (EATA-C) & \checkmark & \checkmark & \checkmark & \checkmark & 65.9  & 3.3 & 82,364 & 32,364\\
        \bottomrule
        \end{tabular}
        }
         \end{threeparttable}
         \end{center}
    \end{minipage}
\end{table*}

\begin{table}[t]
    \centering
    \renewcommand\thetable{D}
    \caption{Comparison of parameter update strategies. EATA-C (Ours) updates only the normalization layer parameters, while EATA-C$^\dagger$ updates all model parameters. Results are obtained on ImageNet-C (Gaussian, level 5) with ViT-Base. \textit{Loss} denotes the average KL-Divergence between the full network and the sub-network across the dataset during online TTA. Time and peak memory usage are measured by processing 50,000 images.}
        \label{tab:norm_or_all_parameters}
    \vspace{-0.1in}
\newcommand{\tabincell}[2]{\begin{tabular}{@{}#1@{}}#2\end{tabular}}
 \begin{center}
 \begin{threeparttable}
    \resizebox{.5\linewidth}{!}{
        \begin{tabular}{l|ccccc}
        \toprule
        \multirow{1}{*}{\tabincell{c}{Method}}&\multirow{1}{*}{\tabincell{c}{\textit{Loss}}}&\multirow{1}{*}{\tabincell{c}{Acc. (\%)}}&\multirow{1}{*}{\tabincell{c}{ECE (\%)}}&\multirow{1}{*}{\tabincell{c}{Time (s)}}&\multirow{1}{*}{\tabincell{c}{Memory (MB)}}\\
        \midrule
        Source & 0.565 & 12.9 & 14.2 & 55.7 & 816.6\\
        SAR & - & 43.1 & 8.0 & 167.3 & 7433.2 \\
        \midrule
        EATA-C$^\dagger$ & 0.437 & 56.0 & 5.9 & 140.9 & 9920.5 \\
        EATA-C (Ours) & 0.451 & 56.8 & 5.2 & 114.9 & 5786.6 \\
        \bottomrule
        \end{tabular}}
         \end{threeparttable}
         \end{center}
\end{table}

\end{document}